\documentclass[journal]{IEEEtran}
\ifCLASSINFOpdf
\else
\usepackage{ifpdf}
 \ifpdf
 \else
 \fi
 
 \ifCLASSINFOpdf
   \usepackage[pdftex]{graphicx}
\else
\fi

\usepackage{enumitem}
\usepackage{tabularx,booktabs}

\usepackage{amsmath,amssymb,amsfonts}
\usepackage{array} 
\usepackage[justification=centering]{caption}
\makeatletter
\let\NAT@parse\undefined
\makeatother
\usepackage{hyperref}  
\usepackage{caption}
\usepackage{subcaption}
\usepackage{float}
\setlist[enumerate]{listparindent=\parindent}
\usepackage{algpseudocode}
\usepackage{epsfig}
\usepackage{multirow}
\usepackage{url}
\usepackage{footnote}
\usepackage{cite}

\begin{document}
\captionsetup{font={footnotesize}}


\title{Evolutionary Programmer: Autonomously Creating Path Planning Programs based on Evolutionary Algorithms}

\author{Jiabin Lou, 
            Rong Ding,
            and Wenjun Wu
\thanks{Manuscript received XX XX, 2021}
\thanks{This work was supported by the National Key Research and Development Program of China under Grant No.2017YFB1001802.}
\thanks{The authors are with the National Laboratory of Software Development Environment, School of Computer Science, Beijing University of Aeronautics and Astronautics, Beijing, 100191, China (e-mail: loujiabin@buaa.edu.cn; dingr@buaa.edu.cn; wwj@nlsde.buaa.edu.cn).}
}

%
%

\markboth{IEEE TRANSACTIONS ON EVOLUTIONARY COMPUTATION}%
{Shell \MakeLowercase{\textit{et al.}}: Bare Demo of IEEEtran.cls for IEEE Journals}

\maketitle

\begin{abstract}
Evolutionary algorithms are wildly used in unmanned aerial vehicle path planning for their flexibility and effectiveness. Nevertheless, they are so sensitive to the change of environment that can't adapt to all scenarios. Due to this drawback, the previously successful planner frequently fail in a new scene. In this paper, a first-of-its-kind machine learning method named Evolutionary Programmer is proposed to solve this problem. Concretely, the most commonly used Evolutionary Algorithms are decomposed into a series of operators, which constitute the operator library of the system. The new method recompose the operators to a integrated planner, thus, the most suitable operators can be selected for adapting to the changing circumstances. Different from normal machine programmers, this method focuses on a specific task with high-level integrated instructions and thus alleviate the problem of huge search space caused by the briefness of instructions. On this basis, a 64-bit sequence is presented to represent path planner and then evolved with the modified Genetic Algorithm. Finally, the most suitable planner is created by utilizing the information of the previous planner and various randomly generated ones.
\end{abstract}


\IEEEpeerreviewmaketitle

\section{Introduction}
\label{Introduction}
\IEEEPARstart{U}{nmanned}  Aerial Vehicles (UAVs)  have shown great potential in several areas, e.g., military, logistics, agriculture, disaster rescue, etc. In recent years,  autonomous path planning has drawn growing significance to the navigation process of UAVs, as the traditional human pilots, even have been highly trained, can't sustain control precision when the UAV is entrusted in complex missions \cite{Survey_of_UAV_mp}. \par
The path planning problem is often formulated as a Constraint Satisfaction Problem (CSP), where the feasibility of a path lies on the constraints imposed by environments, missions and UAV itself. Meanwhile, the evaluation of a feasible path depends on the cost including length, risk, and fuel. Generally, a path is represented as a set of segments, i.e., line or curve, from the start point to the desired target. Therefore, the path planning problem is a CSP to find out a set of waypoints optimizing multiple objectives subject to various constraints. It has been verified this problem is NP-hard, which means the time complexity increases rapidly when the problem size grows, especially in a 3-dimensional environment with various obstacles. \par 
Over the past few years, a bunch of UAV path planning methods have been proposed by scholars, for example, search-based methods, such as voronoi diagram methods \cite{vonoi}, A-star algorithm \cite{A-star} and Gradient-based methods \cite{gradian}; sampling-based methods, such as Rapidly-exploring Random Tree (RRT) \cite{RRT}, Probabilistic Road map (PRM) \cite{PRM}; and nature-inspired methods,i.e.,Evolutionary Algorithms (EAs), including Genetic Algorithm (GA) \cite{Roberge2013, Cao2019, Karakatic2015, Razali2011, Sahingoz2014, Elhoseny2018,  Ghambari2019,Roberge2018}, Particle Swarm Optimization (PSO) \cite{Wu2018,Shao2020, Roberge2018,Wang2016}, Differential Evolution (DE)\cite{Yu2020,Pan2020,Yang2015,Zhang2015}, Grey Wolf Optimization (GWO) \cite{YongBo2017,Qu2020}, etc. Among these methods, the EA-based planners are generally more effective and flexible to this problem, especially in cases where complex environments and numerous obstacles lead to the rapid shrink of the feasible solution area in the huge search space. However, in the problem of motion planning for robotics, which is quite similar to UAV path planning, the performance of  traditional methods is remarkable because of the relatively small search space (i.e., 2-dimensional space) and constraints (e.g., it doesn't require considering the smoothness of the path). But as the problem gets more complicated, the more obvious EA's superiority is. This is because that EA needn't compute the configuration space, which is quite time-consuming, and more importantly, they normally have powerful global search capabilities due to their population-based nature, which partly prevents the algorithm from falling into local optima.\par
In EAs, when searching a feasible path, it generally consists of five phases, i.e., initialization, fitness computing, sorting and selection, exploitation and exploration. Although with different names for these phases in different algorithms, they are essentially similar, e.g., when using GA to solve this problem, it instantiate these phases as \textit{genome} initialization, \textit{fitness} computing, \textit{selection}, \textit{crossover} and \textit{mutation}. Hence, these algorithms have the same framework.  But there are quite differences among these algorithms, and even the same phase in same algorithm has various variants with numerous parameters to adapt to different environments. In other words, the EAs are very sensitive to the environment, so they still often fail or fall into local optima in practice.\par
Taking a closer look at this specific problem, the search space changes dramatically when the UAV fly in a completely different area. This implies that an algorithm that works effectively in one area does not always perform well in another region, and researchers are usually unable to determine which algorithm has a high performance before implementation. Therefore, the main idea adopted in this paper is to autonomously create EA-based path planning algorithm for UAV in a given flying environment.\par
There are always a dream for Artificial Intelligence (AI) to write programs. However, significant challenge still exist when the AI tries to create a generic program without human participation; that is, the AI can hardly learn to read codes written by humans and thus can't understand human logic. The existing auto programmers, such as DeepCoder \cite{Deepcoder}, AI Programmer \cite{Becker2017} and Robustfill \cite{RobustFill}, etc., all have their own internal limitations. First, the instructions used to construct a program should be Turing complete. However, this character makes the instructions are always not very complex and thus may easily lead to bugs that need human involvement to repair. Moreover, the simple instruction set also means the huge search space so that creating a program is very time-consuming. Second, the auto programmers are generally unable to accept tasks directly, in other word, the communication between human programmers and machines is always difficult. It usually requires humans to abstract the mathematical model out of the problem, and it's not a piece of cake. \par
On this basis, we have proposed a first-of-its-kind auto programmer framework-Evolutionary Programmer (EP), which can automatically create suitable UAV path planning program based on EA in different flying environments. Distinct from the generic auto programmer discussed above, the instruction set in EP is high-level integrated and internal sequential because of the stationary five phases of EAs. This way, the EP can focus more on seeking a feasible solution with fewer bugs, and the performance of generated program depends a lot on the operator library. Therefore, we have chosen the most commonly used algorithms (i.e., GA, PSO, DE, GWO, Symbiotic Organisms Search (SOS) and their variants) with some other general strategies like Elitism, Rank system, Migration, etc., to build the operator library, which will be further introduced in Section \ref{using_EA}.\par
Finally, all operators are encoded as binaries, and every 64-bit sequence represent a path planner. So we can focus on predicting an order on these operators and use it to guide the working of EP. Due to the outstanding performance of GA in automatic programming \cite{Yampolskiy2018}, we have chosen it to evolve the path planners (i.e., the original one as well as various randomly generated planners) when the UAV enters a new environment. After a few periods of iterations, the most suitable planner for this environment is created.\par
The rest of this paper is organized as follows: Section \ref{Problem Descroption} proposes the mathematical model of the UAV path planning problem, including the representation and generation of scenarios and paths, the evaluation functions of the objectives and constraints. The framework of EAs to solve the problem is further introduced in Section \ref{using_EA}. In Section \ref{EP}, the operator libraries is constructed and the EP is then proposed in detail. Finally, the conclusions are discussed in Section \ref{conclusion}, and the source code is released for the reference: \url{https://github.com/loujiabin1994/Evolutionary_Programmer}.\par
\section{Problem Description}
\label{Problem Descroption}
There are quite a few important factors (e.g., the description of environments, the representation of a path, the cost of missions, the constraints of the problem, etc.) that should be taken into consideration when planning a path for UAV. In this paper, the UAV path planning problem is treated as a multi-objective constrained optimization problem, and the mathematical model is described as follow.\par
\subsection{Scenarios Representation}
\label{Scenarios}
The scenarios for UAV path planning problem consist of the following components, terrain, obstacles, and the start point as well as the target point.\par
\begin{figure}[htb!]
\centering
   \begin{subfigure}[b]{0.24\textwidth}
        \centering
       \includegraphics[width=0.85\textwidth]{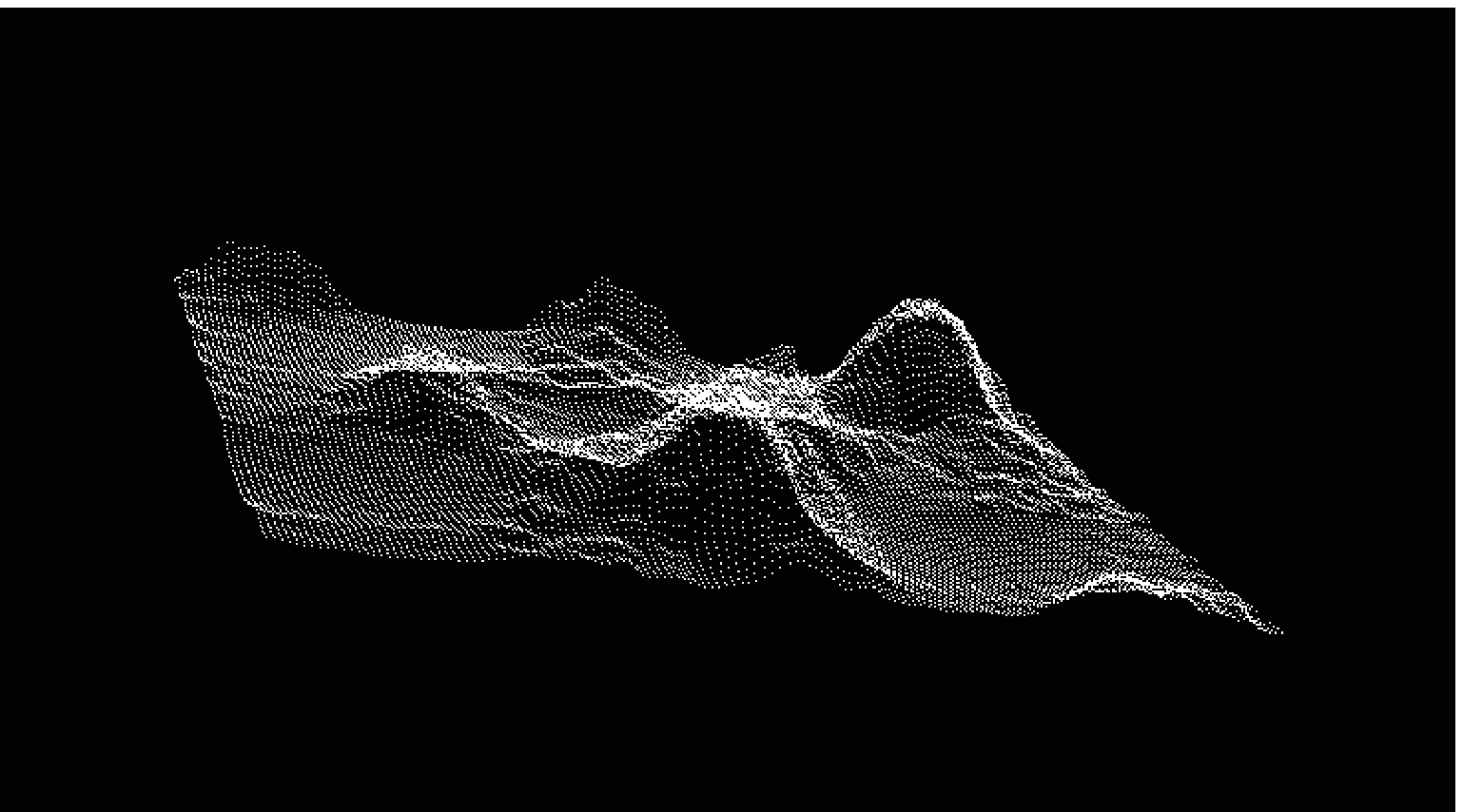}
        \caption{}
        \label{terrain.a}
    \end{subfigure}
    \hfill
   \begin{subfigure}[b]{0.24\textwidth}
        \centering
       \includegraphics[width=0.95\textwidth]{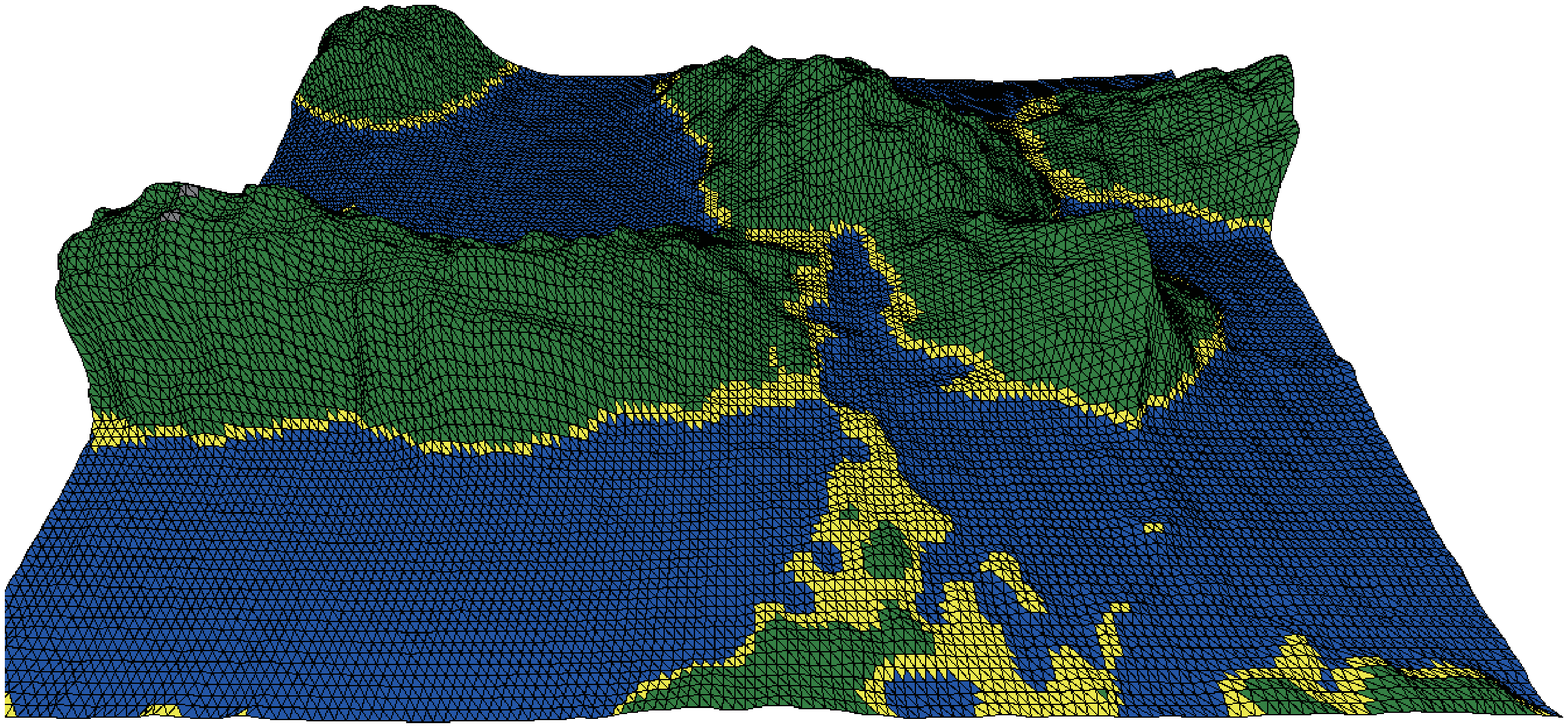}
        \caption{}
        \label{terrain.b}
    \end{subfigure}
\caption{
\centering
Terrain Representation. (a) describes the terrain by point cloud, which is randomly generated using noise, (b) renders (a) by interpolation method so that every point in space is constrained by terrain}
\label{terrain}
\vspace{-5mm}
\end{figure}

 
In this paper, we assume the terrain is open, implying that we can discretize the broad planning space into a surface. On this basis, the main physics obstacles in the space are the terrain, which is depicted as the point cloud and randomly generated using several layers of noises with diverse frequency and amplitude, see Fig.\ref{terrain.a}. But such a discretized representation of terrain can't constrain all points in planning space, so we have used triangle interpolation among every three nearest points thus confirmed the minimum height of all points in the space, as shown in Fig.\ref{terrain.b}.\par
Conventionally, other obstacles (i.e., radar detection, missile attacking, and entering no-fly zone) are treated as soft barriers \cite{Xue2018}, we call them threats in the latter. 
Finally, the starting and ending points are randomly generated within the safe area of the scenario, but in the simulation of this paper, we selected some scenarios with the same beginning and terminus as the display for easy comparison. \par

\subsection{Path representation and heuristics}
\label{path representation}
    In EAs, the initial population is usually randomly generated without any constraints or heuristics for the sake of ignoring the characteristics of the specific problem. But in this problem, such a strategy will fail with large probability because of the strong constraints inside a path and the huge search space. It's wildly accepted that some heuristics can help the algorithm converge faster.
    \subsubsection{Rotated Coordinate System}
\begin{figure}[htb!]
        \centering
        \includegraphics[width=0.42\textwidth]{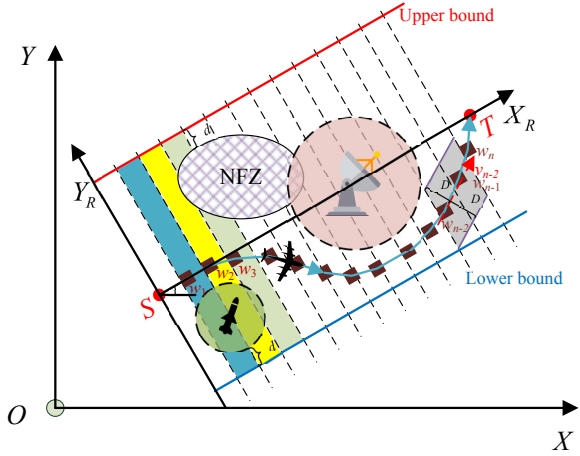}
        \caption{Division of UAV mission space}
        \label{division of mission space}
    \end{figure}

     In most studies, a path is usually represented as a set of waypoints within a global 3-D Cartesian coordinate system, which leads to a large search space since the direction of UAV is rarely consistent with the axes. To solve this problem, the rotated coordinate system $O\!_{R}$-$X\!_{R}Y\!_{R}Z$, as shown in Fig.\ref{division of mission space}, is established, where $X_R$ is the direction of the starting point to the target and $Y_R$ is perpendicular to $X_R$ on the horizontal plane. The transformation between the two coordinates can be obtained by the rotating matrix:
    \begin{equation}
        \label{rotate_matrix}
        \left(\!\begin{array}{l}x \\ y \\ z\end{array}\!\right)\!=\!\left(\!\begin{array}{l}x_{R} \\ y_{R}\\z_{R}\end{array}\!\right)\\\!\left(\!\begin{array}{ccc}\cos \theta & \sin \theta& 0 \! \\\!-\sin \theta & \cos \theta & 0 \!\\\! 0 & 0 & 1\end{array}\!\right)\! +\!\left(\begin{array}{l}x_S \\ y_S \\ 0\end{array}\!\right)
    \end{equation}
    where $\theta$ is the angle from $X$ axis of the origin frame to $X_R$ axis of the rotated frame.\par
    \subsubsection{Control points}
    To ensure $C^2$ continuity of the generated curve, we use control points $\boldsymbol{p}=[p_1, p_2,\ldots,p_n]$ to denote a path. In $O_{R}$-$X_{R}Y_{R}Z$, we can assume that $\boldsymbol{p}$ is monotonically increasing along the $X_R$ axis which means that the UAV cannot move backwards. Actually, some other scholars \cite{Zhang2015, Yang2015} have also utilized this benefit in the phase of path initialization to reduce the search space and they have proved that the solution set is almost never lost in this way. Furthermore, the $x$ coordinate of each waypoint can roughly be locked in a fixed area since the velocity of the UAV should approximately remain constant or tiny change for the purpose of saving fuel. So We divide the $ST$ by $(n+1)$  $\Delta l$ segments, and each equilateral point is the expected position of the control points. Considering that the distribution of each point is different and correlated, we can limit a point's $x$ coordinate value between the expected value of the previous and the next.\par
    Although we have effectively decreased the search space of the problem, the computational cost of the task remains huge. Considering that the $y$ coordinate is unconstrained, there could be a lot of invalid points that away from the planned path. Therefore some external constraints can be added to the control points. As is shown in fig.\ref{division of mission space}, two lines have limited the upper bound as well as lower bound, which are determined by extending outward a constant distance $\Delta d$ of the points from the nearest safe areas around the line $ST$. Under the boundary, the $y$ coordinate is restricted in a certain scope $[y_{min}, y_{max}]$, which are calculated as \eqref{y_min} and \eqref{y_max}, respectively:
    \begin{equation}
        \label{y_min}
        y_{\min} =\min\left\{\min _{i}\left\{y_{\text {threat}, i}^{*}-R_{i}\right\}, 0\right\}-\Delta d
    \end{equation}
    \vspace{-4mm}
    \begin{equation}
        \label{y_max}
        y_{\max} =\max\left\{\max _{i}\left\{y_{\text {threat}, i}^{*}+R_{i}\right\}, 0\right\}+\Delta d
    \end{equation}
    where $R_i$ is the radius of the $i$th threat, $y_{\text {threat}, i}^{*}$ is the vertical coordinate of $i$th threat in the rotated coordinate frame $O\!_{R}$.\par 
    Until now, every point in $\boldsymbol{p}$ is locked in a fixed area on the horizontal plane (e.g. $w_2$ is locked in blue and yellow areas, and $w_3$ is locked in yellow and green areas, and so forth). In addition, the velocity direction of UAV should not change aggressively for the reason of saving fuel. Based on this criterion, as seen in Fig.\ref{division of mission space}, we define two boundaries parallel to the velocity to constrain the velocity direction variation trend with the metric $\Delta D$ equal to $\Delta l$. So at the time of initialization, we can roughly determine the $x$ and $y$ range of $p_{i+1}$ according the position of $p_i$ (e.g. $p_{n-1}$ is locked in the gray area). This restrict is modeled as a Markov Chains, which require a considerable computational cost to be handled, so it is used only as an initial condition for increasing the feasibility of initial population. \par
    So far we have defined the range of $x$ and $y$ of every points, but things are a little different for $z$ coordinate. As a general rule, when flying at low altitude, the UAV can avoid radars on account of the terrain mask effect, and to a certain extent, fly at high altitude need more energy on takeoff and landing. Therefore, the minimum flight altitude is depicted as a constraint to the problem, and the flight altitude, which should be as lower as possible, is formulated as an objective function.\par

    \subsubsection{Generating approximate continuous path from control points}
    As discussed above, we used control points to represent a  path. Nevertheless, the fuzzy situation between adjacent control points has hitherto caused some trouble for the computation of objective functions and constraint conditions. In other words, the ambiguity of the real behaviors between control points can hardly meet the requirements of kinematics and dynamics. There is a good deal of methods named smoother to address this issue. After this phase, the $C^2$ continuous waypoints $\boldsymbol{w}=[w_1, w_2,\ldots,w_m]$ are generated. More details about the smoother will be introduced in section \ref{smooth_fitness}. \par
\subsection{Optimization Model}
Mathematically, the UAV path planning problem can be modeled as a Multi-Objective Constraint Satisfaction Problem, whose objectives and the constraints are listed as follow.
\subsubsection{Objective function}
In this work, five objectives, i.e., length cost, flight altitude, radar detection, missile attacking and turning angle, will be considered to obtain a shortest, safest and smoothest path. These objectives are commonly in conflict, which means the improvements in one objective cannot be achieved without detriment to the others. In order to solve such a counterbalance problem, a weighted function \eqref{muti-object} was proposed.
    \begin{equation}
        \label{muti-object}    
        F=\omega_1{f_1}+\omega_2{f_2}+\omega_3{f_3} + \omega_4{f_4} + \omega_5{f_5}
    \end{equation}
where $\omega_1$, $\omega_2$, $\omega_3$, $\omega_4$, $\omega_5$ are the weights whose sum is $1$, and $f_1$, $f_2$, $f_3$, $f_4$, $f_5$ denote the objective function value, which will be described in the following subsections.
\begin{enumerate}[labelsep = .5em, leftmargin = 0pt, itemindent = 3 em]
    \item[(1)]Length cost\par
    Traditionally, the energy and time costs are highly related to the length of the UAV flight path. Under the same conditions of velocity and path smoothness, the longer the path, the huger the energy and time costs. The normalized approximate length of a path is described as follow:
    \begin{equation}
        \label{length}
        f_{1}\!=\!\frac{\sum_{i=2}^{m} \!\sqrt{\left(x_{i}\!-\!x_{i-1}\right)^{2}\!+\!\left(y_{i}\!-\!y_{i-1}\right)^{2}\!+\!\left(z_{i}\!-\!z_{i-1}\right)^{2}}}{\sqrt{\left(x_{m}\!-\!x_{1}\right)^{2}\!+\!\left(y_{m}\!-\!y_{1}\right)^{2}\!+\!\left(z_{m}\!-\!z_{1}\right)^{2}}}
    \end{equation}
    where $x_i$ denote the $i$th waypoint of $\boldsymbol{w}$.
    \item [(2)]Flight altitude\par
    According to \ref{path representation}, a lower altitude is desired for the sake of avoiding radars and saving fuel, \eqref{flight_altitude} denotes the mean flight altitude of the path.
    \begin{equation}
        \label{flight_altitude}
        \begin{split}
            & \quad\quad\quad\quad\quad\quad f_{2}=\sum_{i=2}^{m} \mathrm{FA}_{i}\quad \text {with }\\[-2mm]
           \mathrm{FA}_{i}  &
            \begin{aligned}
                =\!\left\{\!\begin{array}{ll}0, \!&\text { if }  z_{i} \!\leq\! Map\left(x_{i}, y_{i}\right) \\ \left(z_{i}\!-\!\textit{Map}\left(x_{i}, y_{i}\right)\right)\! /\! m, \! & \text { otherwise }\end{array}\right.
            \end{aligned}
        \end{split}
    \end{equation}
    Where $Map\left(x_i, y_i\right)$ represent the ground height at the point $\left(x_i,y_i\right)$.
    \item [(3)]Radar detection\par
    The minimized probability of detection by hostile radars is going to guarantee the safety of flight. In this paper, we minimize the probability of drones being detected by radar to zero if it's possible, otherwise to the lowest risk. This probability can be calculated by \eqref{p_radar} \cite{Radar_RCS}.
    \begin{equation}
        \label{p_radar}
        P_{R}=\left\{\begin{array}{ll}0 & \text { if } \quad d>R_{R \max } \\ \frac{1}{1+\zeta_{2}\left(d^{4} / RCS\right)^{\zeta_1}} & \text { otherwise }\end{array}\right.
    \end{equation}
    where $\zeta_1$ and $\zeta_2$ depend on the used radar, $RCS$ denotes the radar cross section, which is calculated by \eqref{RCS},
        \begin{equation}
        \label{RCS}
        RCS=\frac{\pi a^{2} b^{2} c^{2}}{\sqrt{\left(a \alpha_{z} \beta_{\phi}\right)^{2}+\left(b \alpha_{z} \alpha_{\phi}\right)^{2}+\left(c \beta_{z}\right)^{2}}}
    \end{equation}
     where $\alpha_z = sin \psi^e$, $\beta_z = cos \psi^e$, $\alpha_\phi = sin \phi^e $, $\beta_\phi = cos \phi^e$, and $\psi^e$ is the angle between the velocity vector of the UAV and the segment that joins the UAV and radar positions, $\phi^e = \phi - arctan(tan \theta/ sin \psi)$ with $\phi$, $\theta$ and $\psi$ denote the roll, elevation and azimuth between the positions of the UAV and the radar. \par
     To sum up, the objective of radar detection can be represented as follow:
    \begin{equation}
        \label{radar}
        f_{3}= \sum_{i=1}^{R} \sum_{j=2}^{m}{P_R}_{i j}
    \end{equation}
    where $R$ is the number of radars.
    \item [(4)]Missile attacking\par
    For the same reason, the probability of missile attacking should be as lower as possible. This probability, which depend on the distance between UAV and missile center $d$ as well as the maximum hit radius of missile $R_M$, can be calculated by \eqref{missile}, so the objective of this part is given by \eqref{f_missile}.
    \begin{equation}
        \label{missile}
        {P_M}=\left\{\begin{array}{ll}\frac{R_{M }^{4}}{R_{M}^{4}+d^{4}}, & \text { if } d \leq R_{M} \\ 0, & \text { otherwise. }\end{array}\right.
    \end{equation}
    \begin{equation}
        \label{f_missile}
        f_{4}= \sum_{i=1}^{M} \sum_{j=2}^{m} \mathrm{R_M}_{i j}
    \end{equation}
    \item [(5)]Turning angle\par
    This objective is designed to calculate the degree of the path bending, the smaller the average turning angle, the more smooth the path, see \eqref{turning_angle}.
  
        \begin{equation}
        \label{turning_angle}
        \begin{split}
       & \quad\quad\quad\quad\quad\quad f_{5}=\sum_{i=2}^{m} \mathrm{\theta}_{i} \quad \text{with} \\[-2mm]
       &\theta_{i}\!=\!\arccos\!\left(\!\frac{\left(x_{i}\!-\!x_{i-1}, y_{i}\!-\!y_{i-1}\right) \!\cdot\!\left(x_{m}\!-\!x_{i}, y_{m}\!-\!y_{i}\right)^{T}}{\left\|\left(x_{i}\!-\!x_{i-1}, y_{i}\!-\!y_{i-1}\right)\!\cdot\!\left(x_{m}\!-\!x_{i}, y_{m}\!-\!y_{i}\right)\right\|}\!\right)\!  
    \end{split}
    \end{equation}
\end{enumerate}
\subsubsection{Constraints}
    When planning a path, several constraints should be taken into account for safe flight. The mainstream way of constraint-handling consists of penalty function, feasibility keeping methods and hierarchy sort, etc. EP deals with these strategies as operators, seen in \ref{smooth_fitness} with detailed discussion, and here we will introduce the concrete constraints as the following.
    \begin{enumerate}[labelsep = .5em, leftmargin = 0pt, itemindent = 3 em]
    \item [(1)]Climbing/gliding\par
   Since the maneuverability of a UAV, the slope (i.e. the climbing and gliding angle), which describe the change of flying orientation in the vertical coordinate, should be restricted in the range given by \eqref{climbing}-\eqref{s_k}.
   {\setlength\belowdisplayskip{0 em}
    \begin{equation}
    \label{climbing}
       g_{1}=\max \left(s_{k}-\alpha_{k}\right) \leq 0
    \end{equation}}
    \vspace{-3mm}
    \begin{equation}
        \label{gliding}
        g_{2}=\max \left(\beta_{k}-s_{k}\right) \leq 0
    \end{equation}
    For $k$ in $2, \ldots, m,$ where
    \begin{equation}
        \alpha_{k}=-1.5377 \times 10^{-10} z_{k}^{2}-2.6997 \times 10^{-5} z_{k}+0.4211
    \end{equation}
    \begin{equation}
        \beta_{k}=2.5063 \times 10^{-9} z_{k}^{2}-6.3014 \times 10^{-6} z_{k}-0.3257
    \end{equation}
    \begin{equation}
        \label{s_k}
        s_{k}=\frac{z_{k+1}-z_{k}}{\sqrt{\left(x_{k+1}-x_{k}\right)^{2}+\left(y_{k+1}-y_{k}\right)^{2}}}
    \end{equation}
    \item [(2)]Minimum flight altitude\par
    As discussed in \ref{path representation}, the UAV should fly above a appropriate altitude for the sake of safety. \par
    For $k$ in $2, \ldots, m$,
    \begin{equation}
        g_{3}=H_{\text {safe }}-\min \left(z_{k}-Map\left(x_{k}, y_{k}\right)\right) \leq 0
    \end{equation}
    where $H_\text{safe}$ denotes the safe height above the ground.
    \item [(3)]Forbidden flying area\par
    This constraint is introduced to avoid getting into the predefined no-fly zone (NFZ), which is high risky for the UAV. Therefore, in \eqref{NFZ}, the NFZs were established as hard constraints that the UAV must keep away from.
    \begin{equation}
        \label{NFZ}
       h_1 = \sum_{i=1}^{m}{\mathrm{InNFZ}}(w_i) = 0
    \end{equation}
    \item [(4)]Mission space \par
    As depicted in Fig.\ref{division of mission space}, the rectangles have restricted the range of control points, hence:
    \begin{equation}
        \label{mission_range}
        h_2 = \sum_{i=1}^{n}{\mathrm{OutRange}}(p_i) = 0
    \end{equation}
\end{enumerate}

\section{Using EA to solve problem}
\label{using_EA}
\begin{figure}[htb!]
        \centering
        \includegraphics[width=0.35\textwidth]{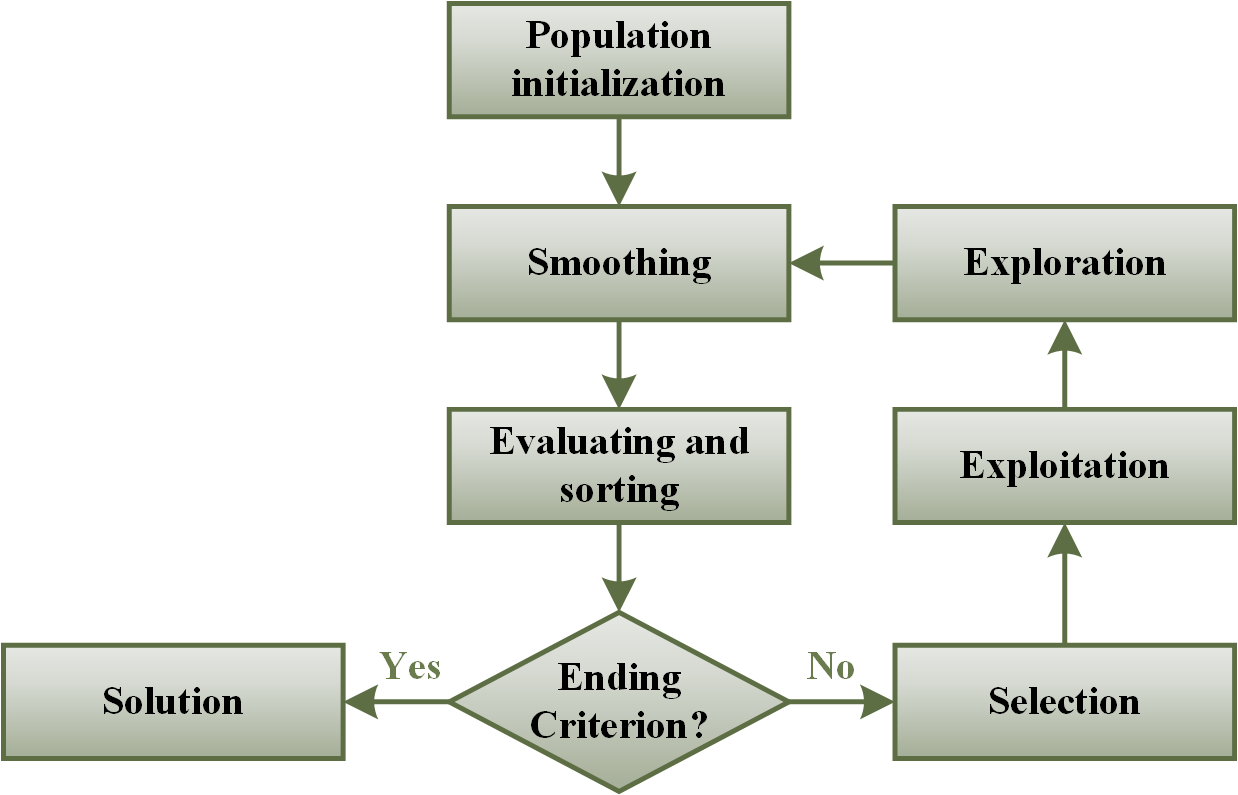}
        \caption{Overview of Evolutionary algorithm for path planning}
        \label{flowchart_EA}
    \end{figure}

    EA is a cluster of Meta-heuristic algorithms inspired by natural intelligence, implying that they are population-based, derived from mimicking the behaviors of the organisms. 
    The main phases of EAs to solve the UAV path planning problem are shown in Fig.\ref{flowchart_EA} and will be discussed in detail in the following subsection.
    \vspace{-5mm}
    \subsection{Encoding Approach and Initialization}
    As discussed in section \ref{path representation}, the initialization of individuals could be guided with some heuristics. In this paper, three criterion are proposed:\par
    \begin{itemize}[topsep=0pt]
        \item \textbf{$x$ coordinate}
    \end{itemize}\par
    In the rotated coordinates system which we have built before, the $x$ coordinate value of the control point is monotonically increasing. Moreover, we assume that $x_i$ obeying the normal distribution with mean $(i\Delta l)$ and standard deviation $(\Delta l/3)$ so that its value will fall in the expected range we defined in \ref{path representation} with 99.7$\%$ probability according to the pauta criterion \cite{pauta}.
     \begin{itemize}[topsep=0pt]
        \item \textbf{$y$ coordinate}
    \end{itemize}\par
    After the $x$-value of a control point is determined, the $y$-value of this point is assumed as obeying the uniform distribution with range of $\left(-\!\Delta\!D \!+ \!y_t, \Delta \!D\! + \!y_t\right)$, where $y_t$ is the $y$-value of the line $P_{i-2}P_{i-1}$ at the point $x_i$, or equal to 0 if this line doesn't exist.
    \begin{itemize}[topsep=0pt]
        \item \textbf{$z$ coordinate}
    \end{itemize}\par
\begin{figure}[htb!]
        \centering
          \includegraphics[width=0.42\textwidth]{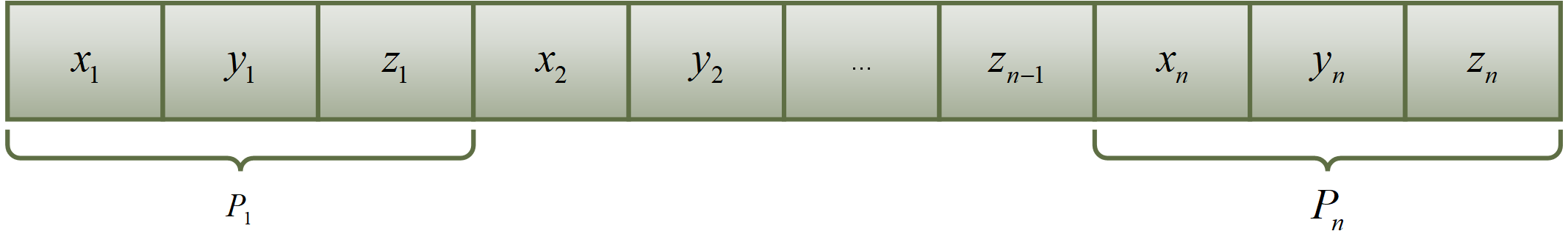}
         \caption{the genotype of individuals, where $x, y , z $ denote the rotated coordinates of control points}
          \label{genotype}
    \end{figure}

    We have hitherto initialized the $x$ and $y$ coordinates of the control points, hence we can obtain the ground constraints corresponding to the control points using $Map(x,y)$ function. Specifically, the $z$-value of $P_0$ is equal to $Map(x_0, y_0)$, the $i$th point obeying the normal distribution with mean $(z_{i-1} + Map(x_i, y_i) - Map(x_{i-1}, y_{i-1}))$ and standard deviation $\Delta h$, where $\Delta h$ is set roughly equal to $\Delta l / 3$ in this paper according to the maneuverability of the UAV.
    Finally, the individual is generated, while the encoding approach is shown in Fig.\ref{genotype}.

    \vspace{-2mm}
    \subsection{path smoother and fitness}
    \label{smooth_fitness}
\begin{figure}[htb!]
    \centering
       \begin{subfigure}[b]{0.23\textwidth}
            \centering
           \includegraphics[width=0.85\textwidth]{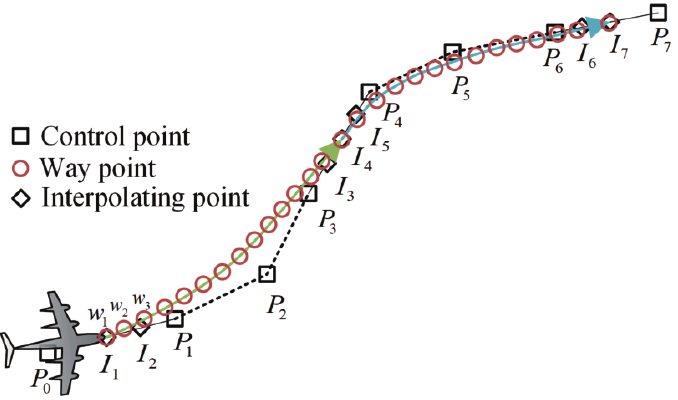}
            \caption{six-order Bezier Curve}
            \label{Bezier}
        \end{subfigure}
        \hfill
       \begin{subfigure}[b]{0.23\textwidth}
            \centering
           \includegraphics[width=0.85\textwidth]{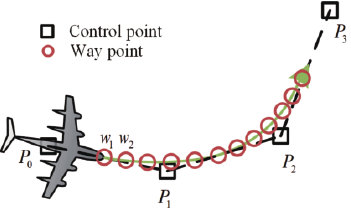}
            \caption{Cubic B-spline Curve}
            \label{B-spline}
        \end{subfigure}
  
        \begin{subfigure}[b]{0.23\textwidth}
            \centering
           \includegraphics[width=0.85\textwidth]{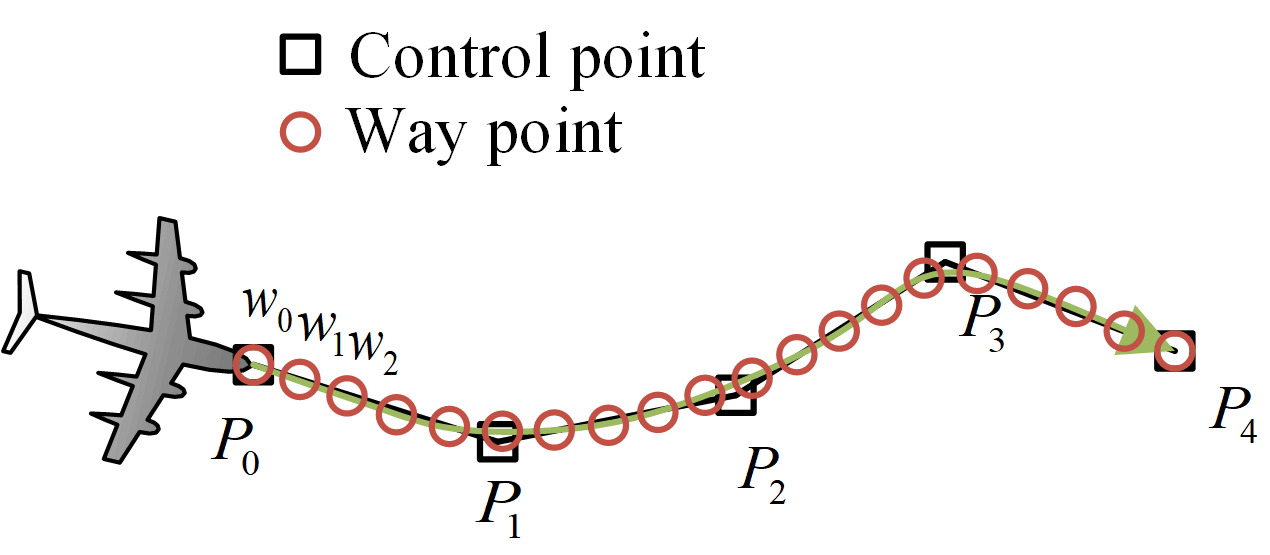}
            \caption{RTS Smoother}
            \label{RTS}
        \end{subfigure}
        \hfill
        \begin{subfigure}[b]{0.23\textwidth}
            \centering
           \includegraphics[width=0.85\textwidth]{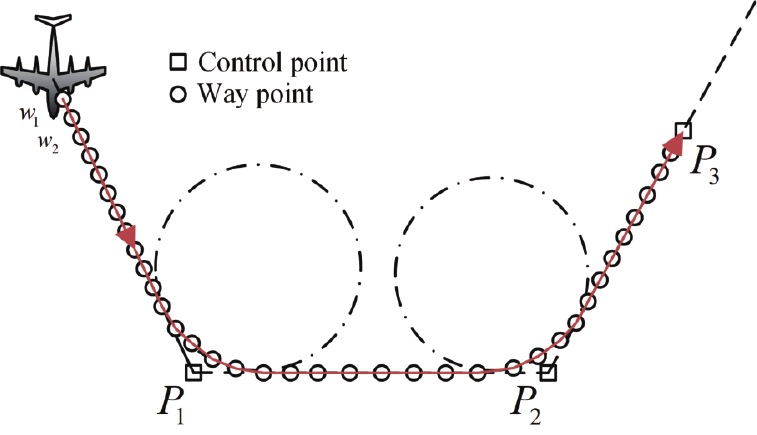}
            \caption{Tangent circle curve}
            \label{Arc}
        \end{subfigure}
    \caption{smooth methods}
    \vspace{-2mm}
    \label{smooth_curve}
    \end{figure}

    Although we have obtained the control points of a path, the concrete situation between adjacent points is still vague. To keep the $C^2$ continuity, we picked four curves respectively, i.e., Bezier curve, B-spline curve, RST smoother and tangent circle curve, as seen in Fig.\ref{smooth_curve}. Each method may have its own inherent distinctive characteristics. Concretely, Bezier curve with three control points and four interpolating points could lie within the convex hull; cubic B-spline curve will pass through the midpoint of a line segment between two adjacent points; RST smoother has the lowest error of these four methods; and Tangent circle curve, owns the minimum computational cost of requirements. Anyhow, one thing is clear, the initial control points will be transformed into the way points $\boldsymbol{w}$ using one of these methods.\par
\begin{figure}[htb!]
    \centering
       \begin{subfigure}[b]{0.22\textwidth}
            \centering
           \includegraphics[width=0.85\textwidth]{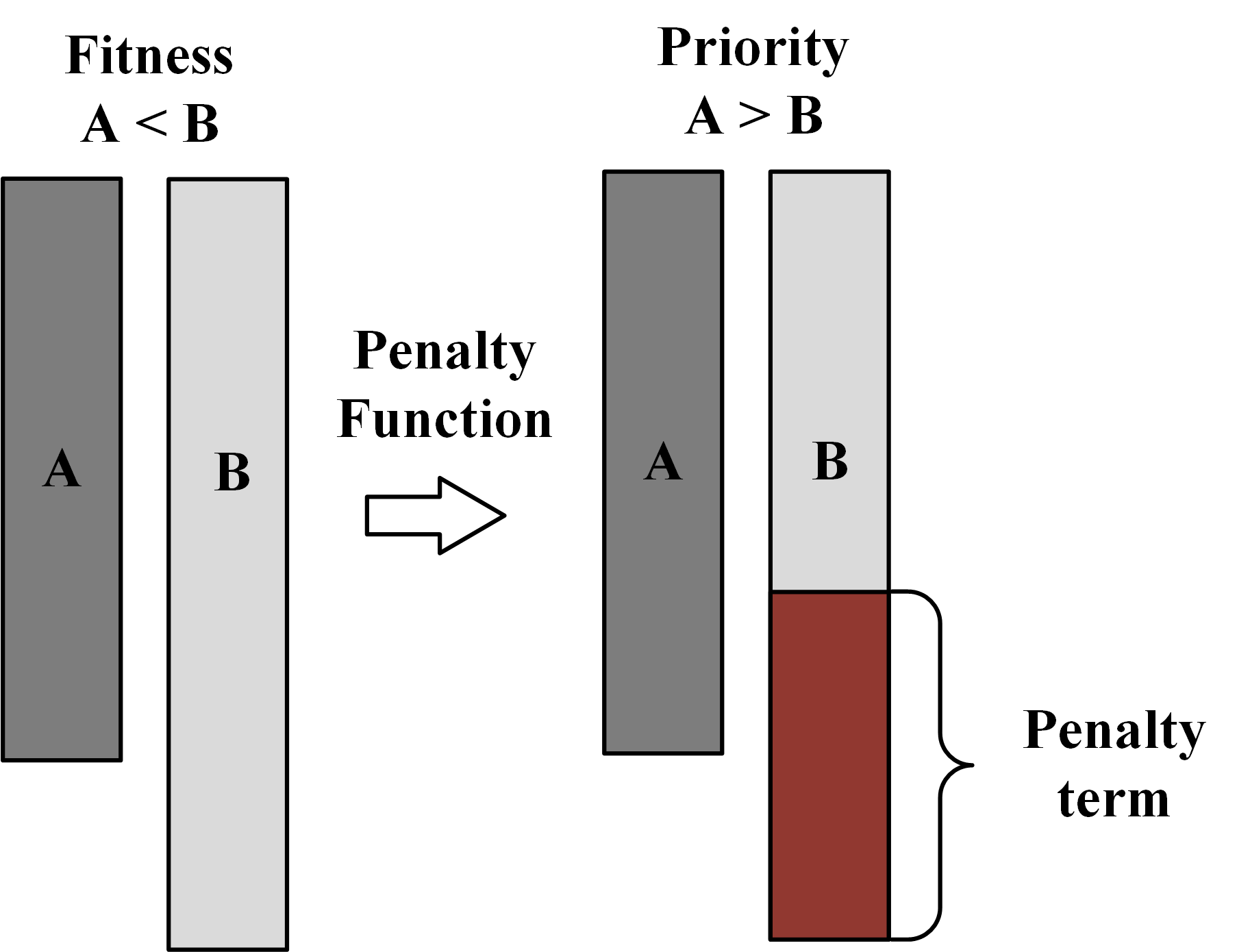}
            \caption{Penalty Function}
            \label{Penalty Function}
        \end{subfigure}
       \begin{subfigure}[b]{0.22\textwidth}
            \centering
            \includegraphics[width=0.99\textwidth]{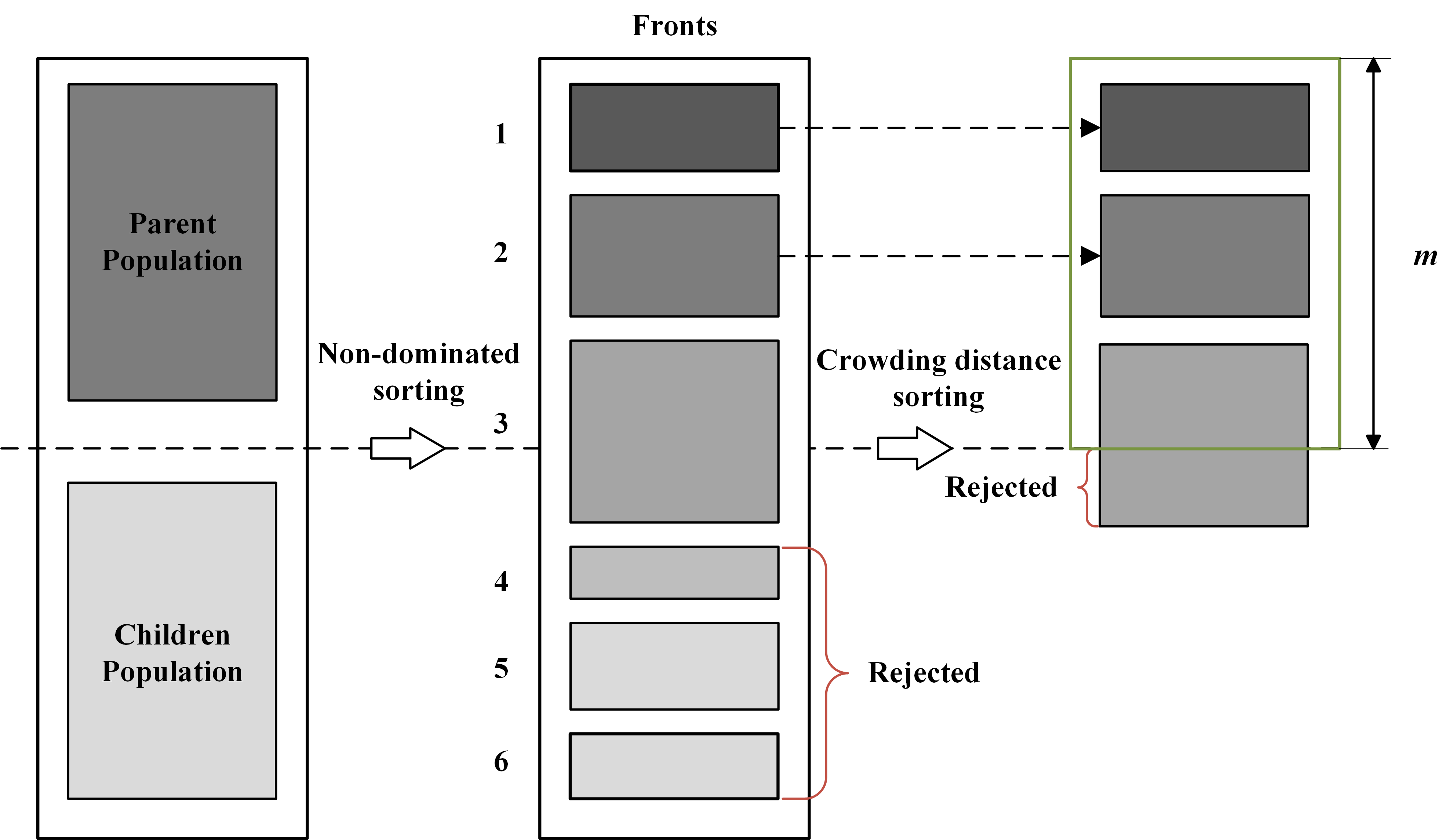}
            \caption{Non-dominated Sorting}
            \label{NS}
        \end{subfigure}
        \begin{subfigure}[b]{0.22\textwidth}
            \centering
           \includegraphics[width=0.75\textwidth]{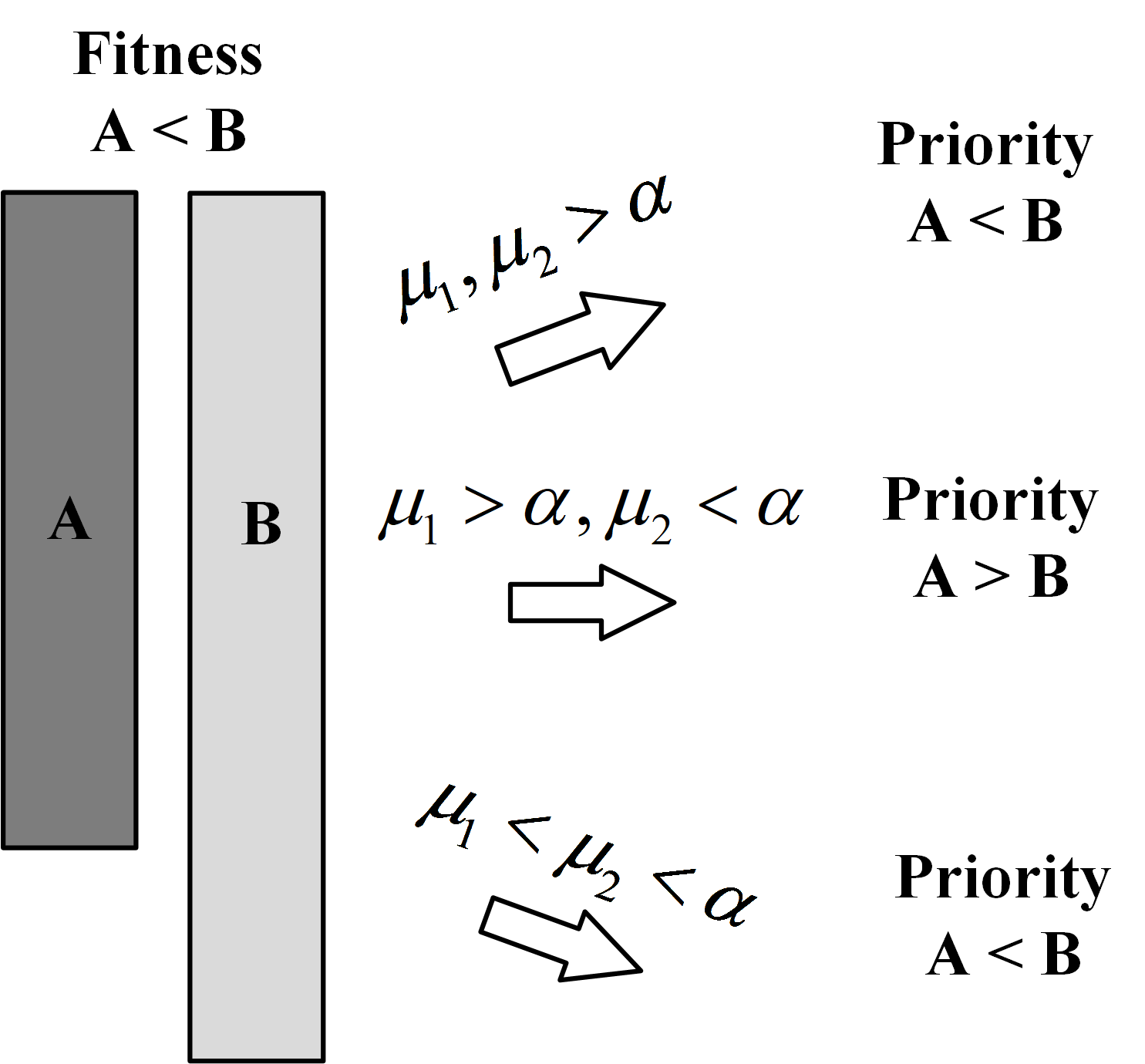}
            \caption{$\alpha$-level comparison}
            \label{alpha}
        \end{subfigure}
        \begin{subfigure}[b]{0.22\textwidth}
            \centering
           \includegraphics[width=0.75\textwidth]{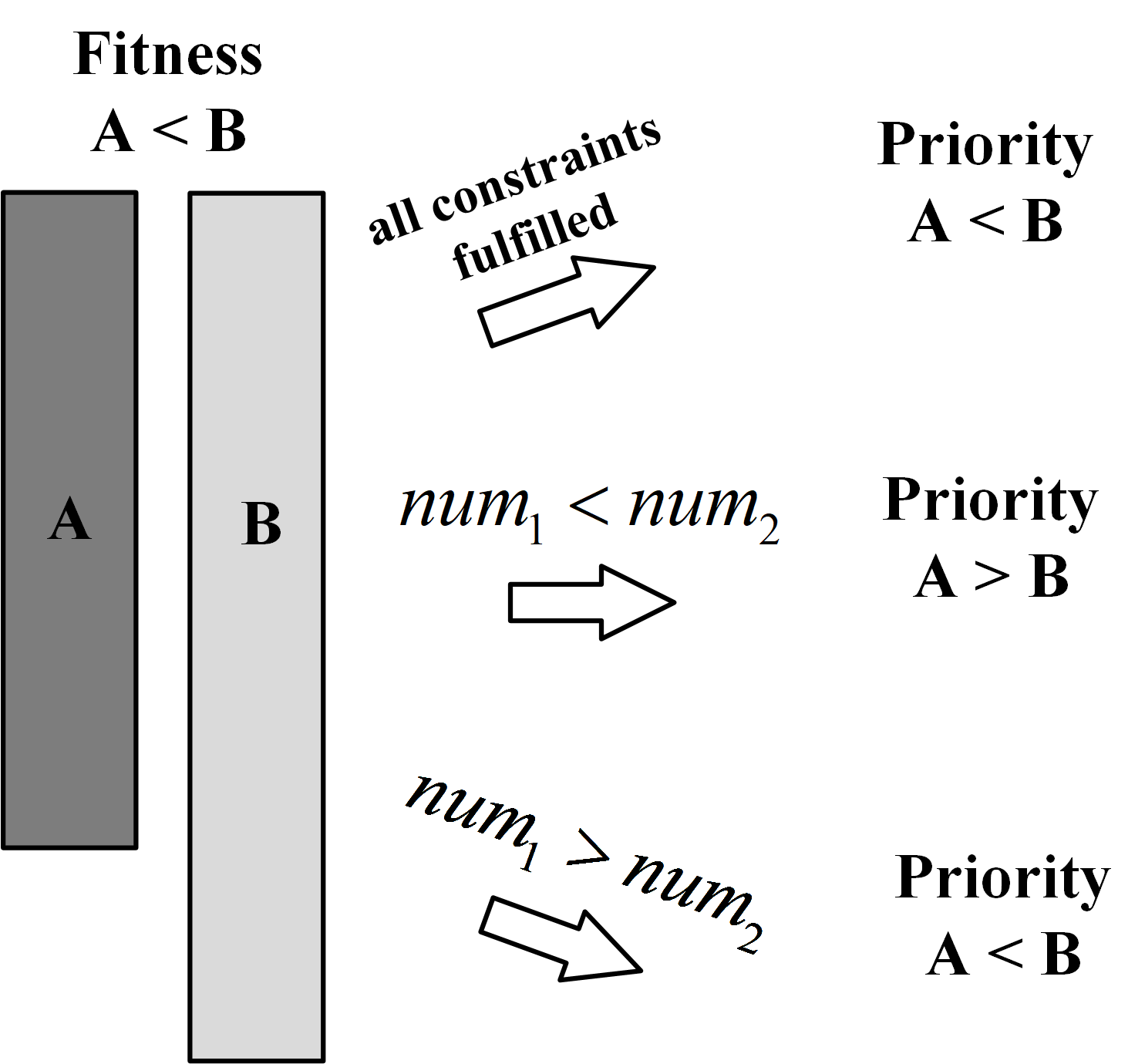}
            \caption{Number of unfulfilled constraints}
            \label{number}
        \end{subfigure}
    \caption{sorting operators}
    \label{constraint handling}
    \end{figure}

    Therefore, the fitness, which is the sole criterion for judging an individual, could be calculated according to \eqref{muti-object}. Although we have tried to increase the probability of valid solutions at the initial phase, the infeasible individuals still exist. To utilize the information of these individuals, the constraint-handling strategies named sorting operators in EP,  are introduced. As shown in Fig.\ref{constraint handling}, four different rules are utilized to handle constraints in this paper. In general, their parameters change with iteration to ensure that infeasible individuals suffer from more selection pressure at the later stage of evolution.\par
    \subsection{Selection}
    \label{selection_oper}
\begin{figure}[htb!]
    \centering
       \begin{subfigure}[b]{0.23\textwidth}
            \centering
            \includegraphics[width=0.85\textwidth]{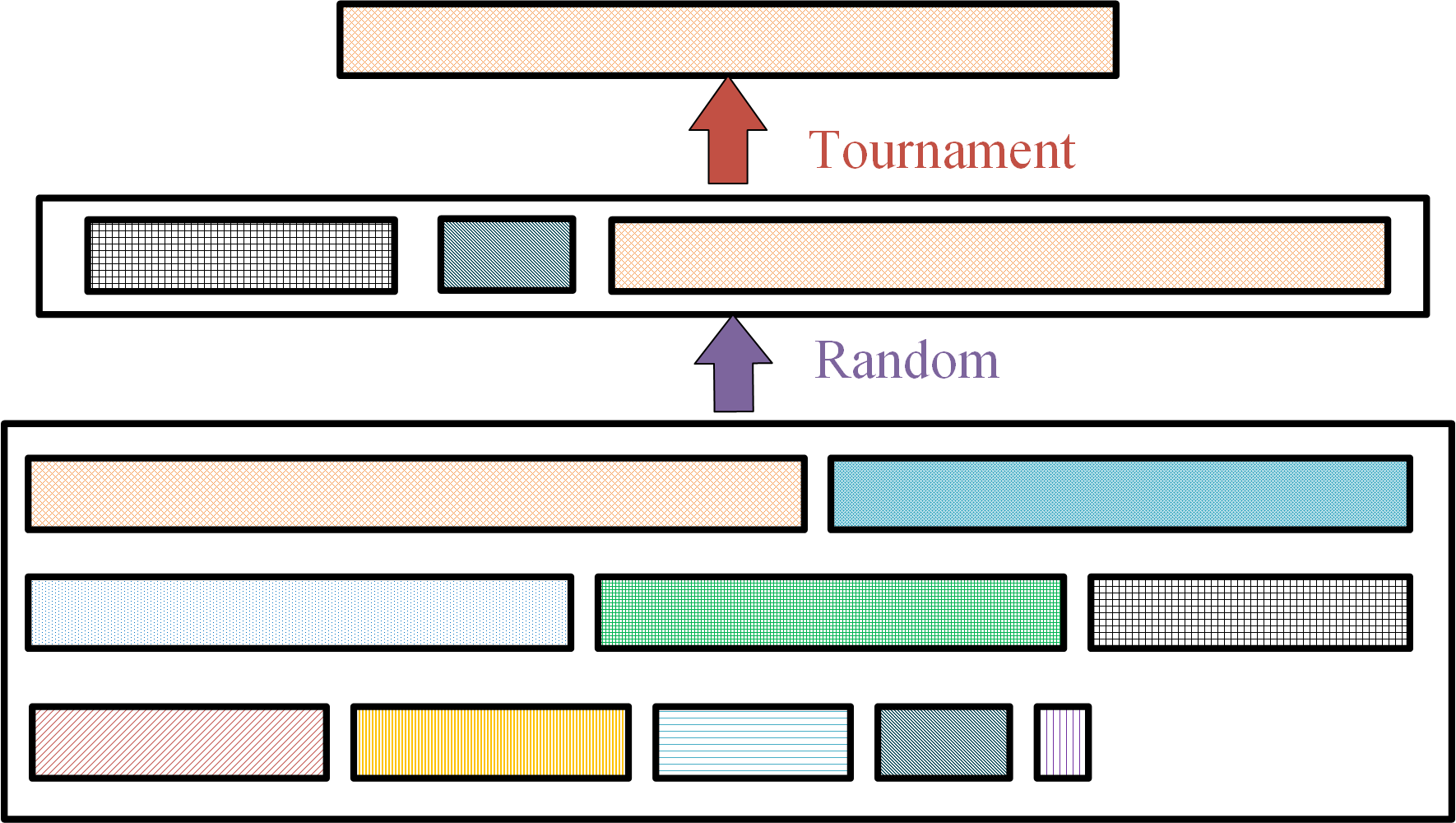}
            \caption{Tournament Selection}
            \label{Tournament}
        \end{subfigure}
        \begin{subfigure}[b]{0.23\textwidth}
            \centering
           \includegraphics[width=0.85\textwidth]{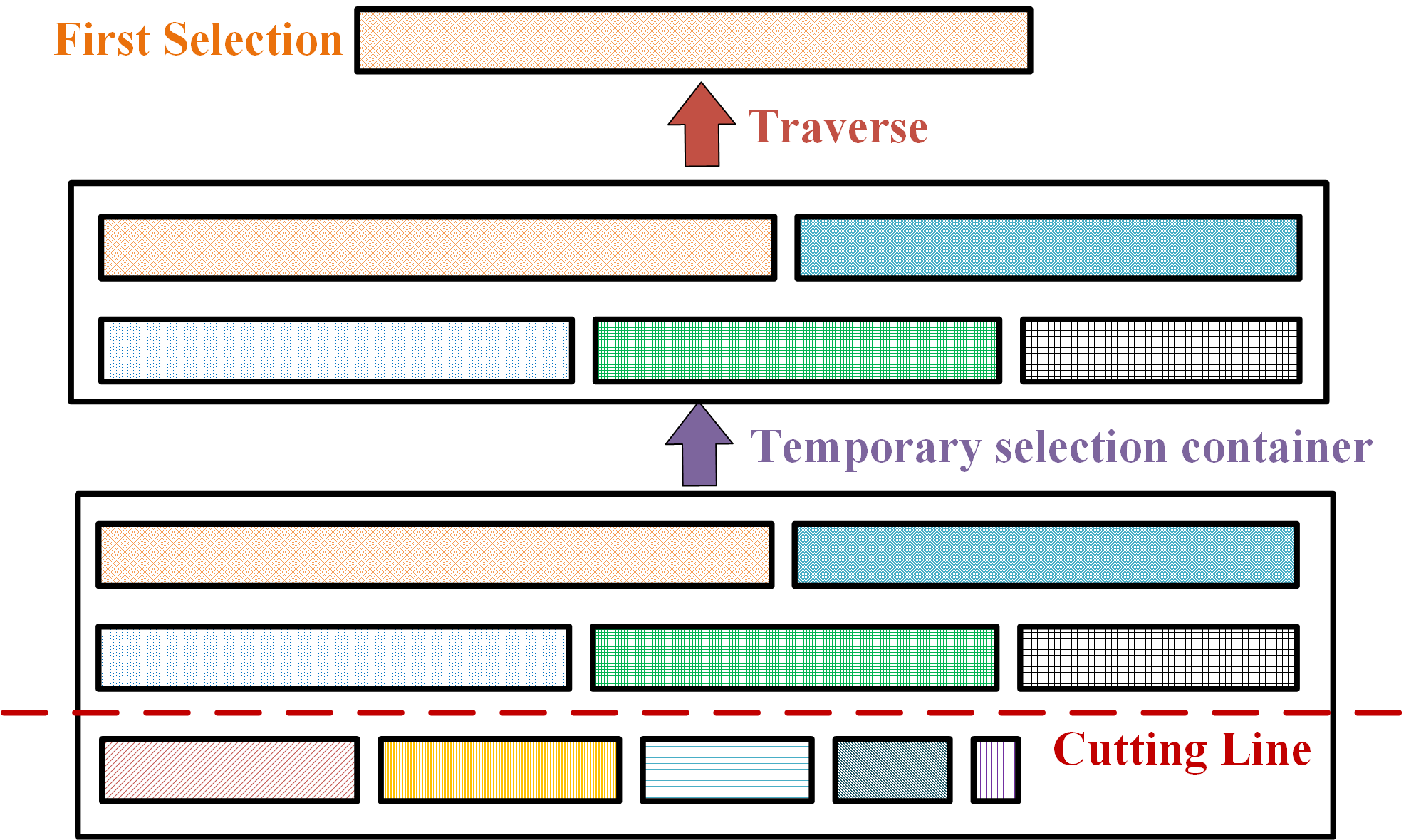}
            \caption{Truncation selection}
            \label{Truncation}
        \end{subfigure}\par
        \vspace{0.3cm}
        \begin{subfigure}[b]{0.23\textwidth}
            \centering
           \includegraphics[width=0.5\textwidth]{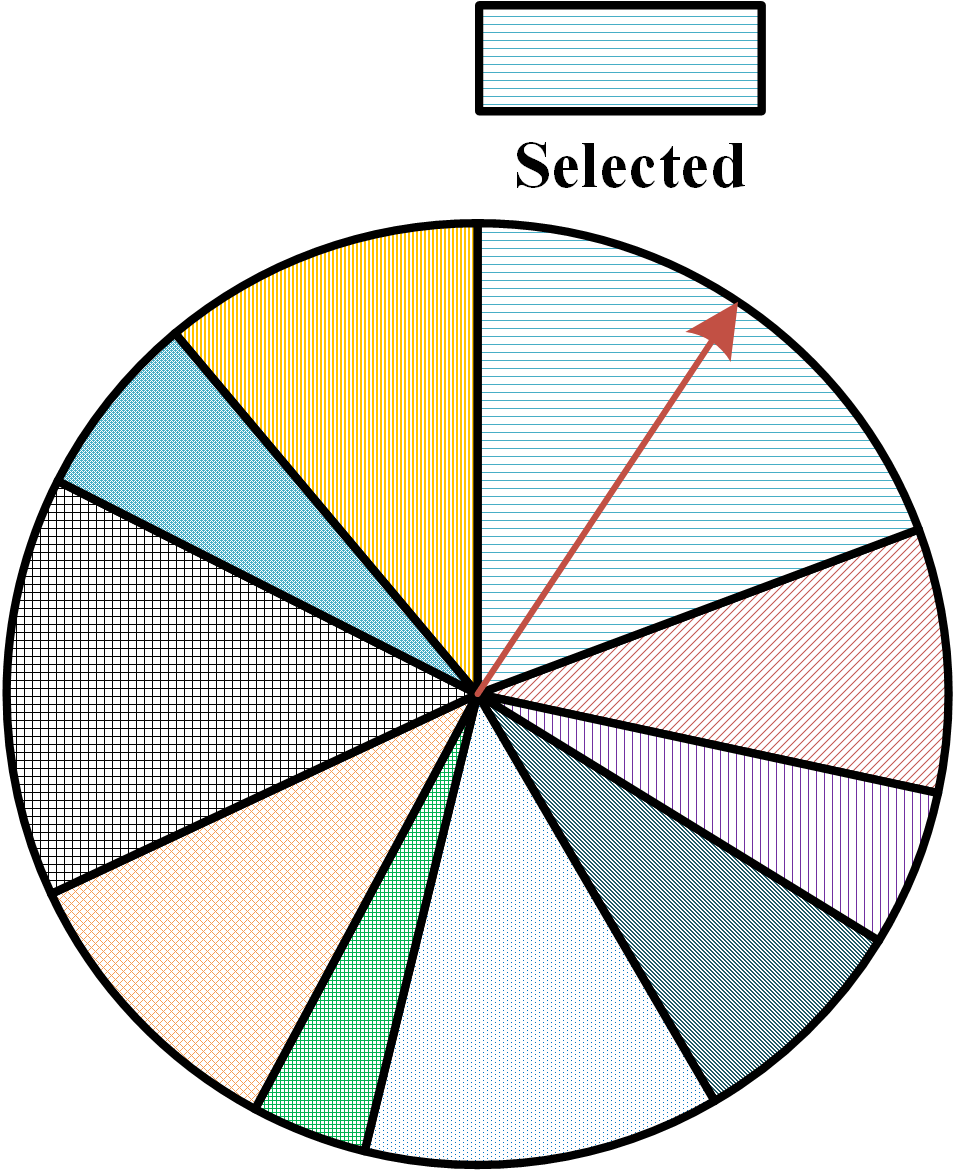}
            \caption{Roulette Wheel Selection}
            \label{RWS}
        \end{subfigure}
        \begin{subfigure}[b]{0.23\textwidth}
            \centering
           \includegraphics[width=0.85\textwidth]{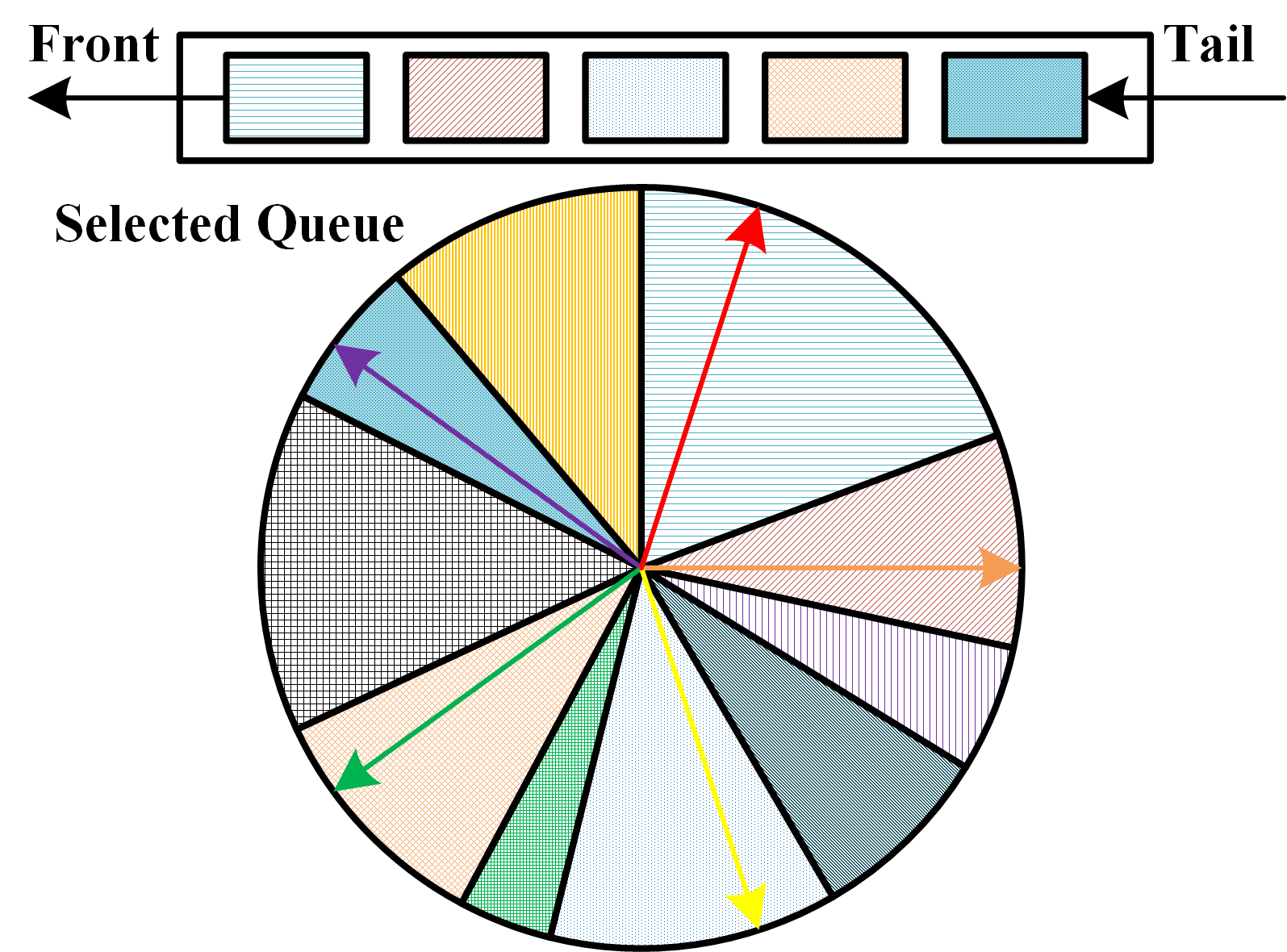}
            \caption{Stochastic Universal Sampling}
            \label{SUS}
        \end{subfigure}
    \caption{selection operators}
    \vspace{-5mm}
    \label{Selection_schemes}
    \end{figure}

    Under the EA framework, selection schemes will make a direct influence on the next generation. As shown in Fig.\ref{Selection_schemes}, these four policies will be identified as operators and added to EP's operator library. There are also various parameters could be tweaked by EP such as the number of candidates in Tournament Selection, the method of assigning areas in Roulette Wheel Selection, the setting of a cutoff line in Truncation Selection, and the step size of every pointer in Stochastic Universal Sampling, etc.
    \subsection{exploitation and exploration}
    \label{exploiation_and_exploration}
    Meta-heuristic algorithms always have a trouble of making an appropriate balance between local searching exploitation and global searching exploration. 
    In other words, these two phases are always in conflict since the computational cost in practice is limited. How to reconcile this balance is the essential problem of EA, where there is no particularly suitable solution at the moment. Although researchers have used abundant techniques (e.g. variation, hybridization, adaptation, etc.) to improve the performance of these algorithms, it's still almost impossible to propose a particular algorithm to fit any situation because this balance is vulnerable to the environment or parameters. This is what the EP is trying to solve. \par
    \subsubsection{Exploitation}
    In the term of exploitation, which focus on the local region, the search behavior should be conscientious. It indicates that the step size should be tiny and the searching information should ideally be provided by more than one individual. In this study, several operators are picked with such characteristics:
     \begin{itemize}[topsep=0pt]
        \item \textbf{Crossover}
    \end{itemize}\par
\begin{figure*}[htb!]
        \centering
       \begin{subfigure}[b]{0.3\textwidth}
            \centering
            \includegraphics[width=0.85\textwidth]{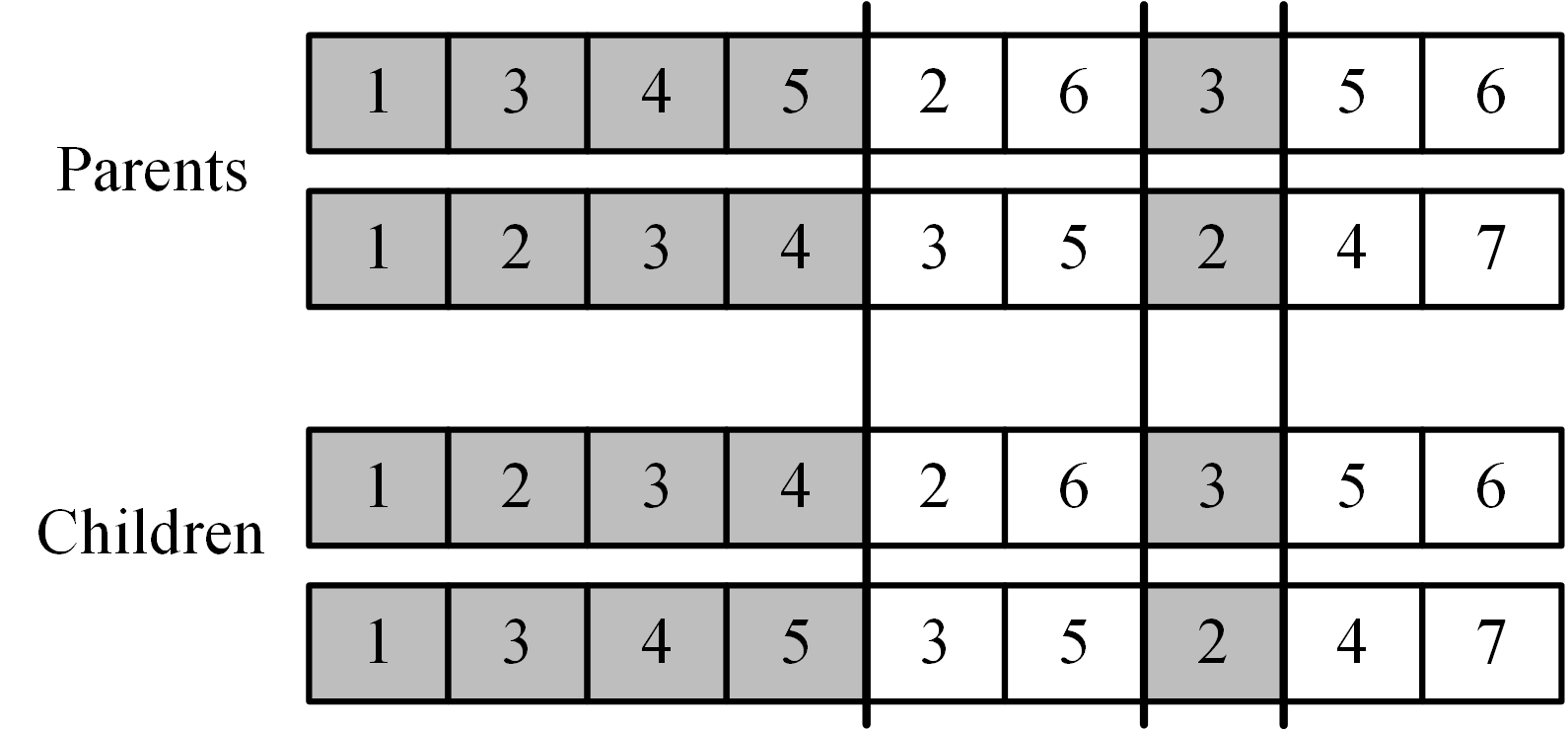}
            \vspace{2mm}
            \caption{nPX, while n = 3}
            \label{OX}
        \end{subfigure}
        \begin{subfigure}[b]{0.3\textwidth}
            \centering
           \includegraphics[width=0.85\textwidth]{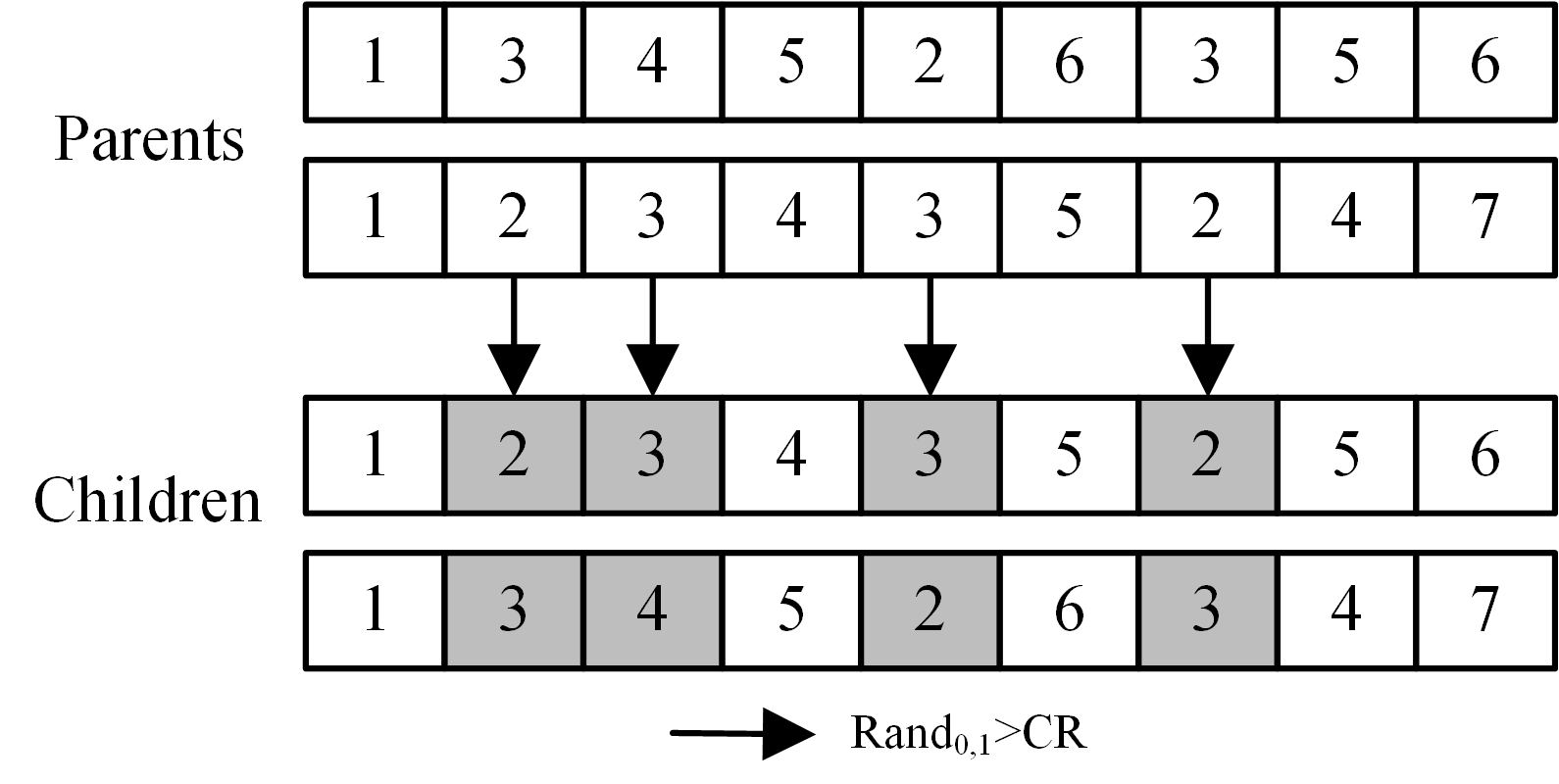}
            \caption{UX}
            \label{UX}
        \end{subfigure}
        \begin{subfigure}[b]{0.3\textwidth}
            \centering
           \includegraphics[width=1\textwidth]{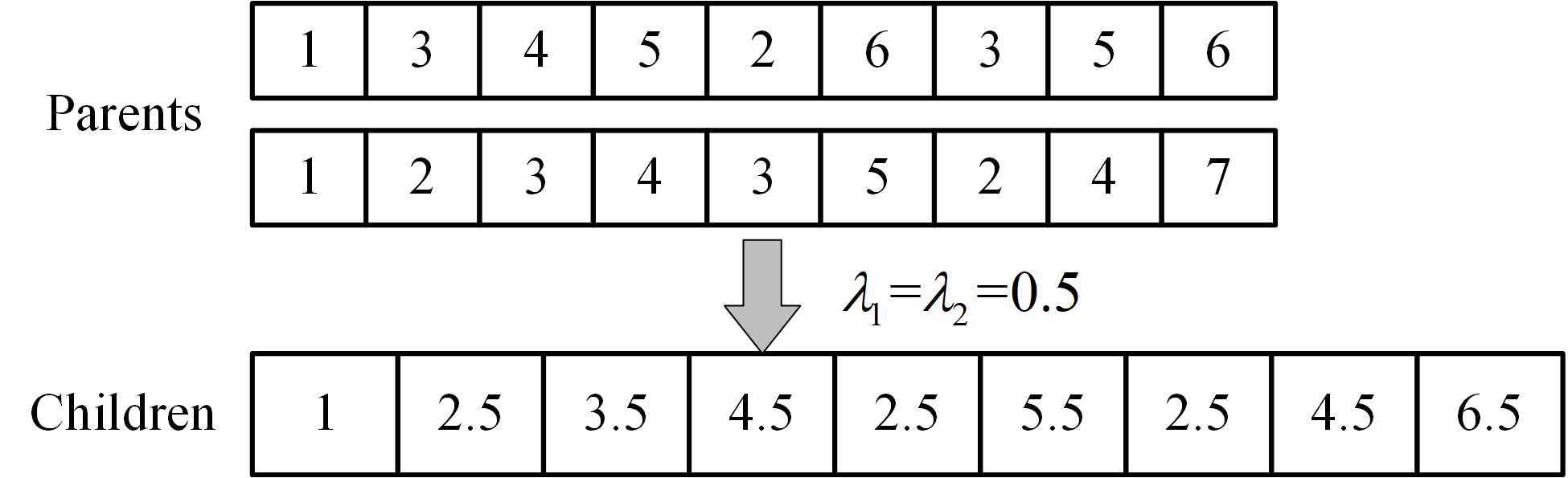}
            \vspace{2mm}
            \caption{AX}
            \label{AX}
        \end{subfigure}
    \caption{Crossover operators}
    \label{Crossover operators}
    \end{figure*}

    The Crossover, which contains numerous variants, is one of the most commonly used exploitation operators. According to the current literature of using this type of operator in the UAV path planning problem \cite{Cao2019, Karakatic2015, Yu2020,Pan2020,Yang2015}, the main variants consist of n-Point Crossover(nPX), Uniform Crossover(UX), Arithmetic Crossover(AX), which are shown in Fig.\ref{Crossover operators}. During this part, the key parameters of these three variants are the number of crossover points $n$, the Crossover Rate $CR$, and the coefficients $\left(\lambda_1, \lambda_2\right)$ ,respectively, which can be adjusted adaptively by EP.  
    \par
    
    \begin{itemize}[topsep=0pt]
        \item \textbf{PSO}
    \end{itemize}\par
   	The CIPSO\cite{Shao2020} has proposed a modified parameter adjustment in PSO, which can also be modeled as an exploitation operator due to its features of small steps and directed searching. In this modified operator, the velocity and position of a particle is updating by \eqref{PSO}.
    \begin{equation}
    \label{PSO}
    \left\{ \begin{array}{l}
        \begin{split}    
    V(t + 1) = &wV\left( t \right) + {c_1}{r_1}\left( {{P_{best}}\left( t \right) - P\left( t \right)} \right) \\ &+ {c_2}{r_2}\left( {{G_{best}}\left( t \right) - P\left( t \right)} \right)
    \end{split}
    \\
    P\left( {t + 1} \right) = P\left( t \right) + v\left( {t + 1} \right)
    \end{array}\right.
    \end{equation}
    where $p(t)$ and $v(t)$ denote the position and velocity of $t$th particle, $w$ is the inertia weight, $r_1, r_2$ are random value distributed in $(0,1)$, $c_1, c_2$ are key parameters to balance the ability of searching and convergence, which are conformed by the function \eqref{c1_c2}.
    \begin{equation}
    \label{c1_c2}
    \left\{ \begin{array}{l}
    {c_1} = {c_{\max }} - {({c_{\max }} - {c_{\min }})t}/{T}\\
    {c_2} = {c_{\min }} + {({c_{\max }} - {c_{\min }})t}/{T}
    \end{array} \right.
    \end{equation}
    Where $c_{\max}$ and $c_{\min}$ are adjusted by EP.
    \begin{itemize}[topsep=0pt]
        \item \textbf{safari}
    \end{itemize}\par
    In the modified Wolf Pack Search algorithm (mWPS)\cite{YongBo2017}, the safari process of wolves could also be encoded as an exploitation operator because of its strong following behavior and directed searching. This process can be illustrated as following:\par
    Firstly, select some of the higher ranking individuals as safari wolves.\par
    Secondly, add some tiny perturbations to these safari wolves, select the best one, denoted as $G_{\text{best}}$.\par
    Finally, the whole wolf pack will approach the best individual:
        \begin{equation}
    P\left( {t + 1} \right) = P(t) + \frac{{{G_{{\rm{best}}}} - P(t)}}{{\left| {{G_{{\rm{best}}}} - P(t)} \right|}}step
    \end{equation}
    where $step$ determines the convergence speed and premature level of the algorithm. This value and the distribution of disturbances are the key parameters in this part, which can be adjusted by EP.\par
    \begin{itemize}[topsep=0pt]
        \item \textbf{Commensalism}
    \end{itemize}\par
    As mentioned in Ref \cite{Qu2020}, the novel hybrid algorithm called HSGWO-MSOS is another efficient planner for solving UAV path planning problem. It has combined the exploration ability of the simplified Grey Wolf Optimizer (SGWO) with the exploitation ability of the modified symbiotic organisms search (MSOS), whose commensalism operator meets the requirement of this section and is described as follows: \par
\begin{equation}
      \left\{ \begin{array}{l}
    {P_i}\left( {t + 1} \right) = {P_i} + {r_{ - 1,1}}({G_{\rm{best}}} - {P_j})\\
    {P_j}\left( {t + 1} \right) = {P_j} + {r_{ - 1,1}}({G_{\rm{best}}} - {P_i})
    \end{array} \right.
\end{equation}
Where $r_{-1,1}$ denotes a random value $r \in (-1,1)$, whose distribution is determined by EP.
    \begin{itemize}[topsep=0pt]
        \item \textbf{DE}
    \end{itemize}\par
    DE is another wildly used algorithms in the area of UAV path planning for its high convergence speed and efficient result, which has developed many variants \cite{Yu2020,Yu2020a,Zhang2015,Yang2015,Chai2020}. We classified these variants into two categories: DE/rand and DE/best. Here we take the most commonly used scheme - DE/rand/1 as an example, which is described as \eqref{DE/rand/1_1}-\eqref{DE/rand/1_3}.
    \begin{equation}
    \label{DE/rand/1_1}
    {v_i}(t) = {P_{r_1}}(t) + F\left[{{P_{r_2}}(t) - {P_{r_3}}(t)} \right]
    \end{equation}
    \begin{equation}
    \label{DE/rand/1_2}
    u_{i,j}\left( t \right) =\left\{ \begin{array}{l}
	v_{i,j}\left( t \right) \ \ \ \ \text{if\ }r_{0,1}\le CR\ \text{or\ }j=rn\\
	P_{i,j}\left( t \right) \ \ \ \ \text{otherwise}\\
    \end{array} \right. 
    \end{equation}
    \begin{equation}
    \label{DE/rand/1_3}
    P_i\left( t+1 \right) =\left\{ \begin{array}{l}
	u_i\left( t \right) \ \ \ \ \text{if\ }f\left[ u_i\left( t \right) \right] \ge f\left[ P_i\left( t \right) \right]\\
	P_i\left( t \right) \ \ \ \ \text{otherwise}\\
    \end{array} \right. 
    \end{equation}
    where $v_i(t)$ and $u_{i}(t)$ denotes the donor vector and trail vector of $i$th individual in $t$th generation respectively, the subscript $j$ denotes the $j$th element in the vector, $r_1,r_2,r_3, r_n \in [1,n]$, while $r_1 \neq r_2 \neq r_3$, and $F$ is a positive parameter called scale factor, which could be adjusted by EP. \par
    \subsubsection{Exploration}
    In exploration, which concentrates more on global scope, the search behavior shows significant randomness. In this paper, various algorithms are picked as exploration operators, they are:
    \begin{itemize}[topsep=0pt]
        \item \textbf{Mutation}
    \end{itemize}\par
    The Mutation is a typical exploration operator. In recent years, it has aroused a lot of attention among researchers and thus many variants have been proposed \cite{Roberge2013, Cao2019, Karakatic2015, Razali2011, Sahingoz2014, Elhoseny2018, Roberge2018, Ghambari2019}. In this paper, we select four most representative mutation operators from relevant research as alternate operators in EP, i.e., Uniform Mutation (UM), Non-Uniform Mutation (NUM), Gaussian Mutation (GM) and Cauchy Mutation (CM). They have some of the same operational processes:\par
    Firstly, they determine whether a genome performs the mutation operation according to the mutation probability $P_m$.\par
    Secondly, the trail variable determined by the concrete operator is generated based on a particular distribution.\par
    Finally, replace the gene selected in the first step with the trail variable, the new vector is generated.\par
    \begin{itemize}[topsep=0pt]
        \item \textbf{PUS}
    \end{itemize}\par
    The population updating strategy (PUS) of hybrid PSO algorithm \cite{Shao2020} is an efficient exploration operator, which leads the algorithm to high performance with a simple method \eqref{PUS}.
    \begin{equation}
    \label{PUS}
    \left\{ \begin{array}{l}
    	P_{small}=P_{large}+ar\\
    	V_{small}=V_{large}\\
    \end{array} \right. 
    \end{equation}
    Where $P_{small}$ and $P_{large}$ represent the half individuals with smaller and larger fitness respectively, $r \in (0,1) $ is a random variable, and $a$ is the key parameter of this operator.
    \begin{itemize}[topsep=0pt]
        \item \textbf{SGWO}
    \end{itemize}\par
    The simplified gray wolf optimizer (SGWO) of HSGWO-MSOS algorithm \cite{Qu2020} is another commonly used exploration operator, which can be described as follows:\par
    Firstly, the coefficient vectors $A$ and $C$ are generated randomly according to the rule of \eqref{coefficient_of_SGWO}.
    \begin{equation}
    \label{coefficient_of_SGWO}
    \left\{ \begin{array}{l}
    	a=2-2t/T\\
    	C=2r\\
    	A=\left( 2r-1 \right) a\\
    \end{array} \right. 
    \end{equation}\par
    Secondly, calculate the value of $D_i$ with \eqref{D_of_SGWO}.
    \begin{equation}
    \label{D_of_SGWO}
    D_i=\left| CG_{\text{best}}\left( t \right) -P_i\left( t \right) \right|
    \end{equation}\par
    Finally, the new individual is produced by \eqref{genotype_of_SGWO}.
    \begin{equation}
        \label{genotype_of_SGWO}
        P_i\left( t+1 \right) =P_i\left( t \right) -AD_i
    \end{equation}
    \begin{itemize}[topsep=0pt]
        \item \textbf{CINF}
    \end{itemize}\par
    For solving the stagnation problem of DE, the collective information (CINF) was proposed in \cite{Zheng2017}, and then was used in the UAV path planning problem \cite{Pan2020}. It came up with the consecutive unsuccessful update $CUU_i$, which is calculated by \eqref{CUU}, to identify the stagnant individual.
    \begin{equation}
    \label{CUU}
    CUU_i\left( t\!+\!1 \right)\! =\! \left\{ 
    \begin{array}{ll}
	0,      & f\!\left( P_i\left( t\!+\!1 \right) \right) \! \ge\! f\!\left( P_i\left( t \right) \right)\\
	CUU_i\left( t \right)\! +\!1, & \text{otherwise}
    \end{array} \right.  
    \end{equation}
    When $CUU_i > T$, where $T$ stands for the threshold of stagnation, whose value can be adjusted by EP, the stagnation occurs and we use the following operation to deal with the stagnation.
    \begin{equation}
    \label{CINF}
    P_{i,j}\left( t+1 \right) =\left\{ \begin{array}{ll}
	P_{i,j}\left( t \right), & r\le a\ or\ j=rn \\
	P_{\text{ci\_mbest}^i,j}\left( t \right) & \text{otherwise}\\ \end{array} \right. 
    \end{equation}
    Where $P_{\text{ci\_mbest}^i,j}(t)$ is the $j$th element of collective vector $P_{\text{ci\_mbest}^i}(t)$, which is defined as follows.\par
    \begin{equation}
    \label{collective_vector}
    P_{\text{c i\_mbest}^i}\left( t \right) =\sum_{k=1}^{m_i}{w_k P_k\left( t \right)}
	\end{equation}
    Where $m_i \in [1,i]$ is a random variable, $w_k$ is a weighting factor, which is calculated by \eqref{w_k}.
    \begin{equation}
        \label{w_k}
        w_k=\frac{m_i-k+1}{1+2+...+m_i}
    \end{equation}
    \subsection{Ending Criterion and others}
    \label{ending_criterion}
    According to the literature, the execution of an algorithm may end up with several conditions, e.g., the limited time running \cite{BERGER20042037}, the limit of generations \cite{Cao2019} and the best individual has not changed for $n$-generations \cite{Yang2015} as well as the $n$\% individuals stay the same \cite{2000The}.\par
    Besides, there are some other operators can be added to EP to avoid prematurity and they fall into two categories. One is the multi-population technique, which divide the whole population to several subpopulations and exchange information around them, such as the island model \cite{Alba1999}, MAPS \cite{Yang2015a}, CEGDA \cite{Lu2005} and so on. Another is the diversity maintenance technique, such as Cellular \cite{2009MOCell}, Injection \cite{2007A}, Decay Factor \cite{Elhoseny2018}, etc. Detailed discussions of these methods are drawn in section \ref{EP}. \par
    Moreover, to speed up the convergence, several operators are proposed in literatures, such as Elitism to select the top $n$\% individual as survival, Repair algorithm \cite{CHOI2003773} to ensure all individuals remain feasible with greedy search after exploration and exploitation, and push-forward insertion heuristics (PFIH) \cite{6253010} to optimize the solutions locally, etc. \par
\section{The Design of Evolutionary Programmer}
    \label{EP}
    EP is a machine learning framework that can automatically create suitable UAV path planning program based on EA. Since the outstanding performance of GA in automatic programming, we choose it as the solver of EP. In this section, we will give a comprehensive overview of EP's software architecture. \par
\begin{figure}[htb!]
        \centering
        \includegraphics[width=0.45\textwidth]{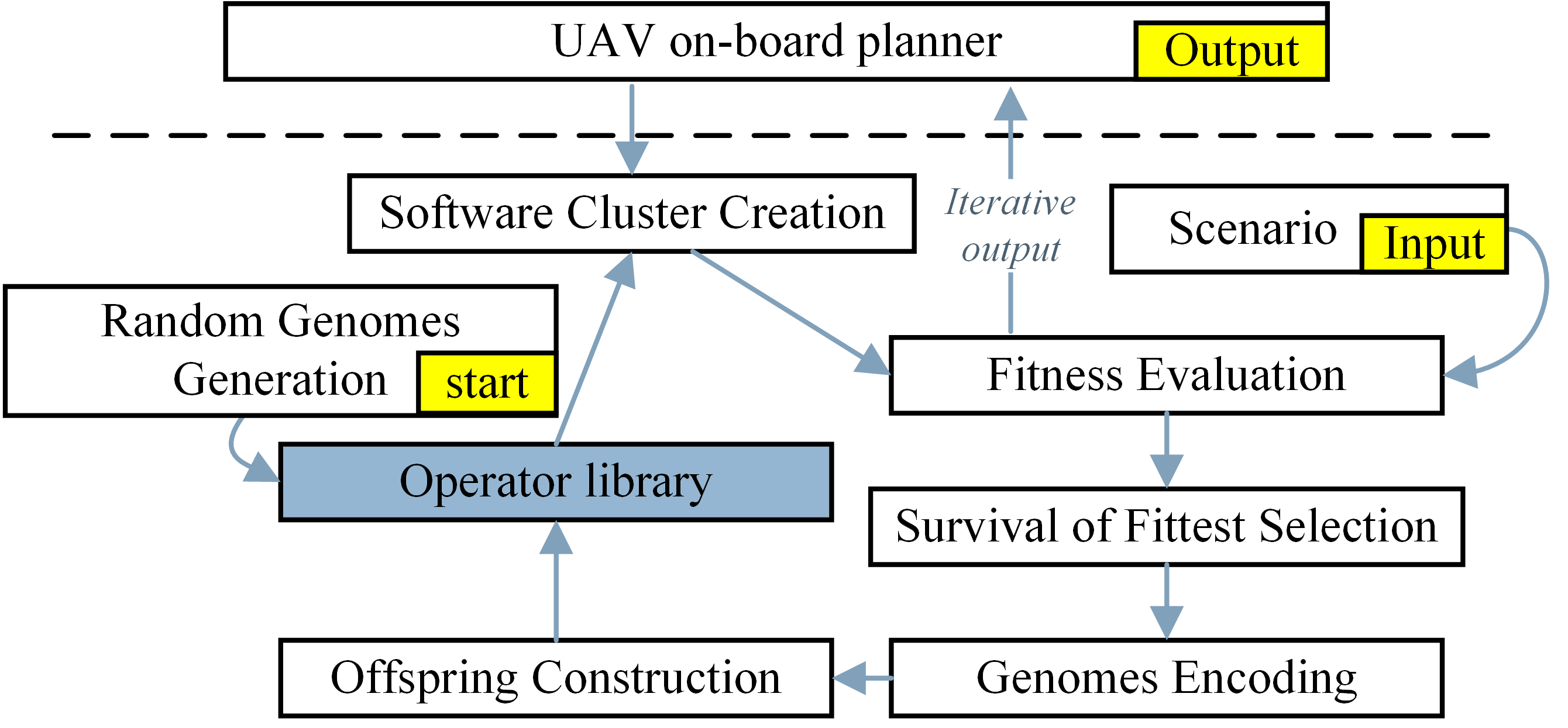}
        \caption{The Evolutionary Programmer software architecture}
        \vspace{-5mm}
        \label{EP_struct}
    \end{figure}

    As shown in Fig.\ref{EP_struct}, when the UAV enters a new environment, the algorithm activates. First, it generates a series of binary arrays, which are mapped to path planners. These planners, together with the UAV on-board planner, formed the initial software cluster. Second, each planner is run based on the input scenario, then EP will record the cost and running time of each planner, and combine them into the fitness. Next, EP selects the planners with high fitness for cross-mutation of their genes and recombines them into a new software cluster. Finally, when the iteration satisfied the ending criterion, the UAV on-board planner is updated to be the individual with the highest fitness in the current software cluster. The detailed description of this process is discussed in the following subsections. 
    \subsection{Operator Library and the Genomes}
    \label{sub_genomes}
   A genome is a sequence of genes used to represent instructions. To translate genomes into programs, the first thing is building an operator library and mapping its components onto the genes. As a result, each genome in EP consists of 64-bit sequence with a value of 0 or 1. The decoder will covert a genome into the related program according to the criterion established at the time of encoding. This section will describe the criterion in detail.
    \begin{itemize}[topsep=0pt]
        \item \textbf{Initialization}
    \end{itemize}\par
    In the term of initialization, 10-bits genes are encoded to represent some specific operators, as seen in Table \ref{Initialization_genome}. Concretely, there are four candidates of $Di$, i.e., default, Island Model, MAPS, CEGDA (see \ref{ending_criterion} in detail), and four options of $Cv$ as smoother mentioned in \ref{smooth_fitness}, i.e., Bezier, B-spline, RTS, Tangent circle.
     \begin{table}[htbp]  
        \centering
        \caption{11-bits genes for Initialization}   
        \resizebox{0.48\textwidth}{!}{
        \begin{tabular}{cccc}
        \toprule
        Token & Summary                      & Coding       & Order \\
        \midrule
    $\#CP$ & the number of Control points                & {[}00, 11{]}   & 0     \\
    $\#P$  & the number of populations                   & {[}00, 11{]}   & 2     \\
    $\#I$  & the number of individuals in a population   & {[}000, 111{]} & 5     \\
    $Di$   & the methods of dividing populations         & {[}00, 11{]}   & 7     \\
    $Cv$   & the Curve used to smooth the Control points & {[}00, 11{]}   & 9     \\  
    \bottomrule 
    \end{tabular}}
    \label{Initialization_genome}
    \end{table}
    
    \begin{itemize}[topsep=0pt]
        \item \textbf{Sorting and Selection}
    \end{itemize}\par
    As shown in Table \ref{sorting_genome}, $\textit{So}$ implies sorting operators, which consist of Penalty Function, Non-dominated Sorting, $\alpha$-level comparison and the Number of unfulfilled constraints; $\textit{Se}$, which denotes selection operators, consists of Tournament selection, Truncation selection, Roulette Wheel Selection and Stochastic Universal Sampling; $Rank$ assigns scores as selection criterion to the sorted individuals, these scores could be linear, logarithmic, exponential, etc. See Section.\ref{smooth_fitness} and \ref{selection_oper} for a detailed explanation of these operators.
     \begin{table}[htbp]  
        \centering
        \caption{13-bits genes for sorting and selection}   
        \resizebox{0.48\textwidth}{!}{
        \begin{tabular}{cccc}
        \toprule
        Token & Summary                      & Coding       & Order \\
        \midrule
    $\textit{So}$  &  the measure method for sorting            & {[}00, 11{]}   & 11     \\
    $\textit{So}_{param}$  & the parameters of the measure method  & {[}000, 111{]}   & 13    \\
    $Elitism$  & select top $n\%$ as survival  & {[}00, 11{]}     & 16   \\
    $Rank$   & the ranking system to evaluate the solutions  & {[}00, 11{]}   & 18     \\
    $\textit{Se}$   & the selection method used to pick a individual & {[}00, 11{]}   & 20    \\ 
    $\textit{Se}_{param}$   & the  parameters of selection method & {[}00, 11{]}   & 22  \\
    \bottomrule 
    \end{tabular}}
    \label{sorting_genome}
     \end{table}
    \begin{itemize}[topsep=0pt]
        \item \textbf{exploitation and exloration}
    \end{itemize}\par
    As illustrated in \ref{exploiation_and_exploration}, the genes of this part are depicted as Table \ref{ex_genome}, where $\textit{Twins}$ and $\textit{Infer}$ operate on the offspring produced by exploitation and exploration, respectively.
     \begin{table}[htbp]  
        \centering
        \caption{12-bits genes for exploitation and exploration} 
        \setlength\tabcolsep{0pt}
        \renewcommand{\arraystretch}{1.1}
        \resizebox{0.48\textwidth}{!}{ 
        \begin{tabular}{cccc}
        \toprule
        Token & Summary                      & Coding       & Order \\
        \midrule
    $Exploit$  &  the algorithm chosen as exploitation operator   & {[}000, 111{]}   & 24     \\
    $Exploit_{param}$  & the parameters of $Exploit$   & {[}00, 11{]}   & 27    \\
    $Twins$   & if allow two children (if exist) to next  & {[}0, 1{]}   & 29    \\
    $Explore$   & the algorithm chosen as exploration operator  & {[}000, 111{]}   & 30    \\
    $Explore_{param}$   & the parameters of $Explore$  & {[}00, 11{]}   & 33 \\
    $Infer$   & if reserve the inferior generated by exploration operator  & {[}0, 1{]}   & 35 \\
    \bottomrule 
    \end{tabular}}
    \label{ex_genome}
     \end{table}
    \begin{itemize}[topsep=0pt]
        \item \textbf{Ending Criterion}
    \end{itemize}\par
    The Ending Criterion is a crucial option to avoid additional calculation. Generally, we end up a EA with fixed generation or running time, which is encoded in the token $End$ and $End_{param}$. But there exist some cases denoting that even we keep executing the algorithm, the progress we make will be quite limited. For example, when the prematurity happens, which means that the algorithm is trapped into local minimum, we could stop this population to avoid wasting computational resources. The definition of prematurity and related parameters are encoded in $\textit{Case}1$ and $\textit{Case}1_\textit{param}$, consisting of default setting (do nothing in this term), evolutionary stagnation, population homogenization and goal achievement. Similarly, in $\textit{Case2}$ which is designed for multi-population situations, when the current population is too similar to a solved population, the options are default, reset, killing, and adjustment.
    \begin{table}[htbp]  
        \centering
        \caption{13-bits genes for Ending Criterion} 
        \renewcommand{\arraystretch}{1.1} 
        \setlength\tabcolsep{0pt} 
        \resizebox{0.48\textwidth}{!}{ 
        \begin{tabular}{cccc}
        \toprule
        Token & Summary                      & Coding       & Order \\
        \midrule
    $End$  &  the method used as ending criterion   & {[}0, 1{]}   & 36     \\
    $End_{param}$  & the parameters of  $End$  & {[}000, 111{]}   & 37    \\
    $Case1$   & the definition of prematurity  & {[}00, 11{]}   & 40     \\
    $Case1_{param}$   & the parameters of $Case1$ &  {[}00, 11{]}  & 42   \\
    $Case2$   & the operation if population is similar to a solved one  & {[}00, 11{]}   & 44 \\
    $Case2_{param}$   & the parameters of $Case2$  & {[}00, 11{]}   & 46 \\
    $Case3$   & if restart the algorithm when it stops early  & {[}0, 1{]}   & 48 \\
    \bottomrule 
    \end{tabular}}
    \label{ending_genome}
\end{table}
 \begin{itemize}[topsep=0pt]
        \item \textbf{other operators}
    \end{itemize}\par
    Besides the previously proposed operations, there are other non-standard operators to improving the performance of algorithm and speed up the convergence. An presented in Table \ref{others_genome}, the $Cell$, $Injc$, $Anti$, $Fbcl$ and $Decy$, which have been explained in the section \ref{ending_criterion}, are used to maintain the diversity of the population. And the rest of operators are used mainly for improving the rate of convergence.
\begin{table}[htbp]  
        \centering
        \caption{15-bits genes for other operators}   
        \renewcommand{\arraystretch}{1.1} 
        \setlength\tabcolsep{0pt} 
        \resizebox{0.48\textwidth}{!}{ 
        \begin{tabular}{cccc}
        \toprule
     Token & Summary                      & Coding       & Order \\
        \midrule
    $Cell$  &  Using Cellular & {[}0, 1{]}   & 49    \\
    $Injc$  & Using Injection method  & {[}00, 11{]}   & 50   \\
    $Rpar$   & Using Repair algorithm to enforce the feasibility  & {[}00, 11{]}   & 52     \\
    $Mgrt$   & Using the migration to keep the info sharing among populations  & {[}00, 11{]} &  54  \\
    $Anti$   & Using Antibody to prevent the domination of one genotype & {[}00, 11{]}   & 56 \\
    $Fbcl$   & Forbidding the clones in the population   &  {[}0, 1{]}  & 58 \\
    $Decy$   & using the Decay Factor Algorithm  & {[}00, 11{]}   & 59 \\
    $PFIH$   & using the PFIH to optimize the solution  & {[}000, 111{]}   & 61 \\
    \bottomrule 
    \end{tabular}}
    \label{others_genome}
\end{table}
  \subsection{Fitness}
  \label{sub_fitness}
  The fitness of EP, which is different from the description in section \ref{smooth_fitness} and used to evaluate how well a created program performs, should be well designed from the user-defined target. In this paper, the efficiency and performance are what we are concerning about. For the former, we use a time variable $F_t$ to represent it. And for the latter, we pick $F$ which is calculated in \eqref{muti-object} to denote it. So the fitness function of EP could be formulated as \eqref{F_EP}.
  \begin{equation}
    \label{F_EP}
      F_{EP} =1/\left( W_1 F + W_2 F_t/E_t \right)
  \end{equation}
  Where $E_t$ is the expectation of the running time that can be adjusted by user, $W_1$ and $W_2$ are weights that sum to 1.
\subsection{Generations}
\label{sub_Generations}
A generation in GA consists of several operators. The better and faster the generated program produces the result, the higher its fitness is, and thus the more likely it is to be selected and maintain to the next generation at each epoch. The process in a generation is carried out as follows.
    \begin{itemize}[topsep=0pt]
        \item \textbf{Selection}
    \end{itemize}\par
    In every epoch, there are a fixed number of genomes as the pool of selection. For the sake of rapid convergence, here we choose the RWS (section \ref{selection_oper}) as the selection operator..
    \begin{itemize}[topsep=0pt]
        \item \textbf{Crossover and Mutation}
    \end{itemize}\par
    In each epoch, a bunch of offspring are generated via crossover and mutation. To be specific, a process called \textit{crossover} implies that parents contribute a part of genome to the child, and the \textit{mutation} is that adding the stochastic disturbance under some controlled rules to specific genes of the child.
    \begin{itemize}[topsep=0pt]
        \item \textbf{Survival of the Fittest}
    \end{itemize}\par
    After the above process, a ranking system is introduced for the purpose of sorting all the genomes relying on their fitness at the earlier searching stage and removing the infeasible genomes whose running time exceeded the $E_t$ at the later stage. Concretely, the worst planner got the lowest score (about 0.8 in this paper), the best one got a score of 1.2 and all others were rank in between. This ranking system prevents better planners from being disproportionately favored in the phase of selection.
    
\subsection{Terminate Conditions and Output}
\label{Output}
The entire evolutionary process has to end with some limited elapsed time. In this paper, we assume that the UAV goes through a phase of calibrating and burning the planner before entering a new scene, that is, EP is given enough time to work. Due to the fast convergence of GA, we limit the running time of EP to a maximum of 20 minutes. And as the process of evolution stopped, the planner with the highest fitness is output and burn onto the UAV's onboard computer.
\section{Simulation Results}
\label{simulation}
In theory, by evolving the operators of EA, EP will get better performance than the origin algorithms under certain circumstances, and, in different environments, EP should have evolved to show better adaptability. To verify these features that do exist, 1) we designed different scenarios for proving the adaptive ability of EP, and 2) in some representative scenarios, we compared EP with the state-of-the-art evolutionary methods, which were decomposed and formed the operator library in this paper, for testifying the performance of EP.  
\subsection{adaptive verification}
\label{adaptive simulation}
\begin{figure}[htb!]
\centering
   \begin{subfigure}[b]{0.25\textwidth}
        \centering  
       \includegraphics[width=1\textwidth]{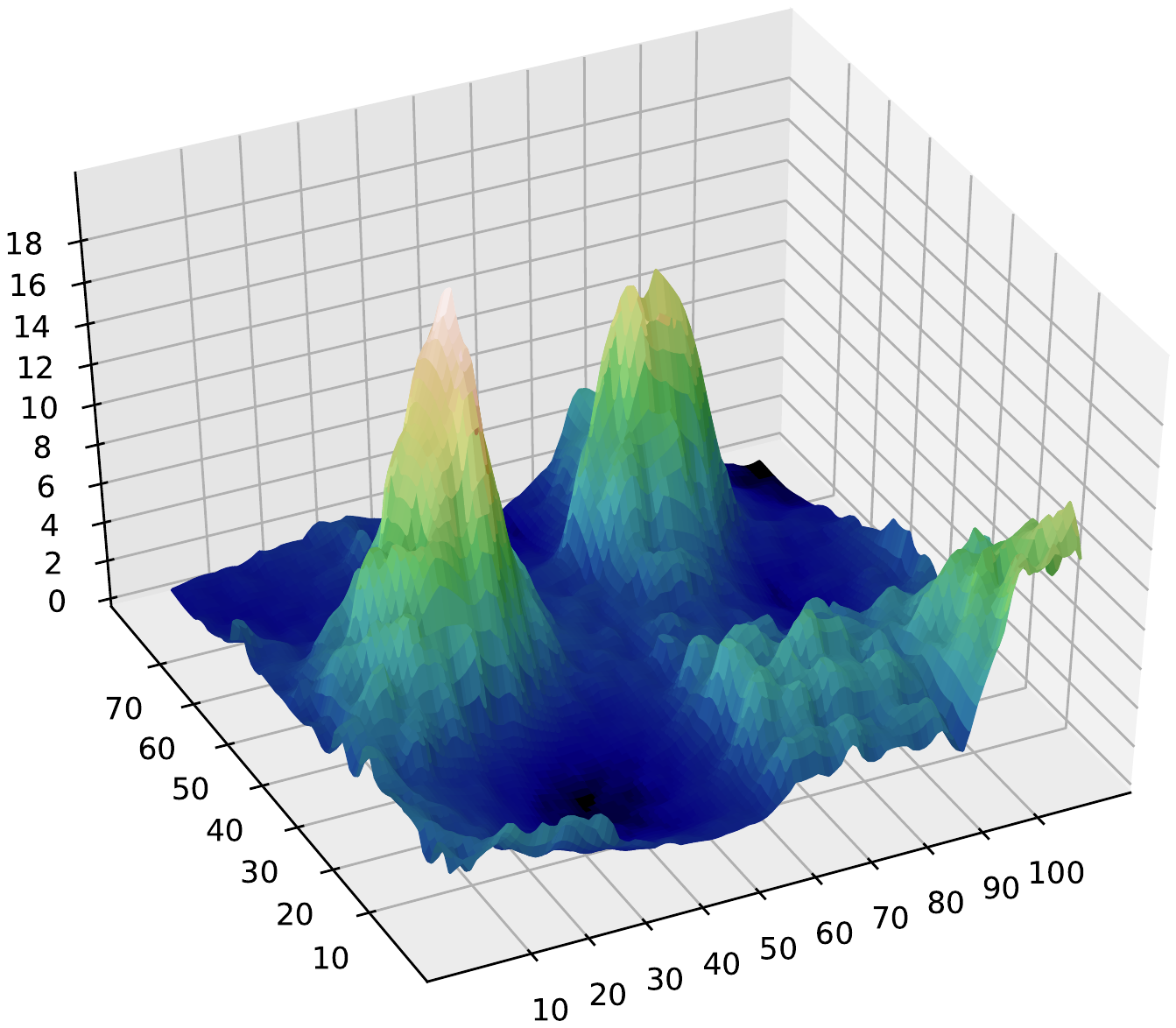}
        \caption{}
        \label{baseline.a}
    \end{subfigure}
    \hfill
   \begin{subfigure}[b]{0.23\textwidth}
        \centering
       \includegraphics[width=0.95\textwidth]{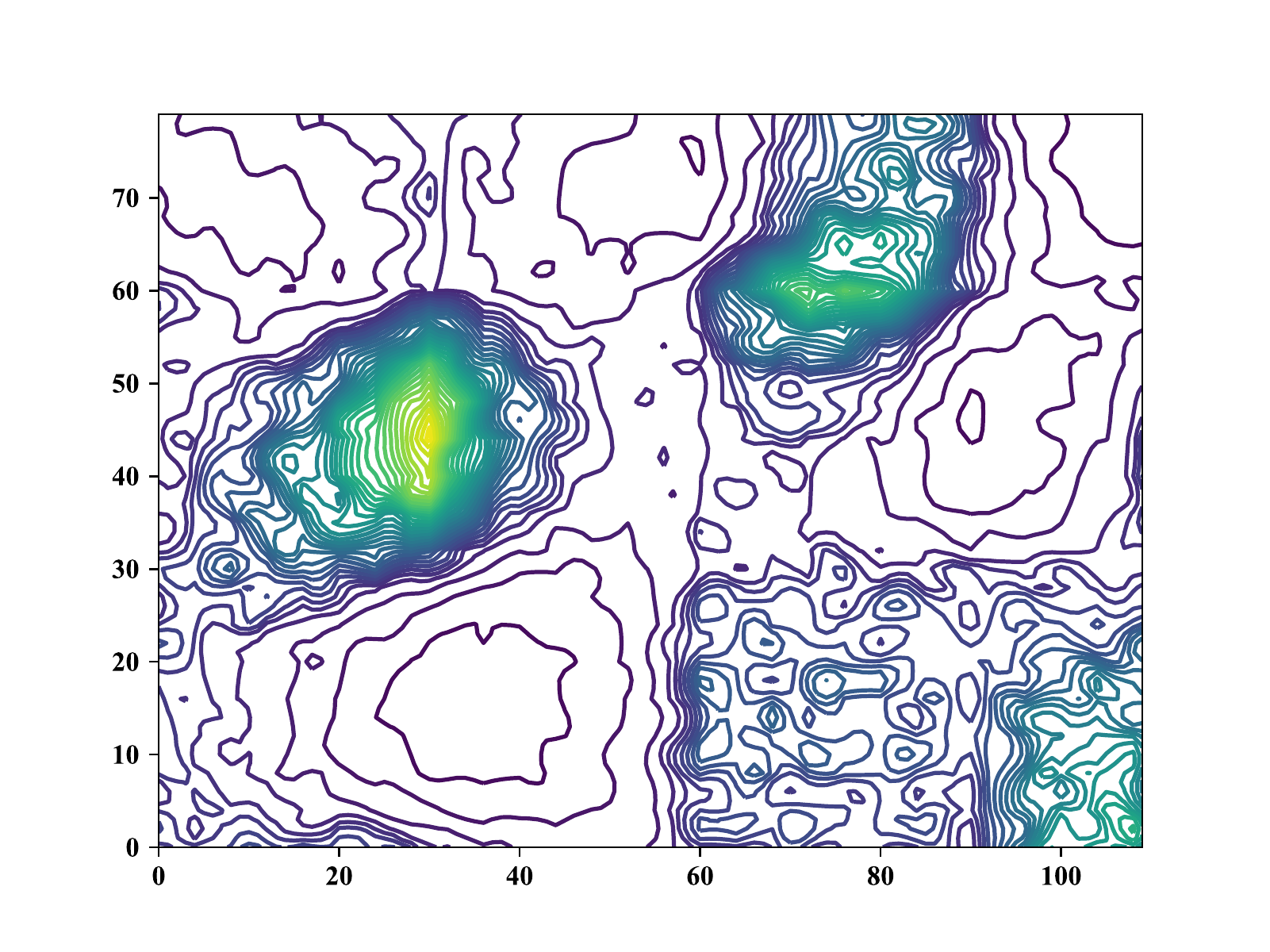}
        \caption{}
        \label{baseline.b}
    \end{subfigure}
\caption{
\centering
base scenario
}
\label{baseline}
\vspace{-5mm}
\end{figure}

In this section, plenty of scenarios were generated to test the adaptive ability of EP to the changes of topographic relief, obstacle density and starting\&ending points. In each of the experimental groups, an identical terrain, as shown in Fig.\ref{baseline}, was used as the base terrain, so that we can compare which factor changes EP is more sensitive to. Moreover, we set up an initial planner (genetic algorithm) as input to the EP, labeled Origin Planner. Uniformly we set the mission space as $\left[ 0,\ 150 \right] \times \left[ 0,\ 100 \right] $ and  in the first two experiments, we set the starting point as $\left[ 0,\ 0,\ Map\left( 0,\ 0 \right) \right] 
$, the destination as $\left[ 100,\ 70,\ Map\left( 100,\ 70 \right) \right] 
$. In each scenario,  we give EP 20 minutes for evolution, then we measure the performance between the Evolved Planner and the Original Planner through the constraint and fitness of their solution as well as the execution time they used.

\begin{enumerate}[labelsep = .5em, leftmargin = 0pt, itemindent = 3 em]
    \item Obstacle density\par
    In most references, obstacle density is one of the most influential environmental factors affecting UAV path planning performance. For the purpose of testifying adaptive capacity to the environment of EP, four scenarios with different obstacle density in 2D view\footnote{In order to display performance effects intuitively, this paper only presents 2D views of scenarios, and all data and 3D views can be accessed in \url{https://github.com/loujiabin1994/Evolutionary_Programmer/tree/numpy/script/data}.} are shown in Fig.\ref{2d_obstacle_density}(a)-(d), whose Obstacle density could be characterized as sparse, not too much, a bit more and dense, respectively. According to the constraints, the UAV should be forbidden from entering rectangular areas, and it is better to keep away from circular areas to reduce the probability of detection and attack. In each scenario, the evolved planner can always find the smooth path to avoid all the obstacles. In contrast, the original planner can only avoid obstacles in simple scenarios, but cannot adapt to the requirements of complex scenarios. \par
    
    \begin{figure}[htb! ]
    \centering
   \begin{subfigure}[b]{0.23\textwidth}
        \centering  
       \includegraphics[width=0.95\textwidth]{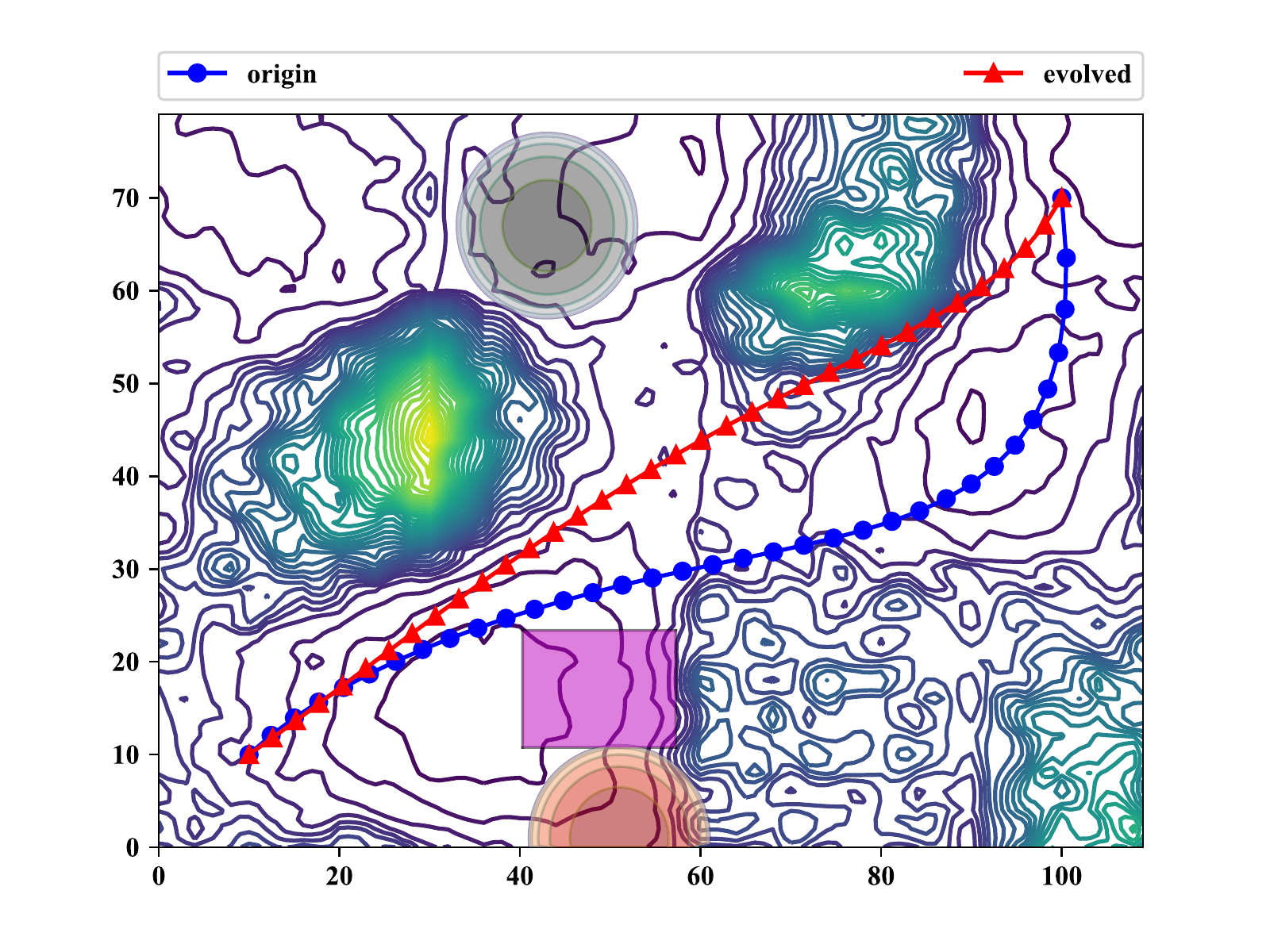}
        \caption{}
        \label{little_map1}
    \end{subfigure}
    \hfill
   \begin{subfigure}[b]{0.23\textwidth}
        \centering
       \includegraphics[width=0.95\textwidth]{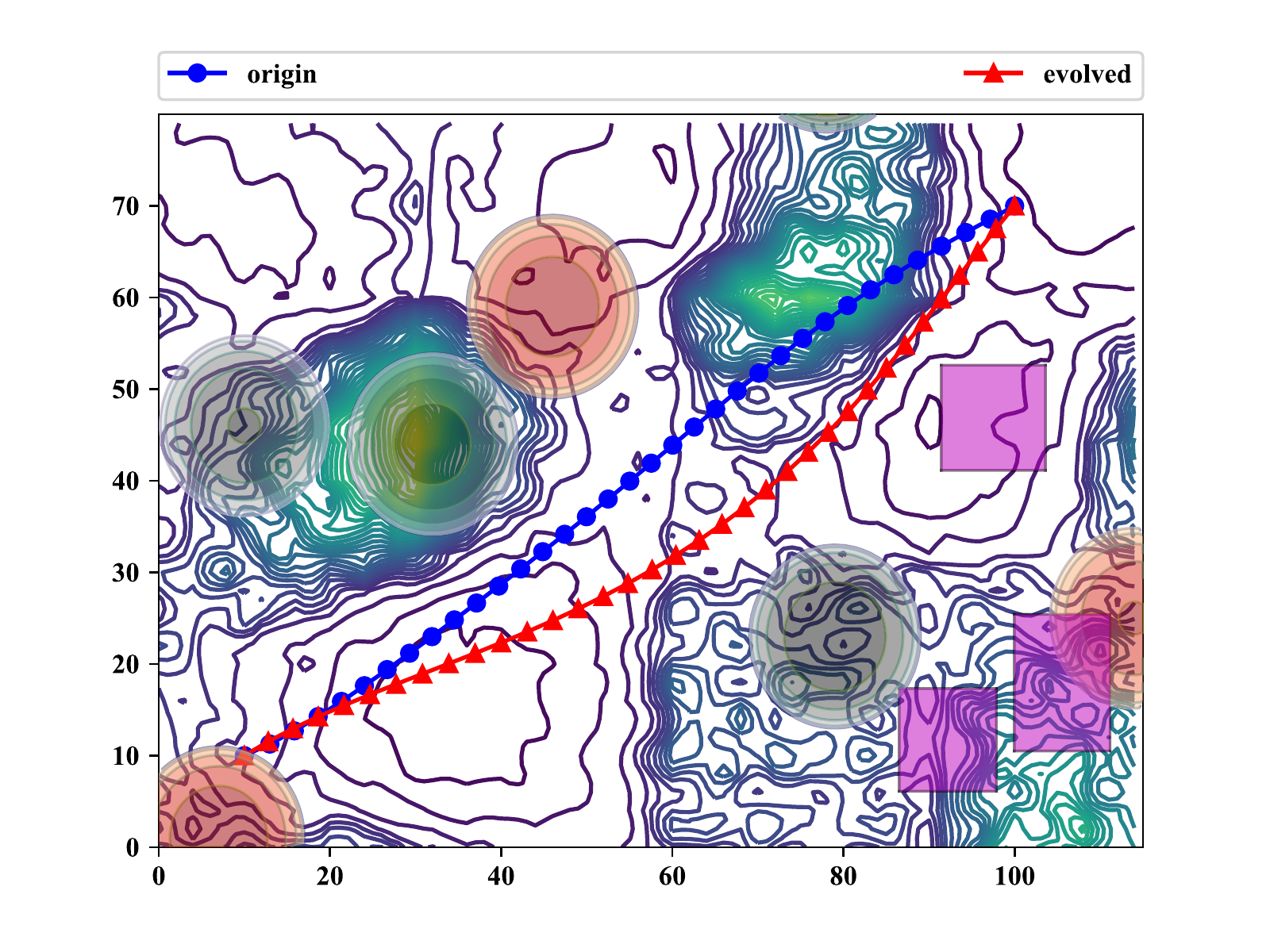}
        \caption{}
        \label{mid_map1}
    \end{subfigure}
        \hfill
   \begin{subfigure}[b]{0.23\textwidth}
        \centering
       \includegraphics[width=0.95\textwidth]{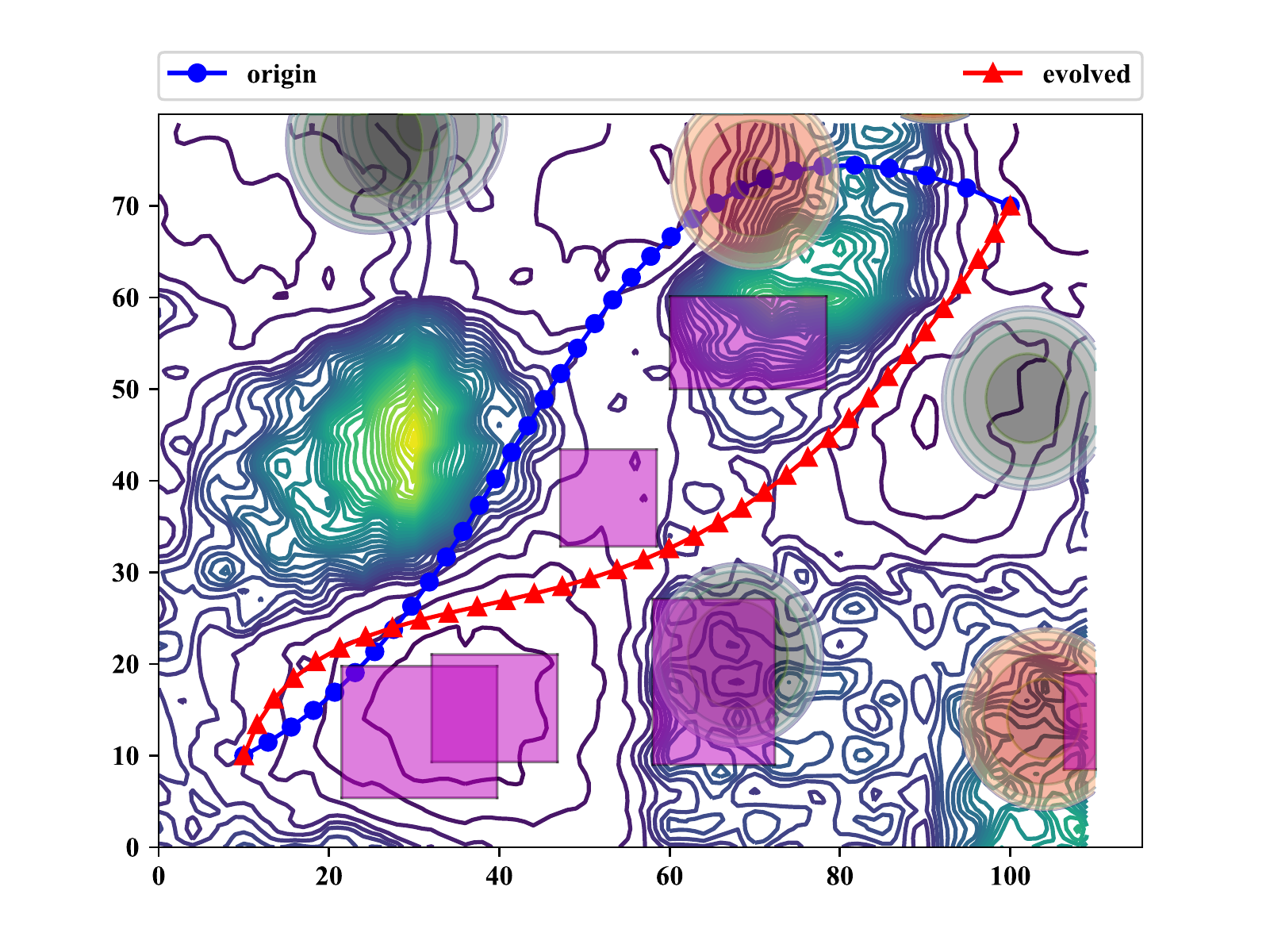}
        \caption{}
        \label{lot_map1}
    \end{subfigure}
        \hfill
   \begin{subfigure}[b]{0.23\textwidth}
        \centering
       \includegraphics[width=0.95\textwidth]{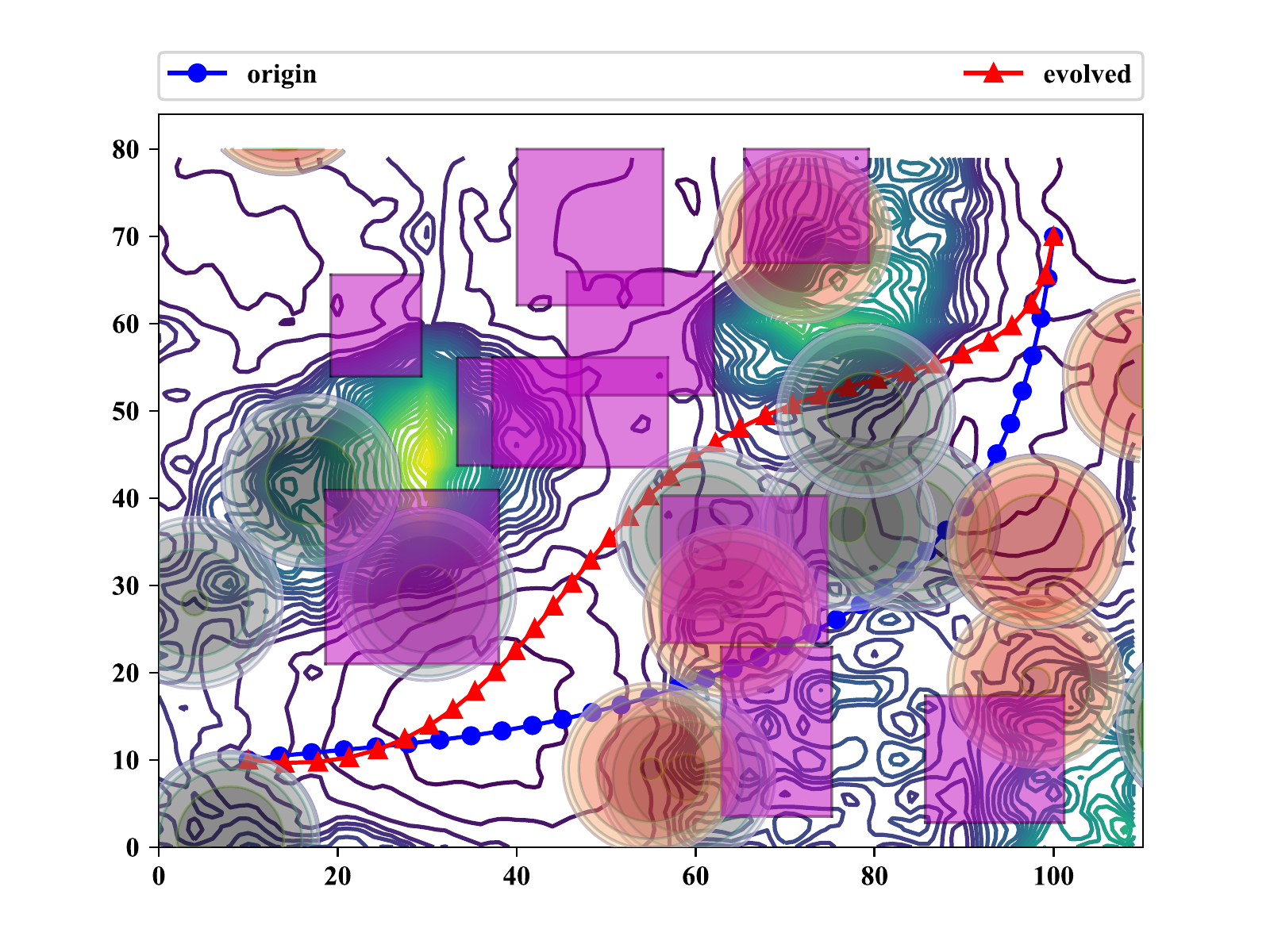}
        \caption{}
        \label{huge_map1}
    \end{subfigure}
    \caption{
    \centering
    The trajectory planning results in four scenarios with different density of obstacles. The rectangle represents the no-fly zone, the gray and orange circle represents the missile and radar areas, respectively.
    }
    \label{2d_obstacle_density}
    \end{figure}
    Besides, the numerical results are list in Table \ref{numerical result of changing obstacle}. We can clear see that evolved planner have been shown superior to the origin one. On the one hand, the former have meet all the constraints in a relatively small amount of time. However, there is an exception in the dense condition, the evolved planner took more time than the original one, which is worth since the result of avoiding breaking into no-fly zones. On the other hand, the evolved planner has got the solution with better fitness in all scenes, which means the trajectory has a smaller length cost, lower flight altitude, smoother turning, lower probability of radar detection and missile attack.\par
    \begin{table}[htbp]  
    \centering
    \caption{Performances Measures Under Changing Obstacles} 
    \renewcommand{\arraystretch}{1.1} 
    \setlength\tabcolsep{2pt} 
    \resizebox{0.48\textwidth}{!}{   
    \begin{tabular}{ccccccc}
    \hline
\multirow{2}{*}{Obstacle Density} & \multicolumn{3}{c}{Original planner}      & \multicolumn{3}{c}{Evolved planner}             \\ \cline{2-7} 
       & Constraint            & fitness & Time(s) & Constraint      & fitness       & Time(s)       \\ \hline
    sparse           & {[}0,0,0,0,0{]} & 0.56    & 0.27          & {[}0,0,0,0,0{]} & \textbf{0.38} & \textbf{0.13} \\
    not too much      & {[}0,0,0,0,0{]} & 0.64    & 0.38          & {[}0,0,0,0,0{]} & \textbf{0.38} & \textbf{0.09} \\
    a bit more       & {[}0,0,0,1,0{]} & 0.52    & 0.44          & {[}0,0,0,0,0{]} & \textbf{0.41} & \textbf{0.12} \\
    dense            & {[}0,0,0,3,0{]} & 1.38    & \textbf{0.58} & {[}0,0,0,0,0{]} & \textbf{0.61} & 0.67          \\ \hline
    \end{tabular}}
    \label{numerical result of changing obstacle}
    \end{table}
    
    \item Topographic relief\par
    In this kind of scenarios, we have removed all artificial restrictions and just simulate different natural wild scenes, e.g. base scenario, high mountain, canyon and hills. The results are shown in Fig.\ref{2d_topographic_relief} and the numerical results are list in Table \ref{numerical result topographic}. From these materials, the performances of evolved planners perform well in all terrains and are all better than the original planners. In addition, we observed that the original planners haven't met the constraint of safe flight height in three scenarios. This may be because in the case of severe terrain fluctuations, the UAV cannot break through the limitation of climbing angle and thus cannot reach a safe height under the expectation of smaller distance cost and lower altitude. But evolved planners do a good job of reconciling these contradictions.
    
    \begin{figure}[htb!]
    \centering
     \begin{subfigure}[b]{0.22\textwidth}
        \centering  
       \includegraphics[width=1\textwidth]{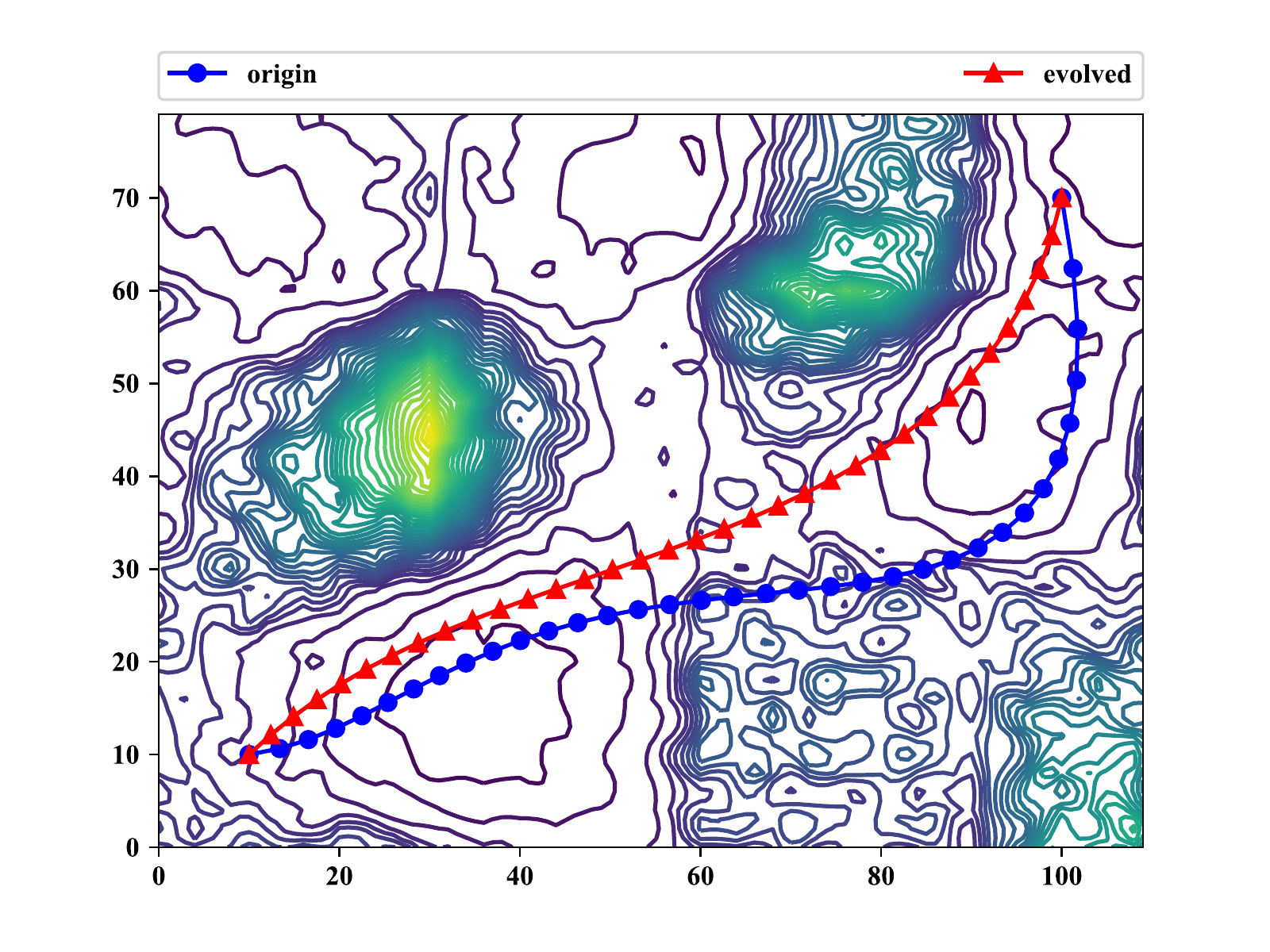}
        \caption{}
        \end{subfigure}
    \hfill
   \begin{subfigure}[b]{0.22\textwidth}
        \centering
       \includegraphics[width=1\textwidth]{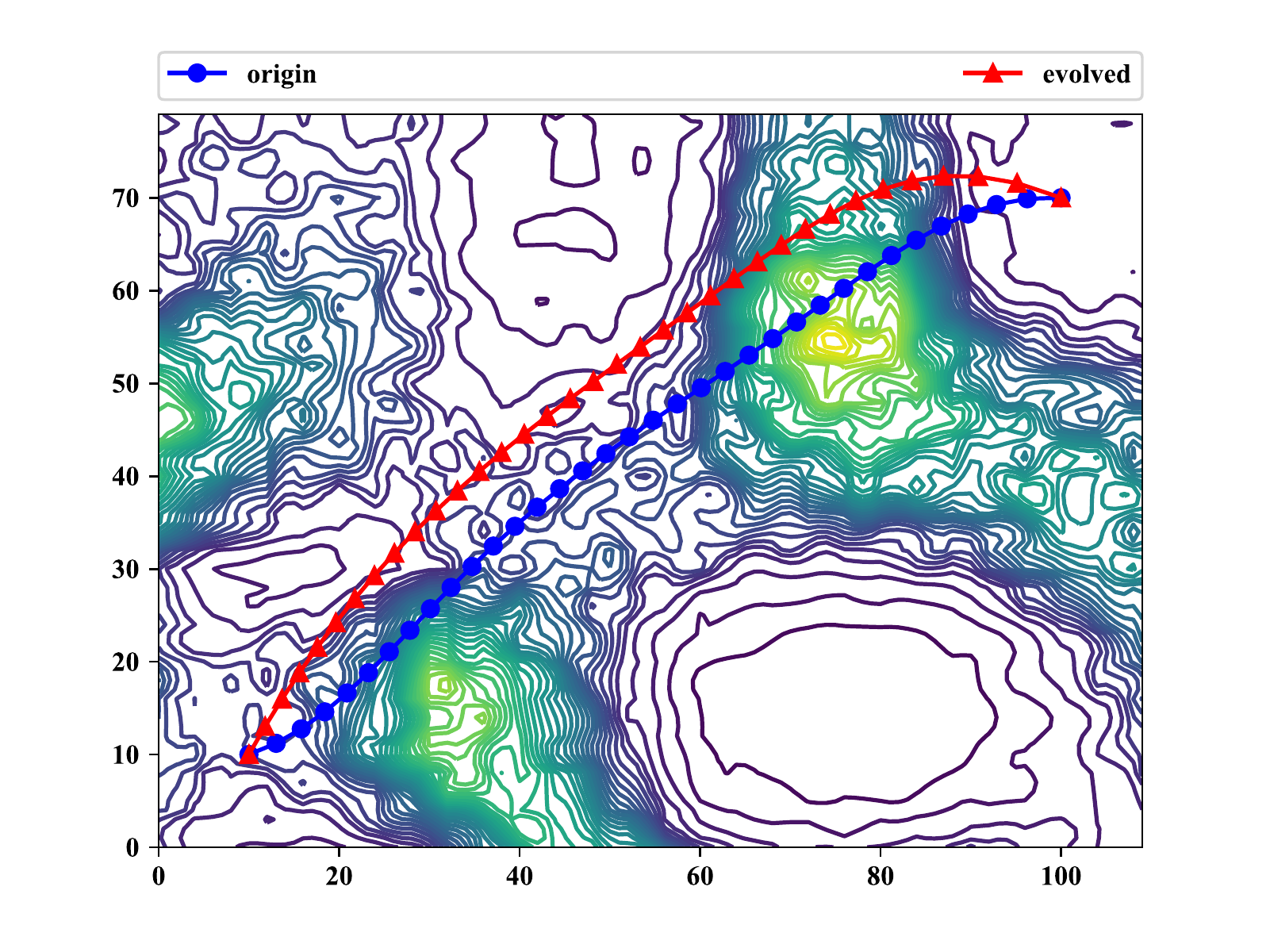}
        \caption{}
        \label{High mountain}
    \end{subfigure}
    \hfill
   \begin{subfigure}[b]{0.22\textwidth}
        \centering
       \includegraphics[width=1\textwidth]{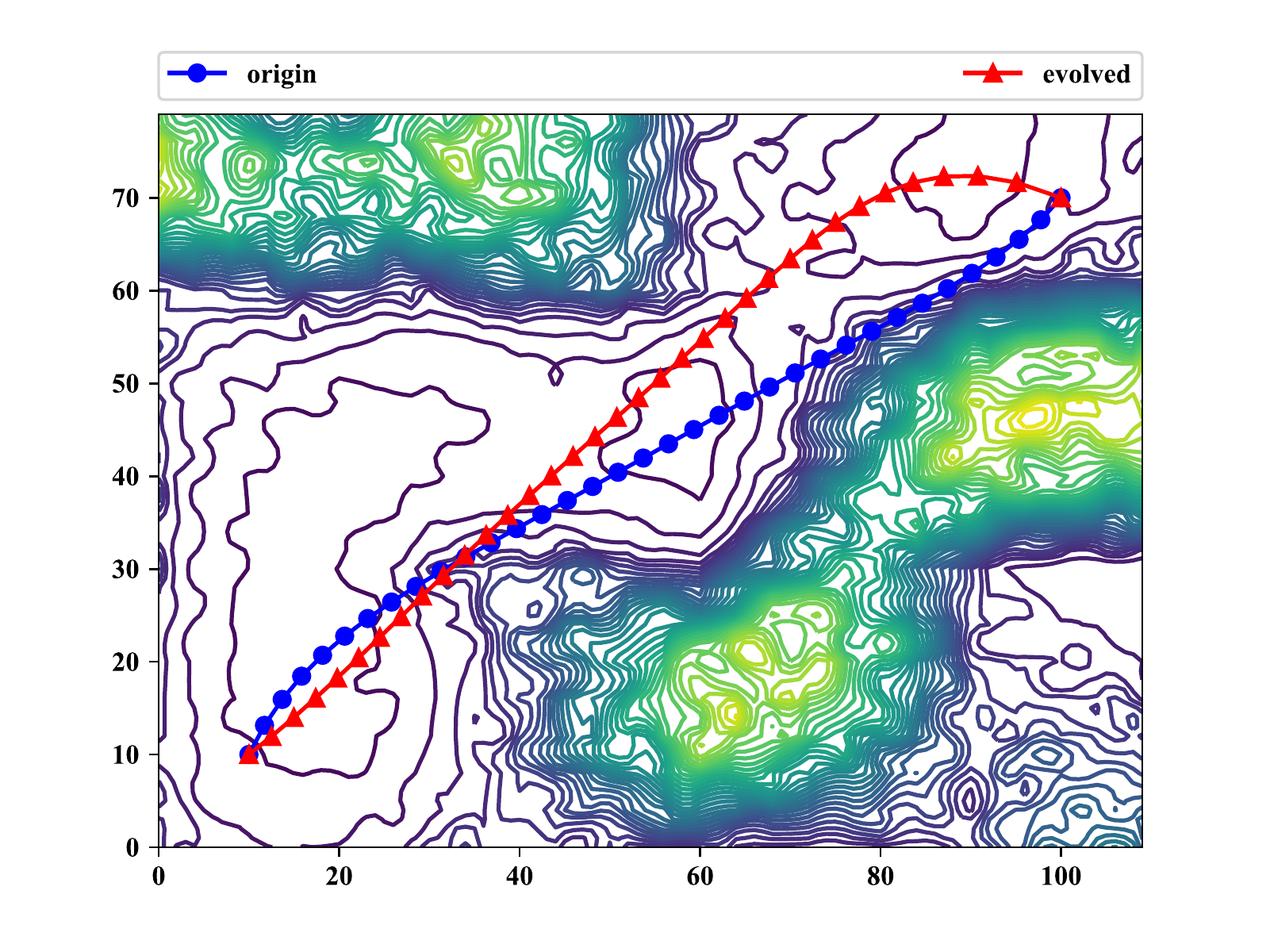}
        \caption{}
        \label{Canyon}
    \end{subfigure}
        \hfill
   \begin{subfigure}[b]{0.22\textwidth}
        \centering
       \includegraphics[width=1\textwidth]{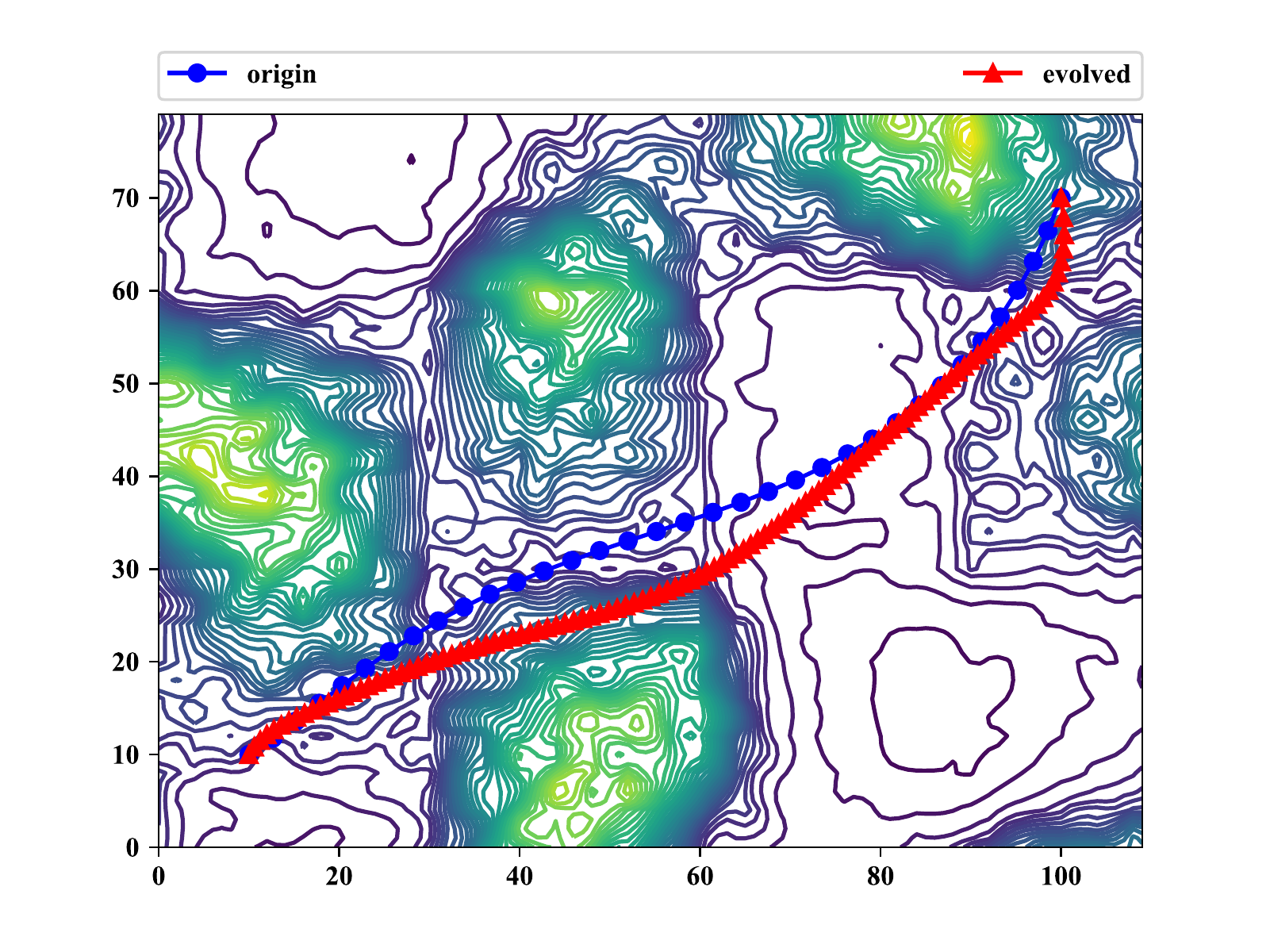}
        \caption{}
        \label{hills}
    \end{subfigure}
    \caption{
    \centering
    The trajectory planning results in four scenarios in four scenarios with different terrains. 
    }
    \label{2d_topographic_relief}
    \vspace{-5mm}
    \end{figure}
    
    \begin{table}[htbp]  
    \centering
    \caption{Performances Measures Under different terrains}  
    \renewcommand{\arraystretch}{1.1} 
    \setlength\tabcolsep{2pt} 
    \resizebox{0.48\textwidth}{!}{ 
    \begin{tabular}{ccccccc}
    \hline
    \multirow{2}{*}{Map Description} & \multicolumn{3}{c}{Original planner}      & \multicolumn{3}{c}{Evolved planner}             \\ \cline{2-7} 
                                     & Constraint            & fitness & Time(s) & Constraint      & fitness       & Time(s)       \\ \hline
    Basic Scenario                   & {[}0,0,0,0,0{]}       & 0.89    & 0.24    & {[}0,0,0,0,0{]} & \textbf{0.53} & \textbf{0.09} \\
    High Mountain                    & {[}0,0.2,0.2,0,0{]} & 1.02    & 0.25    & {[}0,0,0,0,0{]} & \textbf{0.63} & \textbf{0.14} \\
    Canyon                           & {[}0,0,0.1,0,0{]}    & 0.51    & 0.24    & {[}0,0,0,0,0{]} & \textbf{0.37} & \textbf{0.14} \\
    Hills                            & {[}0,0,0.4,0,0{]}    & 1.21    & 0.24    & {[}0,0,0,0,0{]} & \textbf{0.55} & \textbf{0.18} \\ \hline
    \end{tabular}}
    \label{numerical result topographic}
    \end{table}
    
    \item Starting\&ending points\par
    We have now demonstrated that EP is adaptive to new scenarios, but we don't know whether such an adaptation is scenario-specific or mission-specific. In theory, the planner EP created should be scenario-specific since only scenarios are input when EP works. To verify this hypothesis, we assign different starting\&ending points to each scenario shown in Fig.\ref{2d_obstacle_density} to represent diverse missions, and the previous evolved and original planners are ordered to perform these missions. As shown in Fig.\ref{2d_points}, the evolved planners exactly outperform the original ones in the horizontal, vertical and oblique missions of all scenarios. Concretely, the evolved planners always accurately avoid the no-fly zones and try to choose a low-lying place to reach the target. For example, as shown in Fig.\ref{points_4}, three planned paths accurately pass through the small gap where the no-fly zones intersect,  while the last mission is blocked by the no-fly zone in the middle, so the evolved planner chooses to reach the target from the mountain side at the edge.\par
    \begin{figure}[htb! ]
    \centering
   \begin{subfigure}[b]{0.24\textwidth}
        \centering  
       \includegraphics[width=0.95\textwidth]{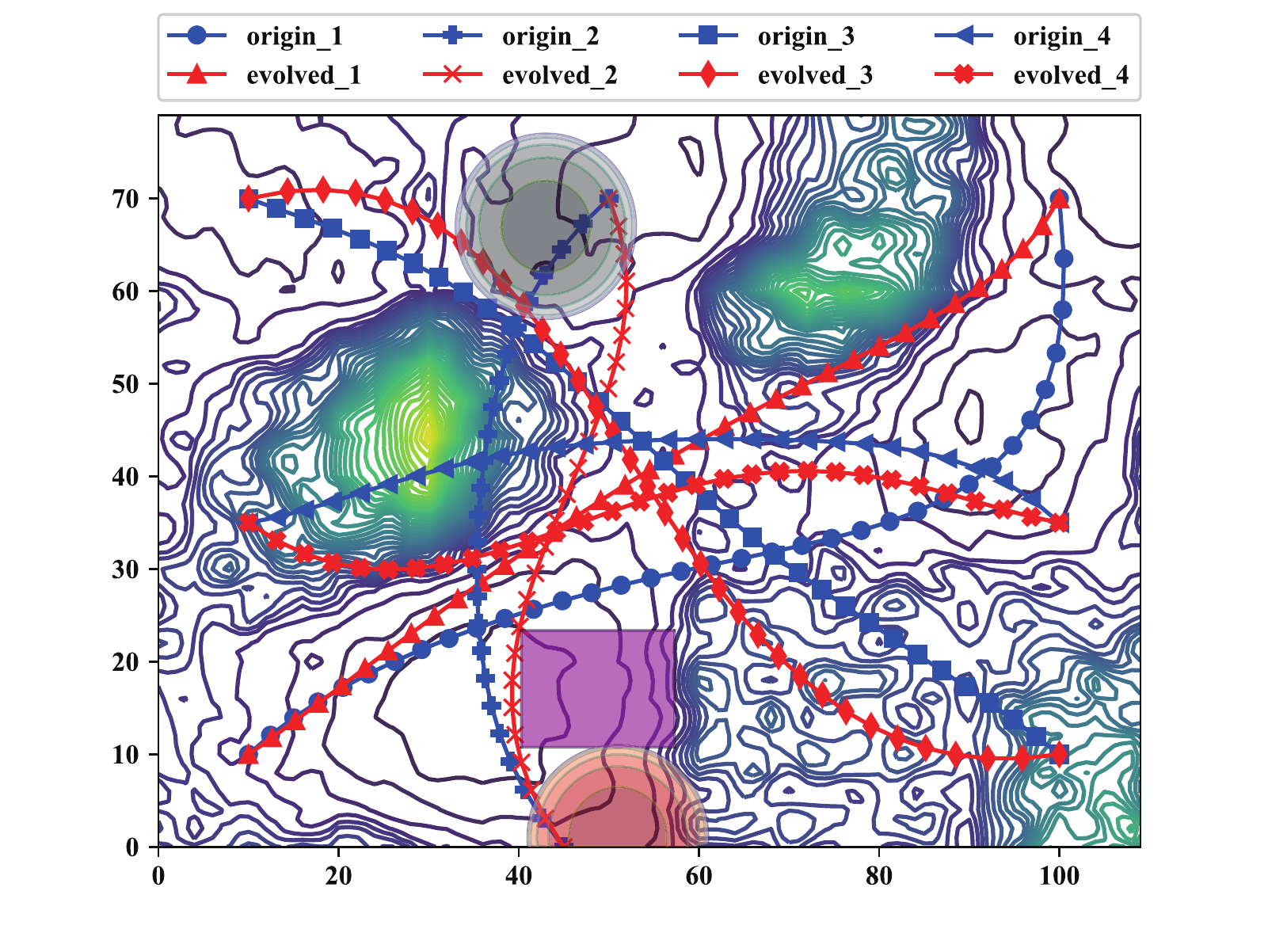}
        \caption{}
        \label{points_1}
    \end{subfigure}
    \hfill
    \begin{subfigure}[b]{0.24\textwidth}
        \centering
       \includegraphics[width=0.95\textwidth]{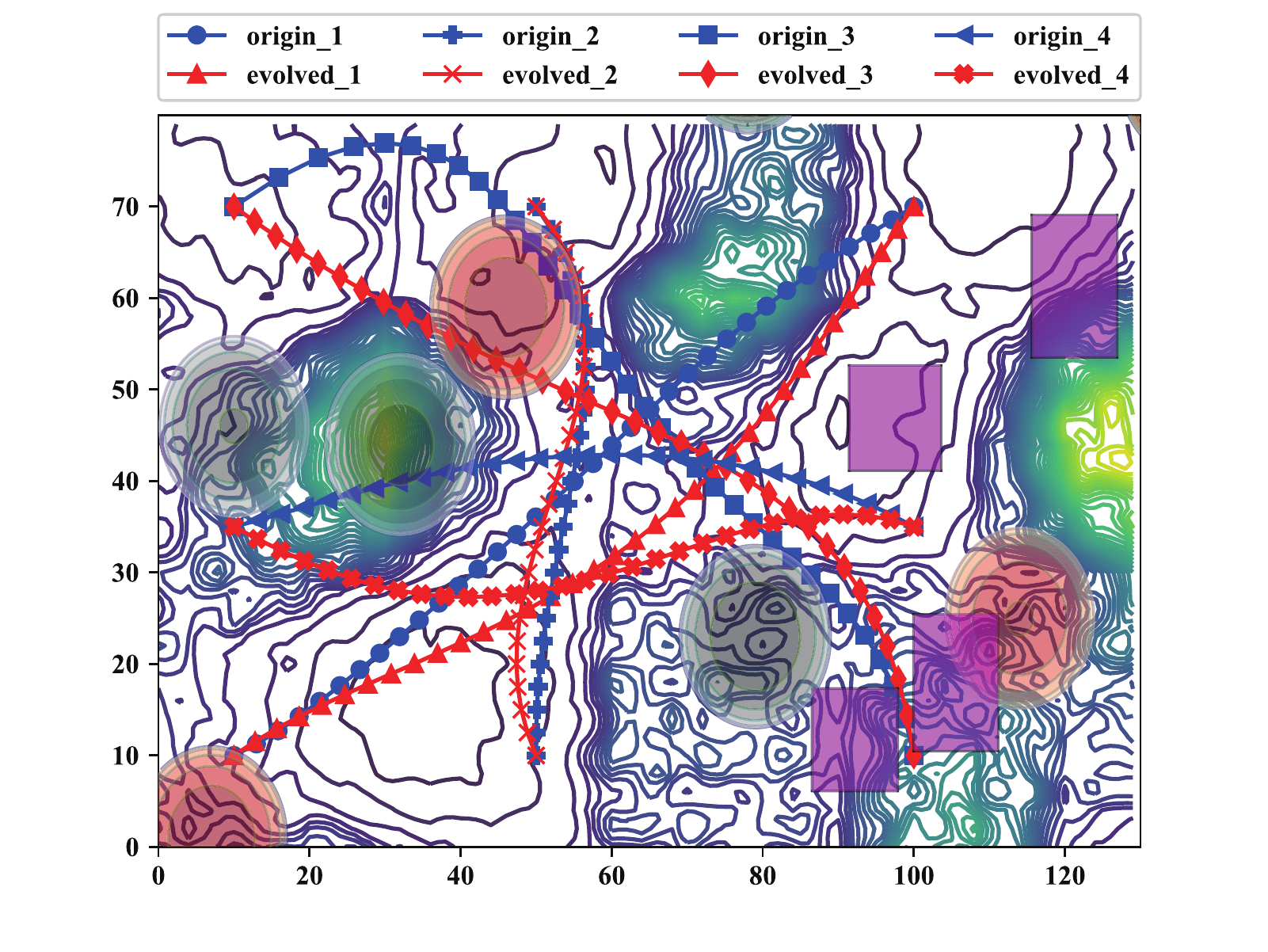}
        \caption{}
        \label{points_2}
    \end{subfigure}
        \hfill
   \begin{subfigure}[b]{0.24\textwidth}
        \centering
       \includegraphics[width=0.95\textwidth]{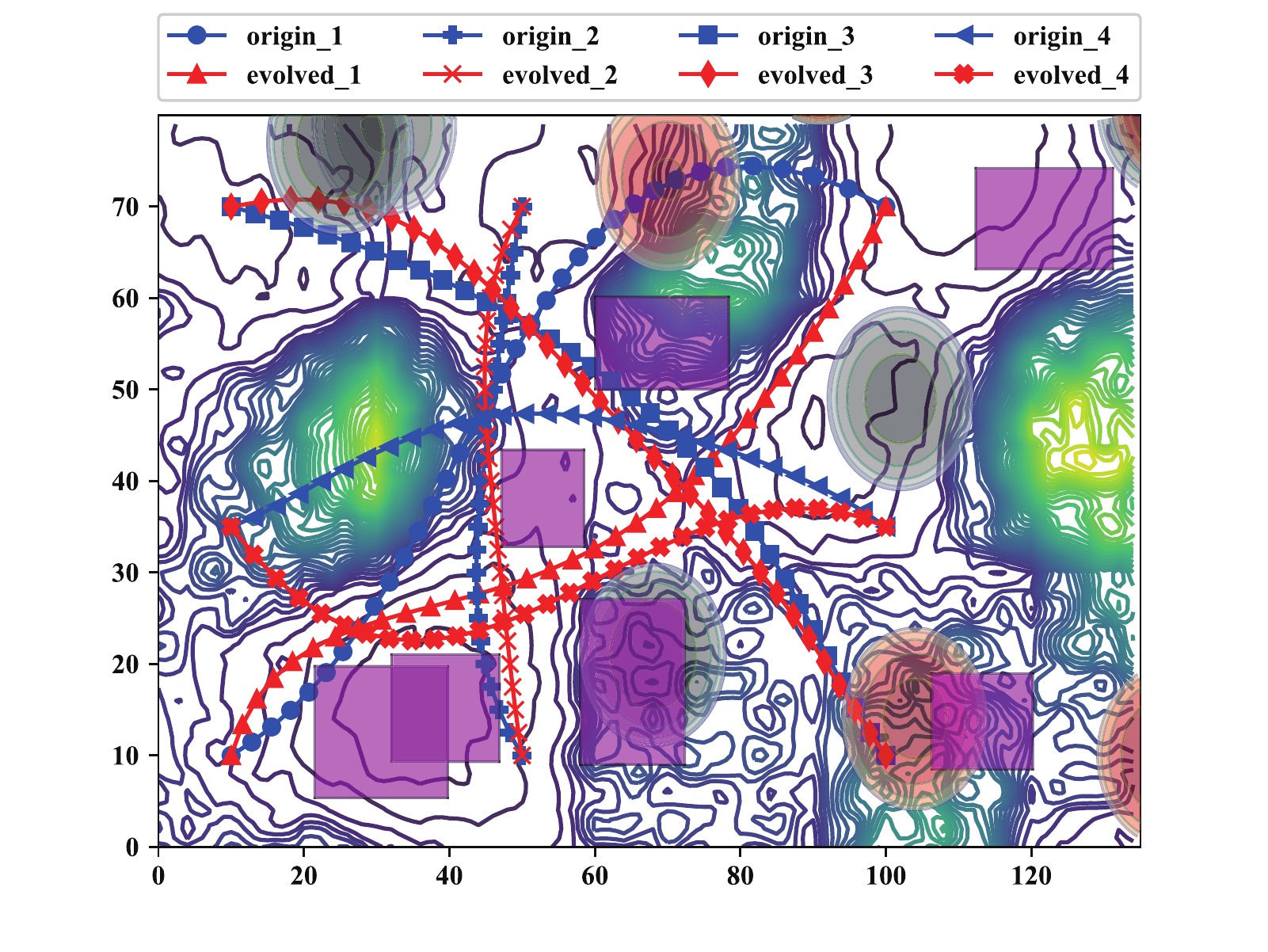}
        \caption{}
        \label{points_3}
    \end{subfigure}
        \hfill
   \begin{subfigure}[b]{0.24\textwidth}
        \centering
       \includegraphics[width=0.95\textwidth]{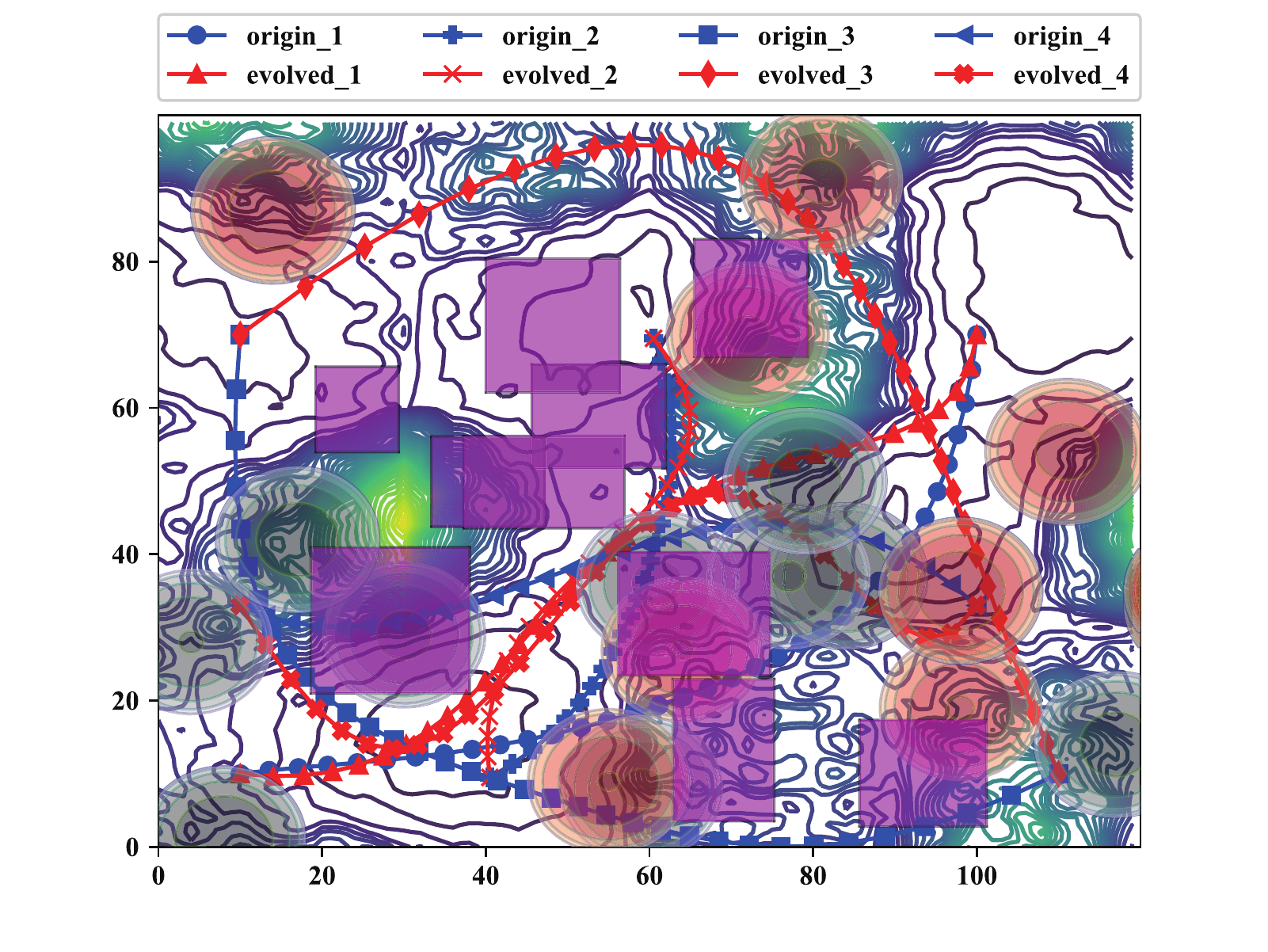}
        \caption{}
        \label{points_4}
    \end{subfigure}
    \caption{
    \centering
    The trajectory planning results with different missions in four environments. 
    }
    \label{2d_points}
    \end{figure}
    
    In order to further illustrate the adaptability of EP is scenario-specific, we conducted path planning in these scenarios for 100 times respectively with randomly generated mission points, and the statistical result were sampled in Table.\ref{numerical result of points}. Different from before, we added the values of constrains together, recorded the average measures in each scenario. It's similar to what we saw before, the evolved planners have a absolute advantage in terms of fitness and constraints, but its running time is relatively sensitive to the complexity of the environment. However, it's worth for the results of excellent performance and the running time is also within the limits permitted by practice.\par
    
    \begin{table}[htbp]  
    \centering
    \caption{Average Performances Measures with different missions in four environments}  
    \renewcommand{\arraystretch}{1.1} 
    \setlength\tabcolsep{2pt} 
    \resizebox{0.48\textwidth}{!}{ 
    \begin{tabular}{ccccccc}
        \hline
        \multirow{2}{*}{Obstacle density}       & \multicolumn{3}{c}{Original planner} & \multicolumn{3}{c}{Evolved planner}        \\ \cline{2-7} 
          & Constraint & fitness & Time(s)       & Constraint & fitness       & Time(s)       \\ \hline
        sparse                                                                            & 0.95       & 0.75    & 0.28          & 0          & \textbf{0.28} & \textbf{0.19} \\
        not too much                                                                      & 1.52       & 0.89    & 0.28          & 0          & \textbf{0.48} & \textbf{0.12} \\
        a bit more                                                                        & 3.52       & 0.65    & 0.52          & 0.3        & \textbf{0.51} & \textbf{0.23} \\
        dense                                                                             & 7.49       & 1.43    & \textbf{0.48} & 0.6        & \textbf{0.91} & 0.75 \\ \hline
    \end{tabular}}
    \label{numerical result of points}
    \end{table}
    
\end{enumerate}
\subsection{Performance Comparison}
\label{Performance Comparison}
Although we have verified that EP is self-adaptive, it has not been testified that whether the evolved planner produced by EP is superior to other planners. Furthermore, we will explore whether the superiority of evolve planner is unique to the current environment. Therefore, we select six recently proposed EA-based planners, i.e., GA\cite{Roberge2018}, CIPSO\cite{Shao2020}, JADE\cite{Zhang2009}, CIPDE\cite{Pan2020}, mWPS\cite{YongBo2017} and HSGWO\cite{Qu2020}, and one planner generated by EP through hundreds of scenarios named Heuristic Hybrid Partical Swarm Optimization(HHPSO) with uniform evolution generations and population size, as compared algorithms. Here,we have selected four representative scenarios Case 1-4 to test the capabilities of algorithms. As shown in \ref{compare_2d_random}, although the evolved planners cannot achieve a remarkable superiority in fitness, they can satisfy the constraints well in all scenarios, which is the most important requirement in practice. In particular, we have selected two extreme cases, i.e., case 2 and case 3, to explore EP's adaptability to soft and hard constraints. In Case 2, the UAV is banned from NFZs, which mean the trajectory planned by algorithm must not cross with the rectangle areas, and we can see that only the result of evolved planner meets this constraint. And in Case 3, the UAV need to avoid radar detection and missile attacks, which doesn't require the trajectory to be completely isolated from the display area since the terrain mask effect and altitude factor, so the path planned by evolved planner is partially within the circle areas but still satisfy constraints. 

    \begin{figure}[htb! ]
    \centering
   \begin{subfigure}[b]{0.241\textwidth}
    \ContinuedFloat
   \begin{minipage}[b]{1.0\linewidth}
           \centering
       \includegraphics[width=1\textwidth]{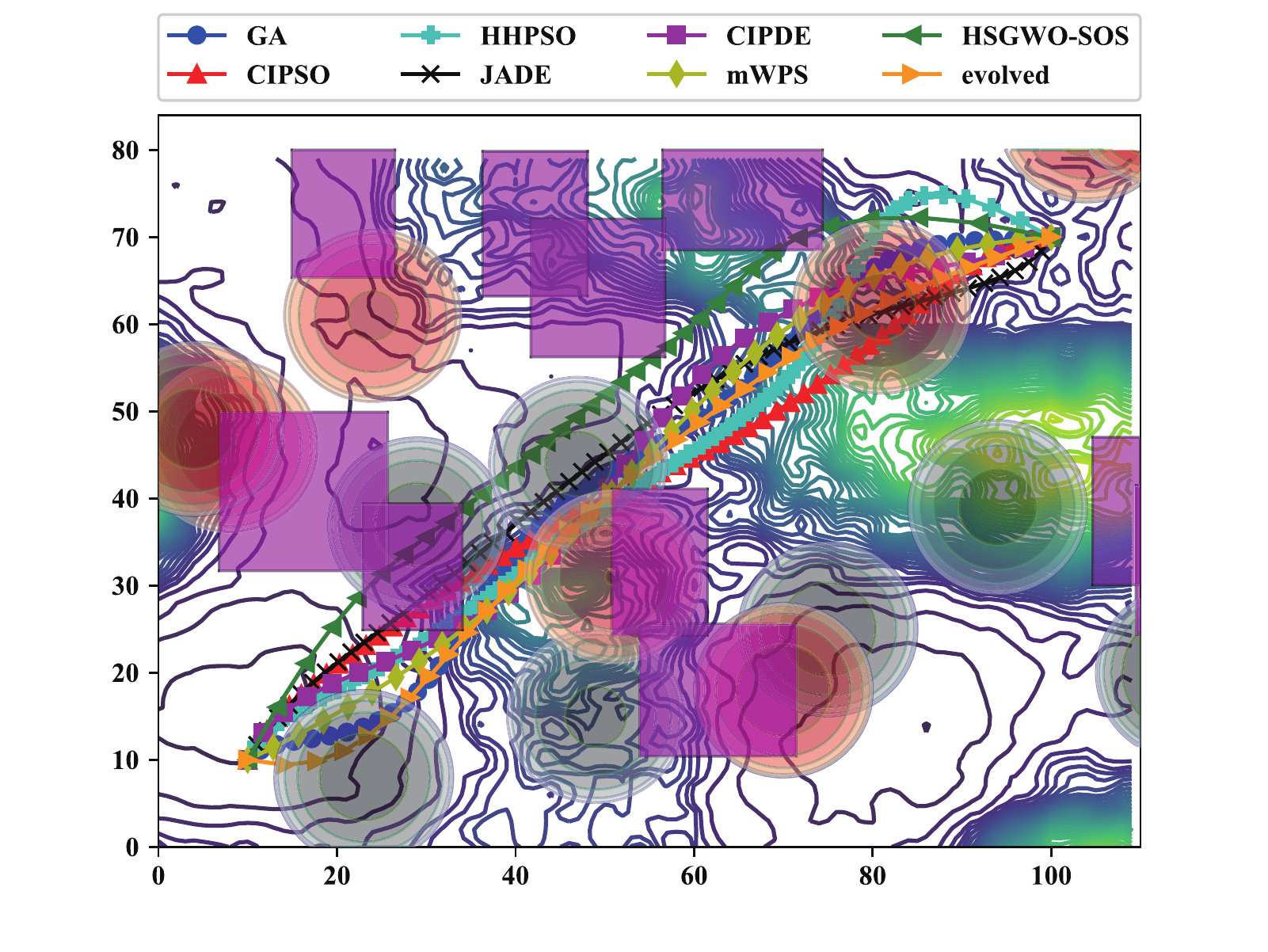}
       \includegraphics[width=0.9\textwidth]{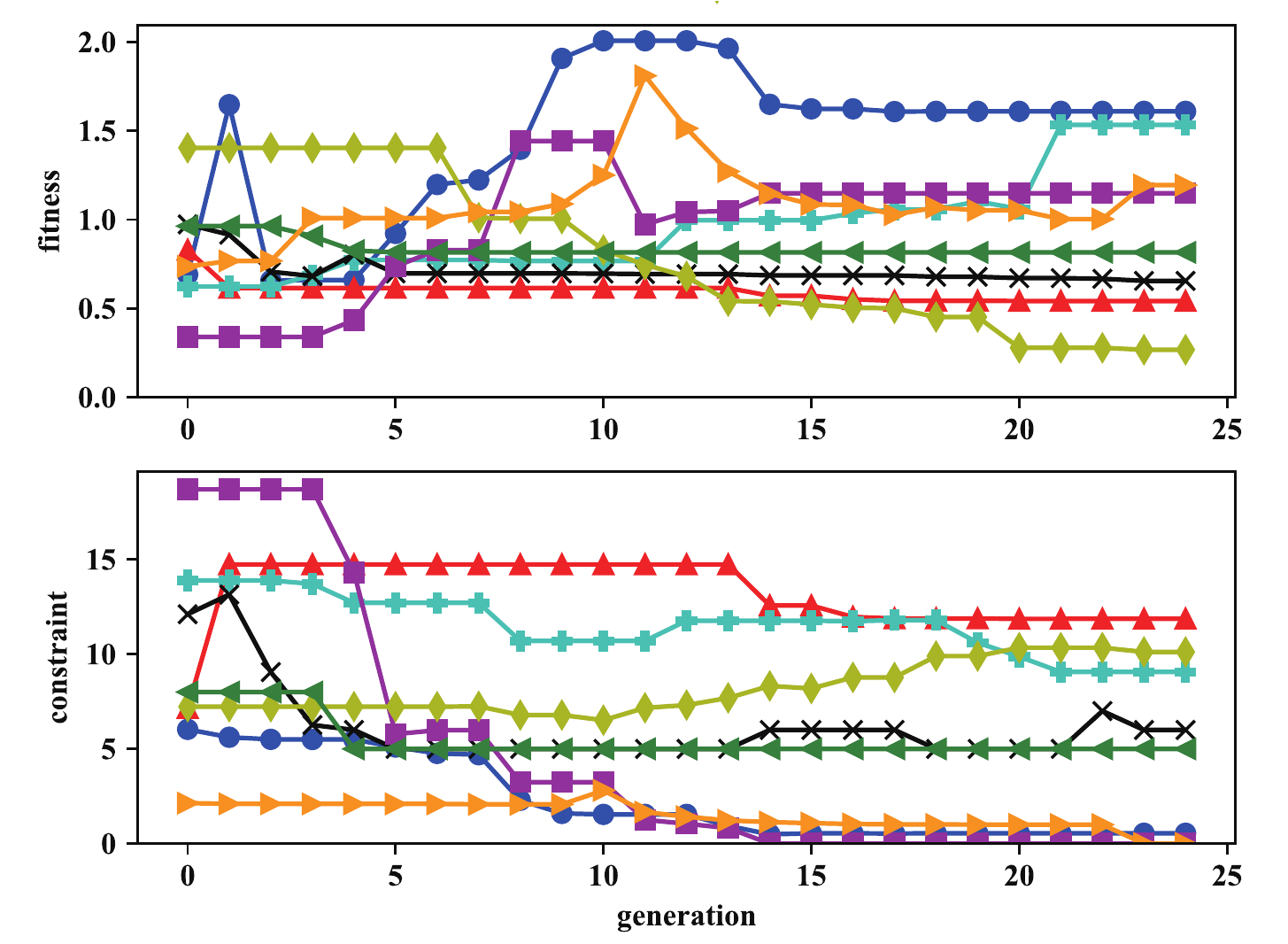}
        \caption{}
    \end{minipage}
    \end{subfigure}
   \begin{subfigure}[b]{0.241\textwidth}
    \ContinuedFloat
     \begin{minipage}[b]{1.0\linewidth}
        \centering
       \includegraphics[width=1\textwidth]{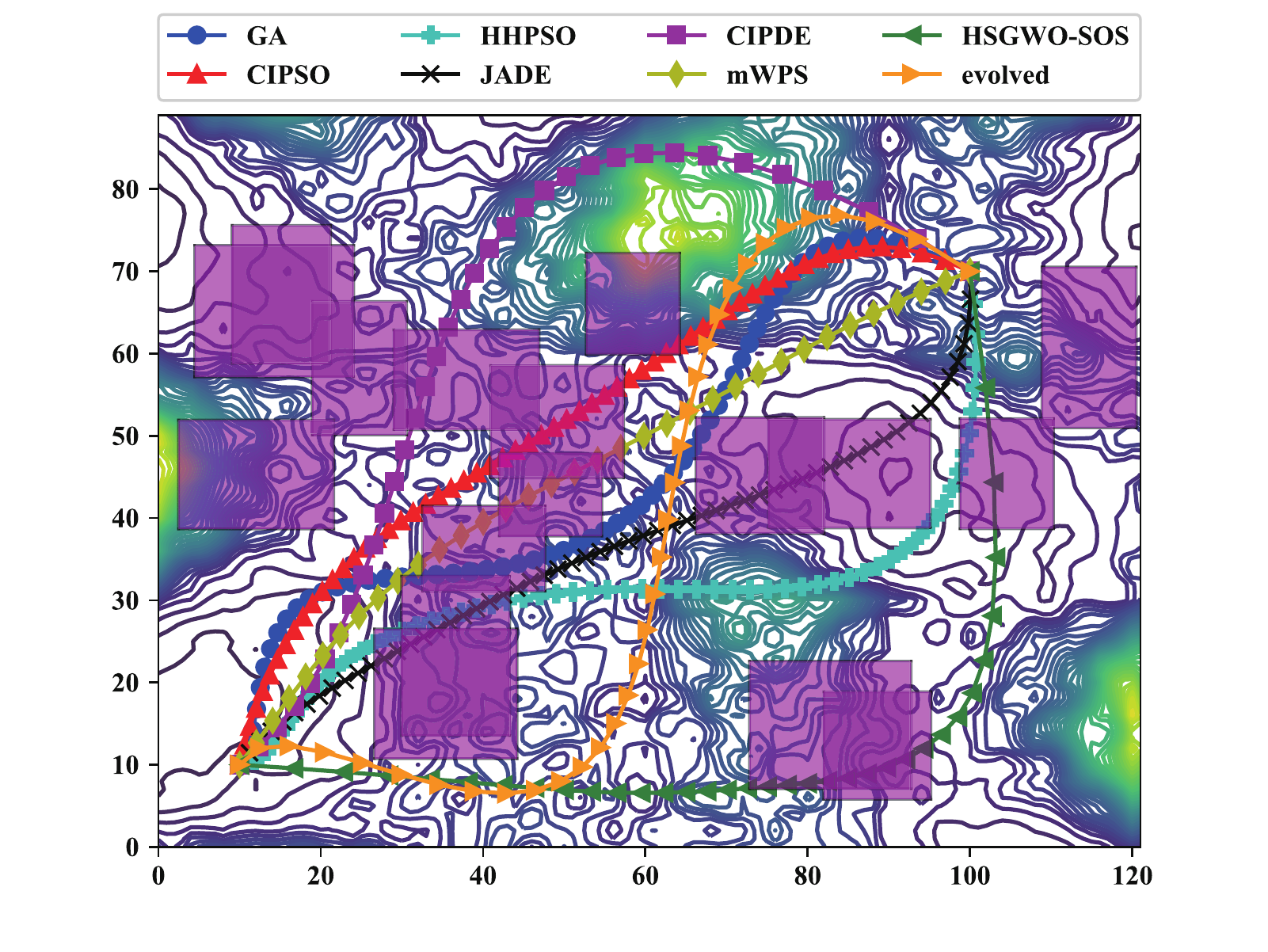}
       \includegraphics[width=0.9\textwidth]{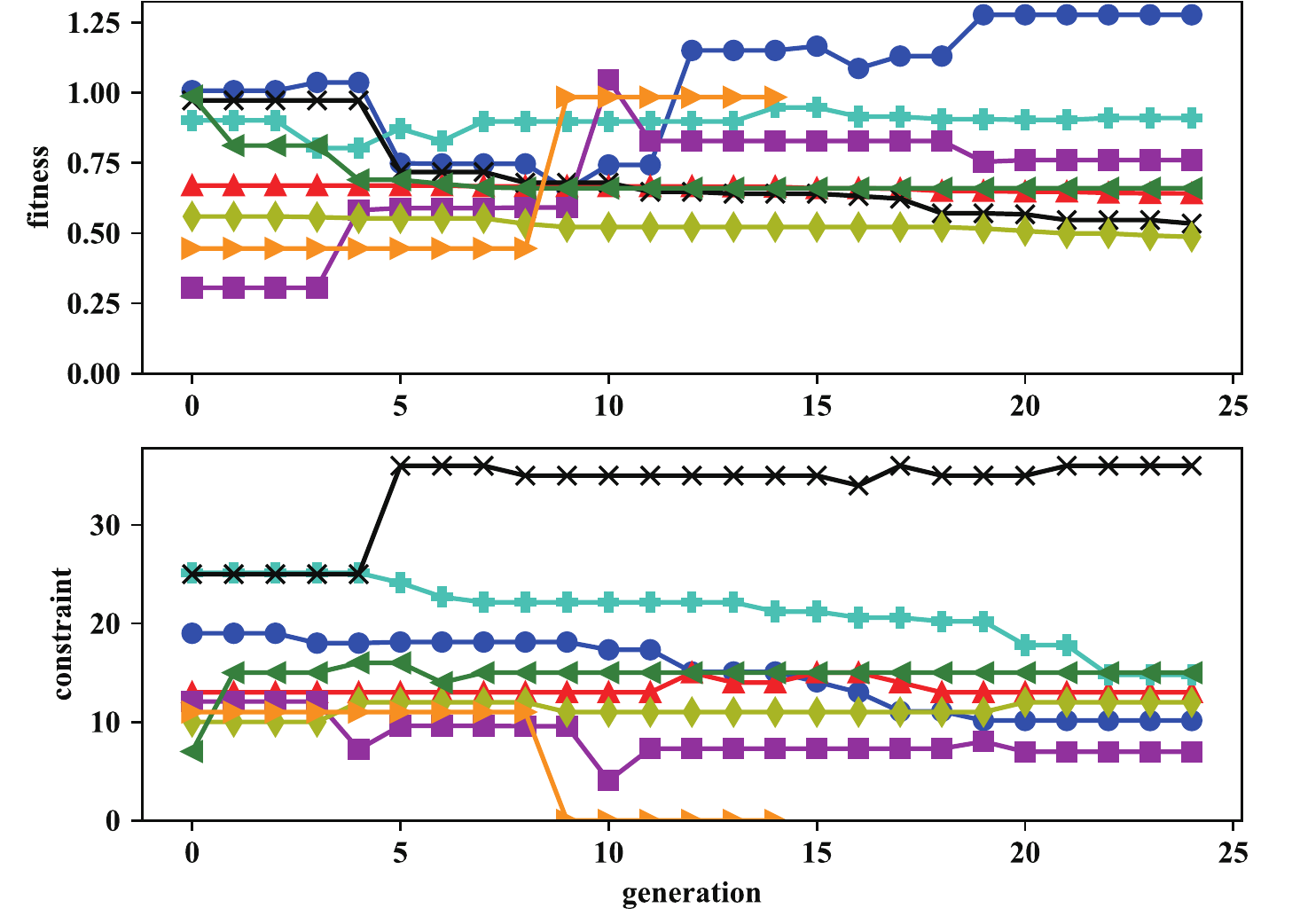}
        \caption{}
    \end{minipage}
    \end{subfigure}
   \begin{subfigure}[b]{0.241\textwidth}
    \ContinuedFloat
      \begin{minipage}[b]{1.0\linewidth}
        \centering
       \includegraphics[width=1\textwidth]{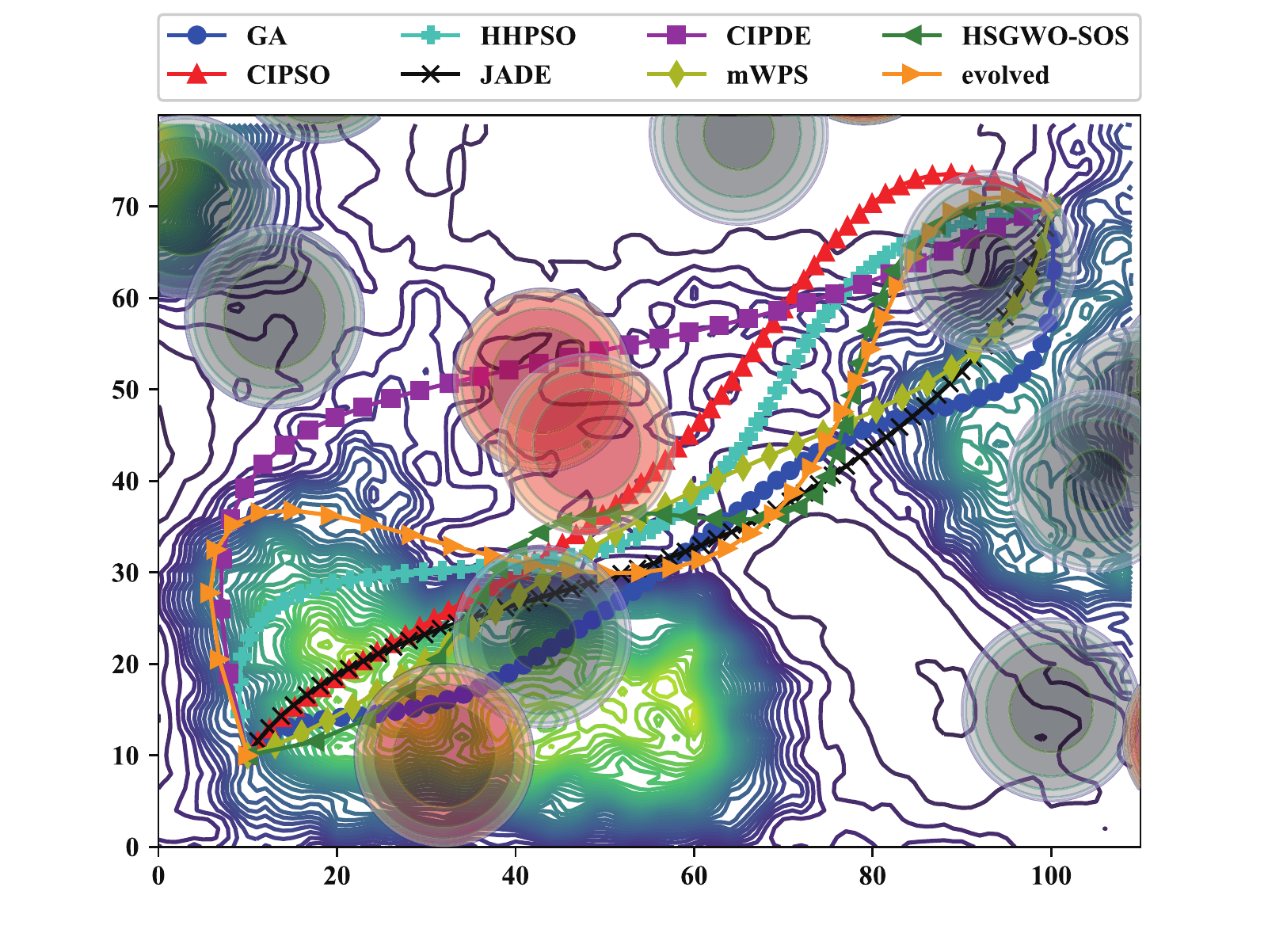}
       \includegraphics[width=0.9\textwidth]{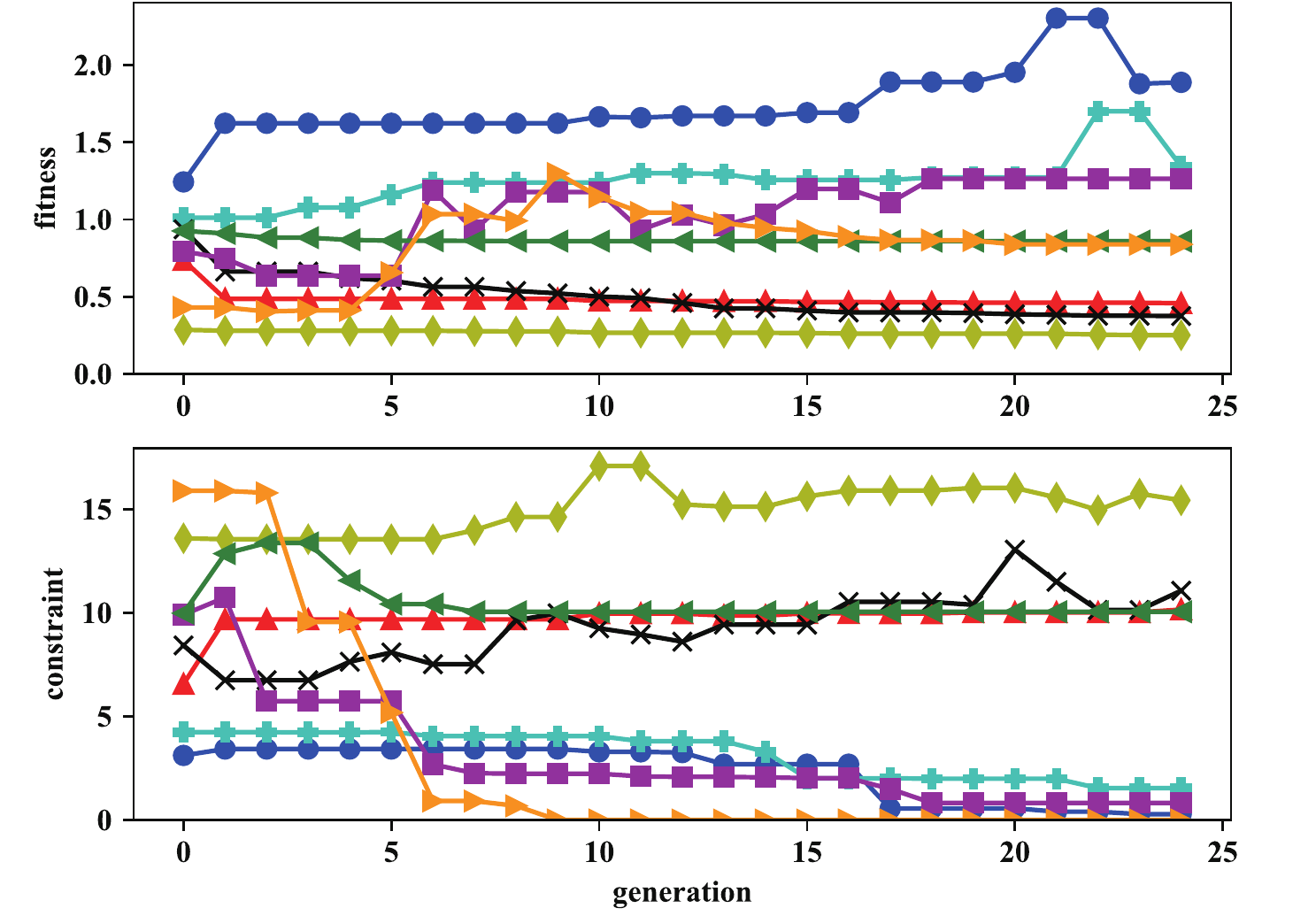}
        \caption{}
        \end{minipage}
    \end{subfigure}
   \begin{subfigure}[b]{0.241\textwidth}
    \ContinuedFloat
       \begin{minipage}[b]{1.0\linewidth}
        \centering
       \includegraphics[width=1\textwidth]{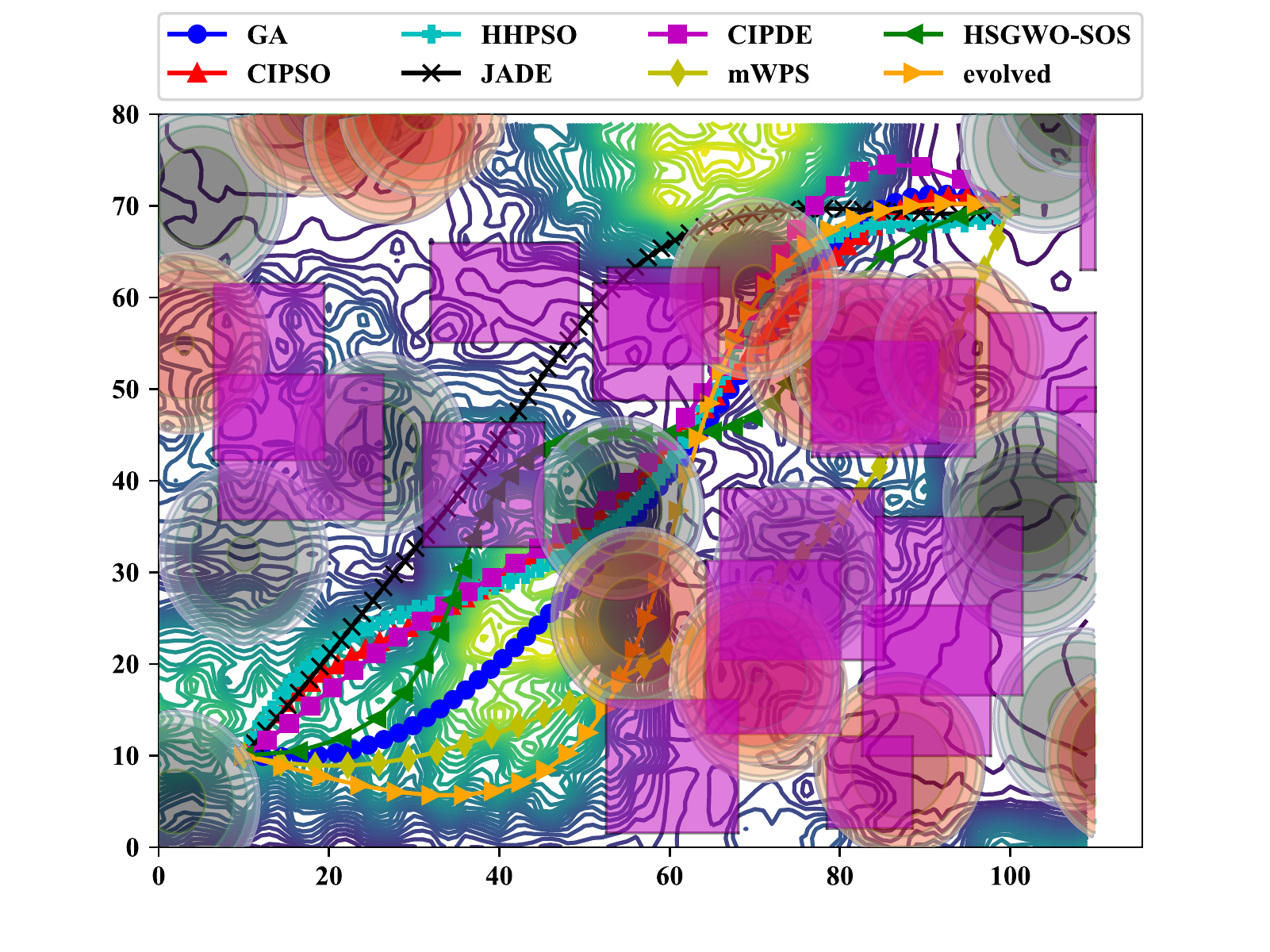}
       \includegraphics[width=0.9\textwidth]{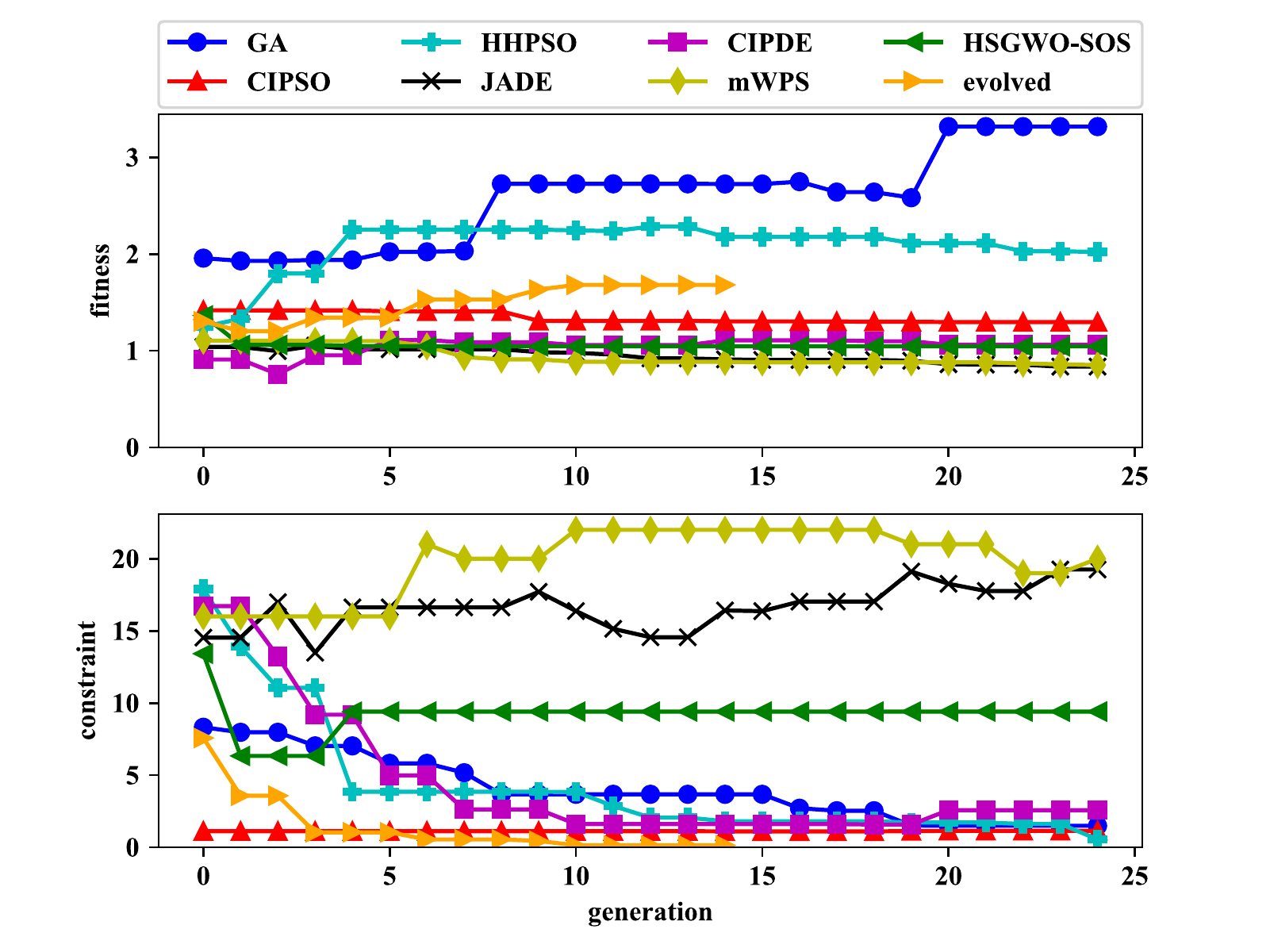}
        \caption{}
        \end{minipage}
    \end{subfigure}
    \caption{
    \centering
    The comparative results among different algorithms.
    }
    \label{compare_2d_random}
    \end{figure}
    To compare the performance of different algorithms, we ran all algorithms 100 times in each scenario and adopted several metrics to measure the final output, which are Successful Rate(\textit{SR}), Average Fitness(\textit{AF}) and Average Time(\textit{AT}). Here we define a successfully planned path where no more than one constraint is violated and the constraint value does not exceed 0.1, which means the UAV can be allowed up to 0.1m bellow the safe height or slightly exceed the climbing/gliding angle. The results are recorded in Table \ref{statistical result compare}, of all the algorithms, EP achieve 96.5\% \textit{SR} in about 0.3 seconds of \textit{AT} in these complex environment, which is superior to the rest of the algorithms. JADE can achieve the optimal \textit{AF}  with a relatively long \textit{At} but only 25.75 \textit{SR}, which is not practical for this particular problem. The same as mWPS, CIPSO and HSGWO. In contrast, other algorithms focus on the satisfaction of constraints and achieve a relatively higher \textit{SR} compared with the aforementioned algorithms, but thus have lost part of fitness. Essentially, the HHPSO is created by EP with training from a large number of scenarios. Although it has a higher \textit{SR} than other algorithms, there is still a big gap with EP. 
    From the above discussions, the planners programed by EP is more effective and efficient than the compared algorithms. On the one hand, the former can achieve the highest \textit{SR} with relative less \textit{AT} in particular scenario. On the other hand, the planners generated by EP with different scenes does not adapt well to the new environment. This illustrates that the superiority of evolve planner is unique to the current environment, and the effect of EP is to constitute such a planner.
    \begin{table}[htbp]  
        \centering
        \caption{Statistical Results of Different Algorithms}  
        \renewcommand{\arraystretch}{1.1} 
        \setlength\tabcolsep{2pt} 
        \resizebox{0.48\textwidth}{!}{ 
\begin{tabular}{lccccccccc}
    \hline
            &                 & GA            & CIPSO       & HHPSO & JADE        & CIPDE  & mWPS   & HSGWO & EP     \\ \hline
\multicolumn{1}{c}{\multirow{3}{*}{AVR}}    & \textit{SR}(\%)          & 65.5          & 16.5        & 74    & 25.75       & 58.25  & 36     & 35.5  & \textbf{96.5}   \\
\multicolumn{1}{c}{}                        & \textit{AF}     & 1.515         & 0.7625      & 1.56  &\textbf{0.545}       & 1.4325 & 0.5475 & 0.88  & 1.19   \\
\multicolumn{1}{c}{}                        & \textit{AT(s)}  & \textbf{0.2275}         & 0.6375      & 0.23  & 0.6325      & 0.6775 & 1.0775 & 1.7   & 0.2925 \\
\\
\multicolumn{1}{c}{\multirow{3}{*}{Case 1}} & SR(\%)          & 67            & 25 & 69    & 37 & 76     & 32     & 55    & \textbf{95}     \\
\multicolumn{1}{c}{}                        & \textit{AF}     & \textit{1.54} & 0.70        & 1.56  & \textbf{0.49}       & 1.57   &\textbf{0.49}   & 0.83  & 1.24   \\
\multicolumn{1}{c}{}                        & AT(s)           & 0.37          & 1.04        & 0.37  & 1.02        & 1.10   & 1.39   & 2.23  & \textbf{0.33}  \\
\\
\multirow{3}{*}{Case 2}                     & \textit{SR(\%)} & 60   & {20} & 66    & 23 & 72     & 52     & 50    & \textbf{96}     \\
            & \textit{AF}     & 1.1           & 0.53        & 1.07  & \textbf{0.41}        & 0.91   & 0.43   & 0.72  & 0.65   \\
            & \textit{AT(s)}  & 0.13          & 0.35        &\textbf{0.13}  & 0.35        & 0.37   & 0.67   & 1.12  & 0.22   \\
            \\
\multirow{3}{*}{Case 3}                     & \textit{SR(\%)} & 86            & 15 & 86    & 32 & 50     & 37     & 25    & \textbf{100}    \\
            & \textit{AF}     & 1.69          & 0.73        & 1.69  &  \textbf{0.44}       & 1.76   & 0.47   & 0.70  & 1.19   \\
            & \textit{AT(s)}  & \textbf{0.17}         & 0.48        & 0.18  & 0.48        & 0.51   & 0.92   & 1.24  & 0.44   \\
            \\
\multirow{3}{*}{Case 4}                     & \textit{SR(\%)} & 49            & 6  & 75    & 11 & 35     & 23     & 12    & \textbf{95}     \\
            & \textit{AF}     & 1.73          & 1.09        & 1.92  & 0.84        & 1.49   & \textbf{0.8}    & 1.27  & 1.68   \\
            & \textit{AT(s)}  & \textbf{0.24}         & 0.68        & \textbf{0.24} & 0.68        & 0.73   & 1.33   & 2.21  & 0.18   \\ \hline
\end{tabular}}
        \label{statistical result compare}
    \end{table}

\section{Conclusion and Discussions}
\label{conclusion}
It's very sensitive for EA-based path planner to the changes of environment. To solve this problem, this paper presents a machine learning system named EP that can autonomously create adaptive path planner. This system is inspired by the obstacles to automatic program generation, i.e., the bugs caused by simplified instruction sets and the communication difficulties between humans and machines. Based on the inspiration, the mathematical model of the problem and a specific operator library are established. In order to further speed up the convergence rate, some heuristics are involved to guarantee the feasibility of initial paths. Comparative simulation shows that the evolved planners produced by EP outperform the original ones and other state-of-the-art EA-based planners, particularly in the complex scenarios with severe terrain relief and dense obstacles around.\par
In theory, creating a program from a few basic instructions is possible but remains a practical gap. So we cut the state-of-the-art EAs of path planning problem to some operators and then recombined them into new algorithms using genetic algorithm. This approach avoids building algorithms from screws and thus can generate effective programs after a few minutes of training. Moreover, such an idea is not only applicable to path planning but also any problems sensitive to environments. \par
Although the performance of EP is remarkable in simulation, it has not been tested in the real world. Part of the reason is that there is no such a large site to place terrain and obstacles. On the other hand, the transformation between simulation and reality has been a long-standing difficulty in robotics. We may further work on these two issues. For the former, we've set up an indoor test environment, including the drones(Crazyflie 2.1)\cite{crazyflie}, the position system(Mocap) and the ground control system. We plan to place obstacles in a simulation environment and then utilize the mixed reality technology to fly the drones in the real world. For the latter, we will adopt Reinforce Learning to generate residual control signal for adapting the real world. Finally, we hope that some studies motivated by our work will be proposed to solve real-world problems, and this work could facilitate the applications of evolutionary algorithms in robotics.\par




\bibliographystyle{IEEEtran}
\bibliography{IEEEexample,ref} 






\end{document}
